\documentclass{article}

\PassOptionsToPackage{numbers, compress}{natbib}

\usepackage[dvipsnames]{xcolor}
\usepackage{color,colortbl}
\usepackage{xspace}
\definecolor{mycitecolor}{rgb}{0,0.08,0.75} 

\usepackage[final]{neurips_2022}

\usepackage[utf8]{inputenc} 
\usepackage[T1]{fontenc}    
\usepackage[colorlinks=true, citecolor=mycitecolor,backref=page]{hyperref}       
\usepackage{url}            
\usepackage{booktabs}       
\usepackage{amsfonts}       
\usepackage{nicefrac}       
\usepackage{microtype}      
\usepackage{xcolor}         

\usepackage{amsmath}       
\usepackage{graphicx}      
\usepackage{booktabs}     
\usepackage{multirow }    
\usepackage{caption}
\usepackage{subcaption}
\usepackage{listings}
\usepackage{wrapfig, blindtext}

\usepackage[capitalize]{cleveref}
\crefname{section}{Sec.}{Secs.}
\Crefname{table}{Table}{Tables}
\Crefname{figure}{Fig.}{Figs.}

\makeatletter
\DeclareRobustCommand\onedot{\futurelet\@let@token\@onedot}
\def\@onedot{\ifx\@let@token.\else.\null\fi\xspace}

\def\ie{{i.e}\onedot} 
\def\eg{{e.g}\onedot}

\newcommand{\na}[0]{\textcolor{gray}{n/a}}

\usepackage{amsmath,amsfonts,bm}

\def\1{\bm{1}}

\def\vc{{\bm{c}}}

\DeclareMathAlphabet{\mathsfit}{\encodingdefault}{\sfdefault}{m}{sl}
\SetMathAlphabet{\mathsfit}{bold}{\encodingdefault}{\sfdefault}{bx}{n}

\def\sR{{\mathbb{R}}}

\newcommand{\accai}{\mathrm{acc_\text{AI}}}
\newcommand{\acchuman}{\mathrm{acc_\text{human}}}
\newcommand{\accteam}{\mathrm{acc_\text{team}}}

\newcommand{\layer}[1]{\ensuremath{\mathsf{#1}\xspace}}
\newcommand{\class}[1]{\texttt{#1}\xspace}
\newcommand{\subsec}[1]{\noindent\textbf{#1}~~}

\newcommand{\increase}[1]{(\textcolor{ForestGreen}{+#1})}
\newcommand{\increasenoparent}[1]{\textcolor{ForestGreen}{+#1}}
\newcommand{\decrease}[1]{(\textcolor{red}{-#1})}
\newcommand{\decreasenoparent}[1]{\textcolor{red}{-#1}}

\definecolor{MyLightGray}{rgb}{0.95, 0.95, 0.95}
\definecolor{MyLightGreen}{rgb}{0.87, 0.93, 0.917}
\definecolor{MyLightPink}{rgb}{0.957, 0.87, 0.914}

\definecolor{ResNetBlue}{rgb}{0.192, 0.454, 0.643}
\definecolor{kNNOrange}{rgb}{0.89, 0.50, 0.169} 
\definecolor{EMDRed}{rgb}{0.757, 0.243, 0.239} 
\definecolor{CHMBrown}{rgb}{0.517, 0.36, 0.322} 

\newcommand{\thresholdcolor}{EEC4FF}

\newcommand{\resnet}[0]{\textcolor{ResNetBlue}{ResNet-50\xspace}}
\newcommand{\knn}[0]{\textcolor{kNNOrange}{kNN}}
\newcommand{\emdcorr}[0]{\textcolor{EMDRed}{EMD-Corr}}
\newcommand{\chmcorr}[0]{\textcolor{CHMBrown}{CHM-Corr}}

\newcommand{\papertitle}{Visual correspondence-based explanations improve AI robustness and human-AI team accuracy}
\title{\papertitle}

\author{%
  Giang Nguyen \thanks{
Equal contribution. 
Listing order is random.
GN led the development of EMD-Corr and human studies on Gorilla.
MRT led the development of CHM-Corr, pilot studies on HuggingFace, and the analysis of human-study data from Gorilla.
AN advised the project.
MRT's work was done before he joined University of Alberta.
}\\
  \texttt{nguyengiangbkhn@gmail.com} \\
   \And
   Mohammad Reza Taesiri $^*$ \\
   \texttt{mtaesiri@gmail.com}\\
   \AND
   Anh Nguyen \\
    \texttt{anh.ng8@gmail.com}\\\\
    Auburn University ~~~~~
}

\begin{document}

\maketitle

\begin{abstract}
    Explaining artificial intelligence (AI) predictions is increasingly important and even imperative in many high-stakes applications where humans are the ultimate decision makers. 
    In this work, we propose two novel architectures of self-interpretable image classifiers that first explain, and then predict (as opposed to post-hoc explanations) by harnessing the visual correspondences between a query image and exemplars. 
    Our models consistently improve (\increasenoparent{1} to \increasenoparent{4} points) on out-of-distribution (OOD) datasets while performing marginally worse (\decreasenoparent{1} to \decreasenoparent{2} points) on in-distribution tests than ResNet-50 and a $k$-nearest neighbor classifier (kNN). 
    Via a large-scale, human study on ImageNet and CUB, our correspondence-based explanations are found to be more useful to users than kNN explanations.
    Our explanations help users more accurately reject AI's wrong decisions than all other tested methods.
    Interestingly, for the first time, we show that it is possible to achieve complementary human-AI team accuracy (i.e., that is higher than either AI-alone or human-alone), in ImageNet and CUB image classification tasks.
\end{abstract}




\section{Introduction}
Comparing the input image with training-set exemplars is the backbone for many applications, such as face identification \cite{hai2022deepface}, bird identification \cite{chen2019looks,zhao2021towards}, and image search \cite{zhao2021towards}.
This non-parametric approach may improve classification accuracy on out-of-distribution (OOD) data \cite{hai2022deepface,zhang2020deepemd,zhao2021towards,papernot2018deep} and enables a class of prototype-based explanations \cite{chen2019looks,nauta2021looks,nauta2021neural,singh2021these,kenny2019twin} that provide insights into the decision making of Artificial Intelligence (AI) systems.
Interestingly, prototype-based explanations are more effective in improving human classification accuracy \cite{nguyen2021effectiveness,fel2021cannot,kim2021hive} than attribution maps---a common eXplainable AI (XAI) technique in computer vision.
Yet, it remains an open question how to make prototype-based XAI classifiers (1) accurate on in-distribution and OOD data and (2) improve human decisions.
For example, in face identification, AIs can be confused by partially occluded, never-seen faces and are unable to explain their decisions to users, causing numerous people falsely arrested \cite{lawsuit2021facial,newjersey2021arrested,detroit2021arrested,michigan2021arrested} or wrongly denied unemployment benefits \cite{MiaSato.2021} by the law enforcement.

\begin{figure*}[t]
  \centering
    \includegraphics[width=1.0\textwidth]{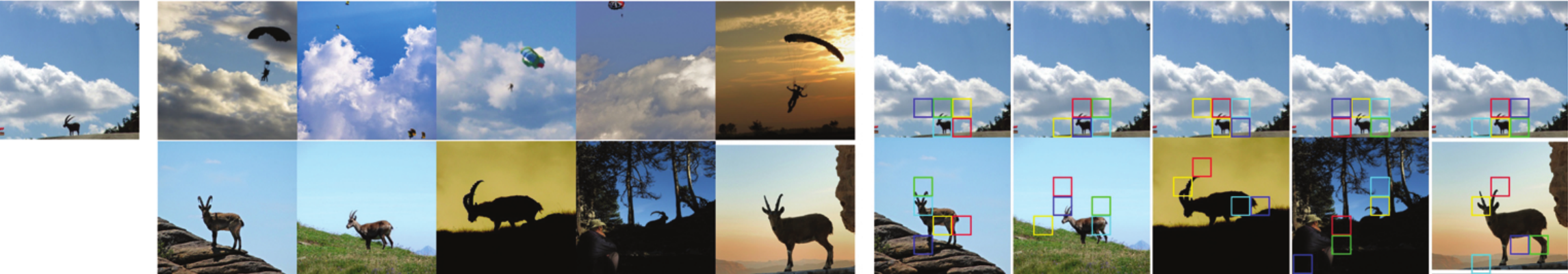}
    {
    \tiny	
		\begin{flushleft}
		\vspace{-1.3cm}
			groundtruth:\\
			\class{ibex}
		\end{flushleft}
	}
	\vspace{1cm}
	{\vspace{-0.6cm}
    \tiny	
		\begin{flushleft}
    		\hspace{1.3cm}
            (a) Explanations for kNN's \textcolor{red}{\class{parachute}} decision (top) and CHM-NN (bottom)
            \hspace{1.3cm}
            (b) Explanations for CHM-Corr's \textcolor{OliveGreen}{\class{ibex}} decision
		\end{flushleft}
	}
    \caption{
    The \class{ibex} image is misclassified into \class{parachute} due to its similarity (clouds in blue sky) to parachute scenes (a).
    In contrast, CHM-Corr correctly labels the input as it matches \class{ibex} images mostly using the animal's features, discarding the background information (b).
    }
    \label{fig:ibex}
\end{figure*}

\begin{figure*}[t]
  \centering
        \includegraphics[width=1.0\textwidth]{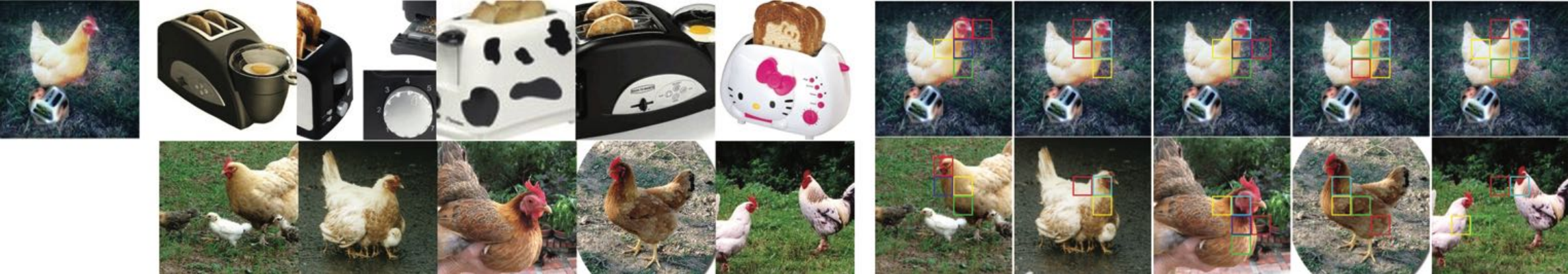}
    {
    \tiny
		\begin{flushleft}
		\vspace{-1.3cm}
			groundtruth:\\
			\class{hen}
		\end{flushleft}
	}
	\vspace{1cm}
	{\vspace{-0.6cm}
    \tiny
		\begin{flushleft}
    		\hspace{1.3cm}
            (a) Explanations for kNN's \textcolor{red}{\class{toaster}} decision (top) and EMD-NN (bottom)
            \hspace{1.3cm}
            (b) Explanations for EMD-Corr's \textcolor{OliveGreen}{\class{hen}} decision
		\end{flushleft}
	}
	
    \caption{Operating at the image-level visual similarity, kNN incorrectly labels the input \class{toaster} due to the adversarial toaster patch (a).
    EMD-Corr instead ignores the {adversarial patch} and only uses the head and neck patches of the hen to make decisions (b).
    }
    \label{fig:adversarial_hen}
    \vspace*{-0.5cm}
\end{figure*}

To address the above questions, we propose two \emph{interpretable} \cite{rudin2019stop}, (\ie, first-explain-then-decide) image classifiers that perform three common steps: (1) rank the training-set images based on their distances to the input using \emph{image-level} features; (2) re-rank the top-50 shortlisted candidates by their \emph{patch-wise} correspondences w.r.t. the input \cite{min2021convolutional,hai2022deepface}; and then (3) take the dominant class among the top-20 candidates as the predicted label.
That is, our classifiers base their decisions on a set of support image-patch pairs, which also serve as \emph{explanations} to users (Figs.~\ref{fig:ibex}b and~\ref{fig:adversarial_hen}b).
Our main findings include:
\footnote{Code and models are available at \url{https://github.com/anguyen8/visual-correspondence-XAI}.}

\begin{itemize}
    \item On ImageNet, a simple $k$-nearest-neighbor classifier (kNN) based on ResNet-50 features slightly but consistently outperforms ResNet-50 on many OOD datasets (\cref{sec:result_finding1}).
    This is further improved after a re-ranking step based on patch-wise similarity (\cref{sec:result_finding2}).
    
    \item Via a large-scale human study, we find visual correspondence-based explanations to improve AI-assisted, human-alone accuracy and human-AI team accuracy on ImageNet and CUB over the baseline kNN explanations (\cref{sec:result_finding3}).
    
    \item Having interpretable AIs label images that they are confident and humans label the rest yields better accuracy than letting AIs or humans alone label all images (\cref{sec:result_finding4,sec:human_ai_team}).
\end{itemize}

To the best of our knowledge, our work is the first to demonstrate the utility of correspondence-based explanations to users on ImageNet \cite{russakovsky2015imagenet} and CUB \cite{wah2011caltech} classification tasks.

\section{Methods}
\label{sec:methods}

\subsection{Datasets}
\label{method:datasets}

We test our ImageNet classifiers on the original 50K-image ILSVRC 2012 ImageNet validation set (i.e., in-distribution data) and four common OOD benchmarks below.


\textbf{ImageNet-R} \cite{hendrycks2021many} contains 30K images in 200 ImageNet categories, mostly artworks -- ranging from cartoons to video-game renditions.

\textbf{ImageNet-Sketch} \cite{wang2019learning} consists of 50,889 black-and-white sketches of all 1,000 ImageNet classes.

\textbf{DAmageNet} \cite{chen2020universal} consists of 50K ImageNet validation-set images that contain universal, adversarial perturbations for fooling classifiers.

\textbf{Adversarial Patch} \cite{brown2017adversarial} 
are 50K ImageNet validation-set images that are modified to contain an adversarial patch that aims to cause ResNet-50 \cite{he2016deep} into labeling every image \class{toaster} (see Fig.~\ref{fig:adversarial_hen}).
Using the implementation by \cite{adv_patch_code}, we generate this dataset, which causes ResNet-50 accuracy to drop from 76.13\% to 55.04\% (\cref{tab:mainresults}).
See \cref{supp:generating_adversarial_patches} for how to download and generate this dataset.




\textbf{CUB-200-2011} \cite{wah2011caltech} (hereafter, CUB) is a fine-grained, bird-image classification task chosen to complement ImageNet.
CUB contains 11,788 images (5,994/5,794 for train/test) of 200 bird species.


\subsection{Classifiers}
\label{sec:method_classifiers}

We harness the same ResNet-50 \layer{layer4} backbone \cite{pretraineds2022pytorch} as the main feature extractor for all four main classifiers, including our two interpretable models. 
Therefore, to test the effectiveness of our models, we compare them with (1) a vanilla ResNet-50 classifier; and (2) a kNN classifier that uses the same pretrained \layer{layer4} features.
We report the top-1 accuracy of all classifiers in Table~\ref{tab:mainresults}.






\subsec{ResNet-50}
For experiments on ImageNet and its four OOD benchmarks, we use the ImageNet-trained ResNet-50 from TorchVision  \cite{pretraineds2022pytorch} (top-1 accuracy: 76.13\%).

For CUB, we take the ResNet-50 pretrained on iNaturalist \cite{van2018inaturalist} from \cite{nauta2021neural} (hereafter, iNaturalist ResNet) and retrain only the last 200-output classification layer (right after \layer{avgpool}) to create a competitive, baseline ResNet-50 classifier for CUB (top-1 accuracy: $85.83$\%).
See \cref{sec:app_resnet_inat} for finetuning details.

\subsec{kNN}
We implement a vanilla kNN classifier that operates at the \layer{avgpool} of the last convolutional layer of ResNet-50.
That is, given a query image $Q$, we sort all training-set images $\{ G_i \}$ based on their distance $D (Q, G_i)$, which is the cosine distance between the two corresponding image features $f(Q)$ and $f(G_i) \in \sR^{2048}$ where $f(.)$ outputs the \layer{avgpool} feature of \layer{layer4} (see \href{https://github.com/pytorch/vision/blob/main/torchvision/models/resnet.py#L205}{code}) of ResNet-50.

The predicted label of $Q$ is the dominant class among the top-$k$ nearest neighbors. 
We choose $k$ = 20 as it performs the best among the tested values of $k \in  \{ 10, 20, 50, 100 \}$.

\begin{figure*}[!hbt]
        \centering
        \begin{subfigure}{1.0\textwidth}
            \centering
            \includegraphics[width=0.70\linewidth]{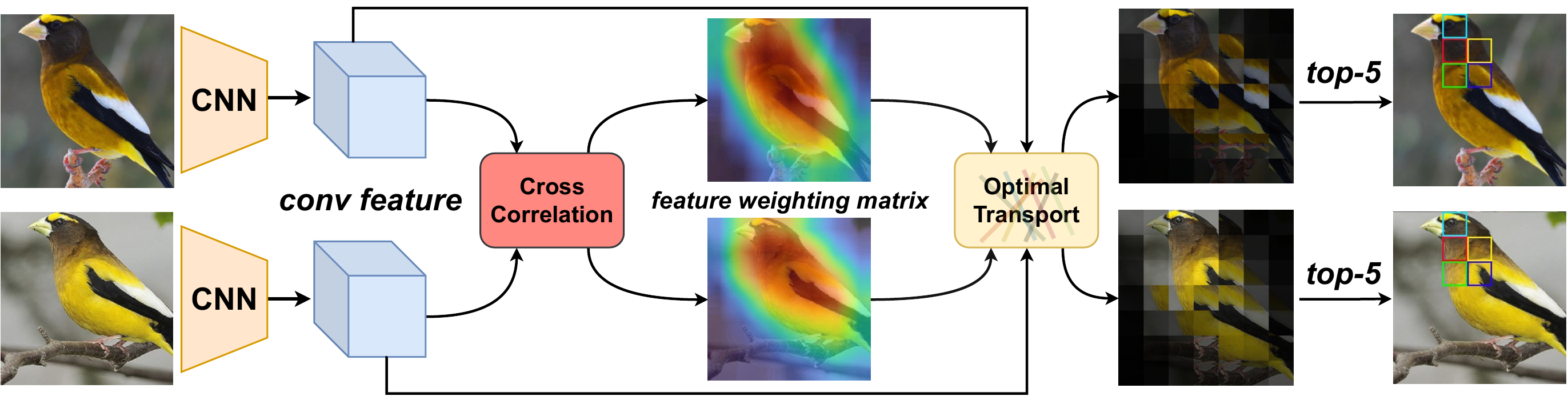}
                \caption{EMD-Corr: First compute patch-wise similarity, and then find correspondences via solving EMD \cite{hai2022deepface,zhao2021towards}.}
                \vspace*{0.3cm} 
                \label{fig:emd_based_class}
        \end{subfigure}
        \hfill
        \begin{subfigure}{1.0\textwidth}
            \centering
            \includegraphics[width=0.83\linewidth]{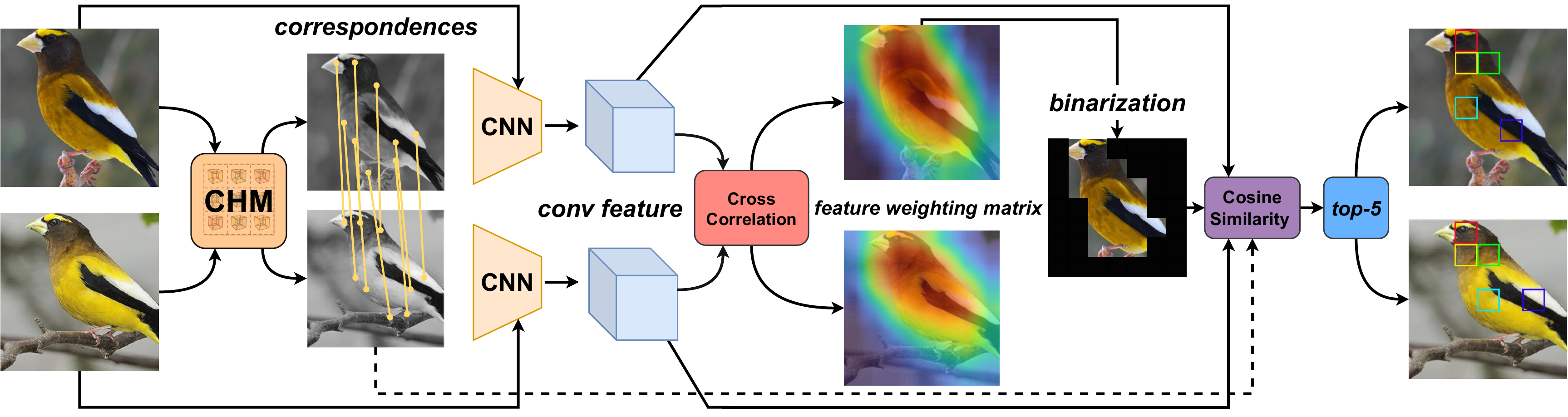}
                \caption{CHM-Corr: First find correspondences via CHM \cite{min2021convolutional}, and then compute patch-wise similarity.}
                \label{fig:chm_based_class}
        \end{subfigure}
        \caption{
        EMD-Corr and CHM-Corr both re-rank kNN's top-50 candidates using the patch-wise similarity between the query and each candidate over the top-5 pairs of patches that are the most important and the most similar (\ie highest EMD flows in EMD-Corr and highest cosine similarity in CHM-Corr).
        }
        \label{fig:classifiers}
\end{figure*}

\subsec{EMD-Corr} 
As kNN compares images using only image-level features, it lacks the capability of paying attention to fine details in images.
Therefore, we propose EMD-Corr, a visual correspondence-based classifier that (1) re-ranks the top-$N$ (here, $N=50$) candidates of kNN using their Earth Mover's Distance (EMD) with the query in a patch embedding space (see Fig.~\ref{fig:emd_based_class});
and (2), similarly to kNN, takes the dominant class among the re-ranked top-20 as the predicted label.

That is, our PyTorch implementation is the same as that in \cite{zhao2021towards,hai2022deepface} except for three key differences.
First, using \layer{layer4} features (7$\times$7$\times$2048), we divide an image into 49 patches, whose embeddings are $\in \sR^{2048}$.
Second, for interpretability, in re-ranking, we only use patch-wise EMD instead of a linear combination of image-level cosine distance and patch-level EMD as in \cite{zhao2021towards,hai2022deepface}, which makes it more opaque how a specific image patch contributes to re-ranking.
Third, while the original deep patch-wise EMD \cite{zhao2021towards,hai2022deepface} between two images (see Eq.~\ref{eq:emd_dist}) is defined as the sum over the \emph{weighted} cosine distances of all $49\times49$ patch pairs, we only use $L = 5$ pairs as explanations and therefore, only sum over the corresponding 5 flow weights returned by Sinkhorn optimization \cite{cuturi2013sinkhorn}.

We find $N=50$ to perform the best among $N \in \{50, 100, 200\}$.
We choose $L=5$, which is also the most common in nearest-neighbor visualizations \cite{li2022structure,krizhevsky2012imagenet}.
In preliminary experiments, we find $L = \{9, 16, 25\}$ to yield so dense correspondence visualizations that hurt user interpretation and $L = 3$ to under-inform users. See \cref{appendix:method_classifier3} for more description of EMD-Corr.

\subsec{CHM-Corr}
EMD-Corr first measures the pair-wise cosine distances for all 49$\times$49 pairs of patches, and then computes the EMD flow weights for these pairs (Fig.~\ref{fig:emd_based_class}).
To leverage the recent impressive end-to-end correspondence methods \cite{min2021convolutional,rocco2018neighbourhood,li2020correspondence}, we also propose CHM-Corr (Fig.~\ref{fig:chm_based_class}), a visual correspondence-based classifier that operates in the opposite manner to EMD-Corr. 
That is, first, we divide the query image $Q$ into 7$\times$7 non-overlapping patches (\ie as in EMD-Corr) and find one corresponding patch in $G_i$ for each of the 49 patches of $Q$ using a state-of-the-art correspondence method (here, CHM \cite{min2021convolutional}).
Second, for the query $Q$, we generate a cross-correlation (CC) map (\cref{fig:classifiers}) \cite{hai2022deepface}, \ie a heatmap of cosine similarity scores between the \layer{layer4} embeddings of the patches of the query $Q$ and the image-embedding (\layer{avgpool} after \layer{layer4}) of each training-set image $G_i$.
Third, we binarize the heatmap (using the optimal threshold $T = 0.55$ found on a held-out training subset) to identify a set of the most important patches in $Q$ and compute the cosine similarity between each such patch and the corresponding patch in $G_i$ (i.e., following the CHM correspondence mappings).
Finally, the similarity score $D(Q, G_i)$ in CHM-Corr is the sum over the $L=5$ patch pairs of the highest cosine similarity across $Q$ and $G_i$.


After testing NC-Net \cite{rocco2018neighbourhood}, ANC-Net \cite{li2020correspondence}, and CHM \cite{min2021convolutional} in our classifier, we choose CHM as it has the fastest runtime and the best accuracy.
Unlike ResNet-50 \cite{he2016deep}, which operates at the 224$\times$224 resolution, CHM uses a ResNet-101 backbone that expects a pair of 240$\times$240 images.
Therefore, in pre-processing, we resize and center-crop each original ImageNet sample differently according to the input sizes of ResNet-50 and CHM.
See \cref{supp:chm_details} for more description of CHM-Corr.








\subsec{CHM-Corr+ classifier based on five groundtruth keypoints of birds}
In EMD-Corr and CHM-Corr, we use CC to infer the importance weights of patches.
To understand the effectiveness of CC in weighting patches, on CUB, we compare our EMD-Corr and CHM-Corr to CHM-Corr+, a CHM-Corr variant where we use a set of five human-defined important patches instead of those inferred by CC.
That is, for each CUB image, instead of taking the five CC-derived important patches (\cref{fig:chm_based_class}), we use at most five patches that correspond to a set of five pre-defined keypoints (beak, neck, right wing, right feet, tail), each representing a common body part according to bird identification guides \cite{feng2018fine} for ornithologists.
From the five patches in the query image, we then use CHM to find five corresponding patches in a training-set image, and take the sum of five cosine similarities as the total patch-wise similarity between two images in re-ranking.

A query image may have $< 5$ important patches if some keypoint is occluded.
That is, evaluating CHM-Corr+ alone provides an estimate of how hard bird identification on CUB is if the model harnesses five well-known bird features.



\subsection{User-study design}

\begin{figure*}[t]
  \centering
        \includegraphics[width=0.9\textwidth]{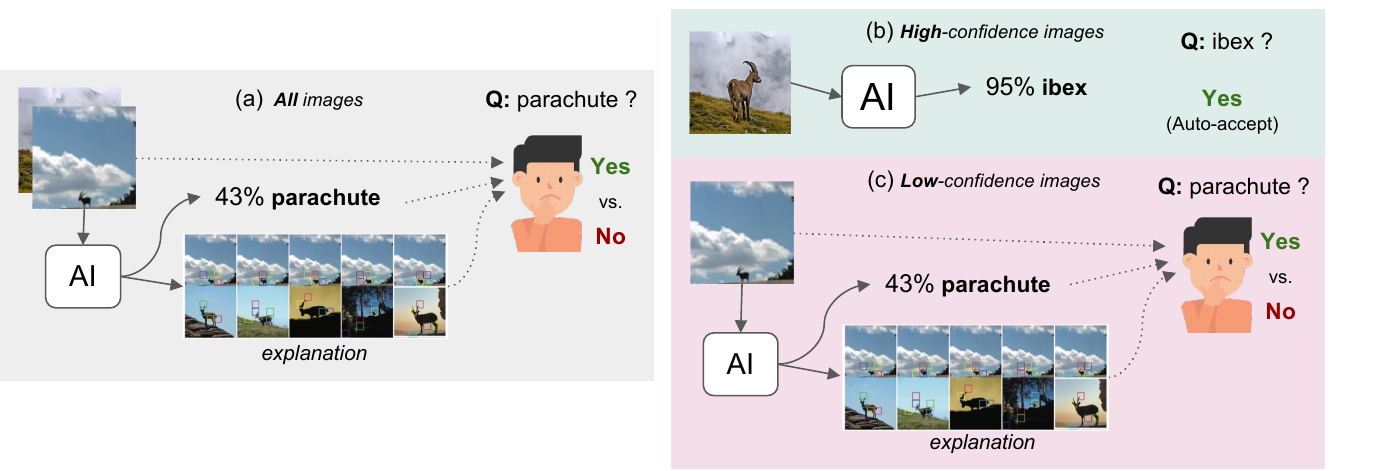}
	{   \small
		\begin{flushleft}
    		\hspace{1.7cm}
            Model 1: Human-only 
            \hspace{2.5cm}
            Model 2: Human-AI team decision makers
		\end{flushleft}
	}
    \caption{
    Two human-AI interaction models.
    In model 1 \colorbox{gray!15}{(a)}, for all images, users decide (Yes/No) whether the AI's predicted label is correct given the input image, AI top-1 label and confidence score, and explanations.
    In model 2, AI decisions are \emph{automatically} accepted if AI is highly confident \colorbox{MyLightGreen}{(b)}.
    Otherwise, humans will make decisions \colorbox{MyLightPink}{(c)} in the same fashion as the model 1.
    }
    \label{fig:interaction-model}
    \vspace*{-0.5cm}
\end{figure*}

The interpretable classifiers (\cref{sec:method_classifiers}) are not only capable of classifying images but also \emph{generating explanations}, which may inform and improve users' decision-making \cite{nguyen2021effectiveness}.
Here, we design a large-scale study to understand the effectiveness of explanations in \textbf{two human-AI interaction models} in classification (see \cref{fig:interaction-model}): 
\textbf{Model 1:} Users make all the decisions after observing the input, AI decisions, and explanations (\cref{fig:interaction-model}a).
\textbf{Model 2:} AIs make decisions on only inputs that they are the most confident, leaving the rest for users to label. 
That is, model 2 (\cref{fig:interaction-model}b--c) is a practical scenario where we offload most inputs to AIs while users only handle the harder cases.

    

Like \cite{nguyen2021effectiveness}, we show each user: (1) a query image; (2) AI top-1 label and confidence score; and (3) an explanation (here, available in kNN, EMD-Corr, CHM-Corr, and CHM-Corr+, but not in ResNet-50).
We ask users to decide Y/N whether the top-1 label is correct (example screen in \cref{fig:sample_trial_screen}).

\subsubsection{Explanation methods}
\label{sec:method_explanations}



We test the explanations of four main classifiers: ResNet-50, kNN, EMD-Corr, and CHM-Corr.
Additionally, we test two \emph{ablated} versions (\ie, EMD-NN and CHM-NN) of the explanations of EMD-Corr and CHM-Corr.
In total, we test 6 explanation methods (see examples in \cref{supp:sample_xais}).

\subsec{ResNet-50} is a representative black-box classifier, which only outputs a top-1 label and a confidence score (\ie, \emph{no explanations}).


\subsec{kNN explanations} 
From the top-20 nearest neighbors (as $k$ = 20 in our kNN), we show the first five images that are from the predicted class (example in \cref{fig:ibex}a).
In some cases where the predicted class has only $M < 5$ exemplars in the top-20, we still show only those $M$ images (see \cref{fig:imagenet_sketch_sample_2}).
That is, we only show at most five neighbors following prior works \cite{nguyen2021effectiveness,singla2014near,mac2018teaching,shen2020useful} that reported utility of such few-image explanations.
We find explanations consisting of $\geq$ 10 images such as those of ProtoPNet \cite{chen2019looks} are hard to interpret for users \cite{kim2021hive}.
Note that our kNN explanations consist of five support images for each decision of kNN (described in \cref{sec:method_classifiers}) as opposed to the \emph{post-hoc} nearest examples in \cite{nguyen2021effectiveness}, which do \emph{not} reflect a classifier's decisions.


\subsec{EMD-Corr and CHM-Corr explanations} 
As EMD-Corr and CHM-Corr re-rank the top-50 candidates (shortlisted by kNN) and take the dominant class among the resultant top-20 as the predicted label, we show the five nearest neighbors from the predicted class as in kNN explanations.
Instead of showing only five post-reranking neighbors, we also annotate, in each image, \emph{all} five patches (example in \cref{fig:ibex}b and \cref{fig:adversarial_hen}b) that contribute to the patch-wise re-ranking (\cref{sec:method_classifiers}).



\subsec{EMD-NN and CHM-NN} To understand the effects of showing correspondences in the explanations to EMD-Corr and CHM-Corr users, we also test an ablated version where we show exactly the same explanations but without the patch annotations (bottom panels in \cref{fig:ibex}a and \cref{fig:adversarial_hen}a).

\subsec{Confidence scores}
While ResNet-50's confidence score is the top-1 output softmax probability, the confidence of kNN, EMD-Corr, and CHM-Corr is the count of the predicted-class examples among the top $k$ = 20.
In human studies, we display this confidence in percentage (\eg 10\% instead of 2/20; \cref{fig:imagenet_sketch_sample_2}) to be consistent with the confidence score of ResNet-50.

\subsubsection{ImageNet and CUB datasets}
\label{sec:method_images}

For XAI evaluation, we run two human studies, one on ImageNet and one on CUB.
For ImageNet, we use ImageNet-ReaL \cite{beyer2020we} labels in attempt to minimize the confounders of the human evaluation as ImageNet labels are sometimes misleading and inaccurate to users \cite{nguyen2021effectiveness}.

\subsec{Nearest-neighbor images} To generate the nearest-neighbor explanations for kNN, EMD-Corr, and CHM-Corr, we search for neighbors in the entire training set of ImageNet or CUB (no filtering).

\subsec{Query images} In attempt to ensure the quality of the query images that we ask users to label, from 50K-image ImageNet validation set, we discard images that: ({a}) do not have an ImageNet-ReaL \cite{beyer2020we} label;
({b}) are grayscale or low-resolution (\ie, either width or height $< 224$ px) as in \cite{nguyen2021effectiveness}; 
({c}) have duplicates in the ImageNet training set (see~\cref{supp:duplicates}), resulting in 44,424 images available for sampling for the study.
In CUB, we sample from the entire 5,794-image test set and apply no filters.

\subsubsection{Training, Validation, and Test phases}
\label{sec:method_tasks}

From the set of query images (\cref{sec:method_images}), we sample images for three phases in a user study: Training, validation, and test.
Following \cite{nguyen2021effectiveness}, we first introduce participants to the task and provide them 5 training examples.
Then, each user is given a validation job (10 trials for ImageNet and 5 for CUB), where they must score 100\% in order to be invited to our 30-trial test phase.
Otherwise, they will be rejected and unpaid.
Among the 10 validation trials for ImageNet, 5 are correctly-labeled and 5 are misclassified by AIs.
This ratio is 3/2 for CUB validation (examples in \cref{supp:validation_images}).




Right before each trial, we describe the AI's top-1 label to users by showing them 3 training-set images and a 1-sentence WordNet description for each ImageNet class.
For CUB classes, we show 6 representative images (instead of 3) for users to better recognize the characteristics of each bird (see \cref{fig:sample_images}).

\subsec{Sampling} 
For every classifier, we randomly sample 300 correctly- and 300 incorrectly-predicted images together with their corresponding explanations for the test trials.
Over all 6 explanation methods, we have 2 datasets $\times$ 600 images $\times$ 6 methods = 7,200 test images in total. 



\subsubsection{Participants}
\label{sec:method_participants}

We host human studies on Gorilla \cite{anwyl2020gorilla} and recruit lay participants who are native English speakers worldwide via Prolific \cite{palan2018prolific} at a pay rate of USD 13.5 / hr.
We have 360 and 355 users who successfully passed our validation test for ImageNet and CUB datasets, respectively. 
We remove low-quality, bottom-outlier submissions, \ie, who score $\leq 0.55$ (near-random accuracy), resulting in 354 and 355 submissions for ImageNet and CUB, respectively.
In each dataset, every explanation method is tested on $\sim$60 users 
and each pair of (query, explanation) is seen by almost 3 users (details in Table~\ref{table:humanaloneperformance}). 







\section{Experimental Results}
\label{sec:result}











\subsection{ImageNet kNN classifiers improve upon ResNet-50 on out-of-distribution datasets}
\label{sec:result_finding1}

Despite impressive test-set performance, ImageNet-trained convolutional neural networks (CNNs) may fail to generalize to natural OOD data \cite{wang2019learning} or inputs specifically crafted to fool them \cite{nguyen2015deep,chen2020universal,brown2017adversarial}.
It is unknown whether prototype-based classifiers can leverage the known exemplars (\ie support images) to generalize better to unseen, rare inputs.
To test this question, here, we compare kNN with the baseline ResNet-50 classifier (both described in \cref{sec:methods}) on ImageNet and related OOD datasets.




On ImageNet and ImageNet-ReaL, kNN performs slightly worse than ResNet-50 by  \decreasenoparent{1.36} and \decreasenoparent{0.99} points, respectively (\cref{tab:mainresults}).
Yet, interestingly, \textbf{on all four OOD datasets, kNN consistently outperforms} ResNet-50.
Notably, kNN improves upon ResNet-50 by \increasenoparent{1.66} and \increasenoparent{4.26} points on DAmageNet and Adversarial Patch.
That is, while ResNet-50 and kNN share the exact same backbone, the kNN's process of comparing the input image against the training-set examples prove to be beneficial for generalizing to OOD inputs.
Intuitively, our results suggest that it is useful to ``look back'' at the training-set exemplars to decide a label for hard, long-tail or OOD images.

Consistently, using the same CUB-finetuned backbone, kNN is only marginally worse than ResNet-50 on CUB (85.46\% vs. 85.83\%; \cref{tab:mainresults}).

\begin{table}
\centering
\small
\caption{Top-1 accuracy (\%).
\resnet 
~models' classification layer is fine-tuned on a specified training set in (b).
All other classifiers are non-parametric, nearest-neighbor models based on pretrained ResNet-50 features (a) and retrieve neighbors from the training set (b) during testing.
\emdcorr~\& \chmcorr~ outperform ResNet-50 models on all \colorbox{MyLightGray}{OOD datasets} (\eg \increasenoparent{4.39} on Adversarial Patch) and slightly underperform on in-distribution sets (\eg \decreasenoparent{0.72} on ImageNet-ReaL).
}
\label{tab:mainresults}
\setlength\tabcolsep{3pt} 
\resizebox{14cm}{!}{
\begin{tabular}{|l|c|c|r|r|l|l|c|}
\hline
Test set & Features (a) & Training set (b) & {\resnet} & {\knn} & {\emdcorr} & {\chmcorr} & {CHM-Corr+} \\ \hline
ImageNet \cite{russakovsky2015imagenet} & ImageNet & ImageNet                                                      & \textbf{76.13} & 74.77 & 74.93 \decrease{1.20} & 74.40 \decrease{1.73}  & \na      \\
ImageNet-ReaL \cite{beyer2020we} & ImageNet & ImageNet                                                      & \textbf{83.04} & 82.05 & 82.32 \decrease{0.72} & 81.97 \decrease{1.07}  & \na      \\ \hline
\rowcolor{MyLightGray}
ImageNet-R \cite{hendrycks2021many} & ImageNet & ImageNet                                      & 36.17 & 36.18 & \textbf{37.75} \increase{1.58} & 37.62 \increase{1.45} & \na       \\
\rowcolor{MyLightGray}
ImageNet Sketch \cite{wang2019learning} & ImageNet & ImageNet                            & 24.09 & 24.72 & 25.36 \increase{1.27} & \textbf{25.61} \increase{1.52} & \na \\
\rowcolor{MyLightGray}
DAmageNet \cite{chen2020universal} & ImageNet & ImageNet                                    & 5.93  & 7.59  & ~~\textbf{8.16} \increase{2.23}  & ~~8.10 \increase{2.17}   &  \na     \\
\rowcolor{MyLightGray}
Adversarial Patch \cite{brown2017adversarial} & ImageNet & ImageNet                               & 55.04 & 59.30  & 59.43 \increase{4.39} & \textbf{59.86} \increase{4.82} &  \na     \\
\rowcolor{MyLightGray}
CUB \cite{wah2011caltech} & ImageNet & CUB
& \na & 54.72 & \textbf{60.29} & 53.65 & 49.63 \\ \hline
CUB \cite{wah2011caltech} & iNaturalist \cite{van2018inaturalist} & CUB & \textbf{85.83} & 85.46 & 84.98 \decrease{0.85} & 83.27 \decrease{2.56} & 81.54 \\ \hline
\end{tabular}
} 
\vspace*{-0.3cm} 
\end{table}

\subsection{Visual correspondence-based explanations improve kNN robustness further}
\label{sec:result_finding2}

Recent work found that re-ranking kNN's shortlisted candidates using the patch-wise similarity between the query and training set examples can further improve classification accuracy on OOD data for some image matching tasks \cite{hai2022deepface,zhao2021towards,zhang2020deepemd} such as face identification \cite{hai2022deepface}.
Furthermore, patch-level comparison is also useful in prototype-based bird classifiers \cite{chen2019looks,donnelly2021deformable}.
Inspired by these prior successes and the fact that EMD-Corr and CHM-Corr base the patch-wise similarity of two images on only 5 patch pairs instead of all 49$\times$49 = 2,401 pairs as in \cite{hai2022deepface,zhao2021towards,zhang2020deepemd}, here we test whether our two proposed re-rankers are able to improve the test-set accuracy and robustness over kNN.

\subsec{Experiment}
We run EMD-Corr and CHM-Corr on all datasets and compare their results with that of kNN (\cref{tab:mainresults}).
Both methods (described in \cref{sec:methods}) re-rank the top $N$ = 50 shortlisted candidates returned by kNN and then take the dominant class in the top-$k$ (where $k$ = 20) as the predicted label.

\subsec{ImageNet results}
Interestingly, despite using only 5 pairs of patches to compute image similarity for re-ranking, both classifiers consistently improve upon kNN further, especially on all OOD datasets.
Overall, EMD-Corr and CHM-Corr outperform kNN and ResNet-50 baselines from \increasenoparent{1.27} to \increasenoparent{4.82} points (\cref{tab:mainresults}).
Intuitively, in some hard cases where the main object is small, the two Corr classifiers ignore irrelevant patches (\eg the sky in \class{ibex} images; \cref{fig:ibex}) and only use the five most relevant patches to make decisions.
Similarly, on Adversarial Patch, relying on a few key patches while ignoring adversarial patches enables our classifiers to outperform baselines (\cref{fig:adversarial_hen}).
See \cref{appendix:compare_methods} for many qualitative examples comparing Corr and kNN predictions.

\subsec{CUB results} Interestingly, using the same ImageNet-pretrained backbones, EMD-Corr outperforms kNN by an absolute \increasenoparent{5.57} points when tested on CUB (60.29\% vs. 54.72\%; \cref{tab:mainresults}).
However, this difference vanishes when using CUB-pretrained backbones (\cref{tab:mainresults}; 84.98\% vs. 85.46\%).


Our CUB and ImageNet results are consistent and together reveal a trend: On i.i.d test sets, Corr models perform on par with kNN; however, on OOD images, they consistently outperform kNN, highlighting the benefits of patch-wise comparison.

\subsection{Corr classifiers leverage five patches that are more important than five bird keypoints}
\label{sec:corr_plus}

EMD-Corr and CHM-Corr harness five patches per image for computing a patch-wise similarity score for a pair of images (\cref{sec:method_classifiers}).
As these five patches are automatically inferred by cross-correlation (\cref{fig:classifiers}), it is interesting to understand further whether replacing these five patches by five user-defined patches in \cite{wah2011caltech} would improve classification accuracy.

\subsec{Experiment}
Since there are no keypoints provided for ImageNet, we test the importance of the five key patches chosen by Corr methods on CUB because CUB provides ornithologist-defined annotations for each bird image.
That is, we create a baseline CHM-Corr+, which is the same as CHM-Corr, except that we use five important patches that correspond to five keypoints in a bird image---beak, belly, tail, right wing, and right foot---as described in \cref{sec:method_classifiers}.
We also test CHM-Corr+ sweeping across the number of keypoints $\in \{5, 10, 15\}$.

\subsec{Results}
On CUB, CHM-Corr outperforms CHM-Corr+ despite the fact that the baseline method leverages \textbf{five} human-defined bird keypoints (\cref{tab:mainresults}; 83.27\% vs. 81.51\%) while CHM-Corr may also use background patches.
Interestingly, when increasing the number of keypoints to 10 and 15, the accuracy of CHM-Corr+ is still lower than that of CHM-Corr (\ie, from 81.51\% to 82.34\% and 82.27\%, respectively).
That is, 15 keypoints may correspond to $\leq$ 15 different patches per image (15 if each keypoint lies in a unique, non-overlapping patch among all the 49 patches per image).

Our results show strong evidence that the five key patches inferred by CC used in EMD- and CHM-Corr do not necessarily cover the birds but are more important than expert-defined bird keypoints.
Qualitative comparisons between CHM-Corr and CHM-Corr+ predictions are in \cref{supp:comapre_method4_methdo4_plus}.

\subsection{On ImageNet-ReaL, correspondence-based explanations are more useful to users than kNN explanations}
\label{sec:result_finding3}

Given that EMD-Corr and CHM-Corr classifiers outperform kNN classifiers on OOD datasets (\cref{sec:result_finding2}), it is interesting to test how their respective explanations help humans perform classification on ImageNet.
Furthermore, in image classification, kNN explanations were found to be more useful to humans than saliency maps \cite{nguyen2021effectiveness}.

\begin{wrapfigure}[26]{r}{.28\textwidth}
\vspace{-0.5cm}
    \begin{minipage}{\linewidth}
    \centering\captionsetup[subfigure]{justification=centering}
    \includegraphics[width=\linewidth]{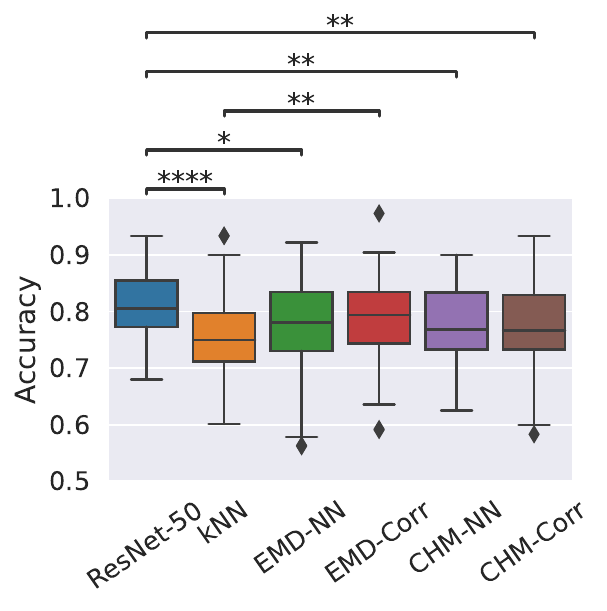}
    \subcaption{ImageNet}
    \label{fig:mannwhittney-utest-imagenet}\par\vfill
     \includegraphics[width=\linewidth]{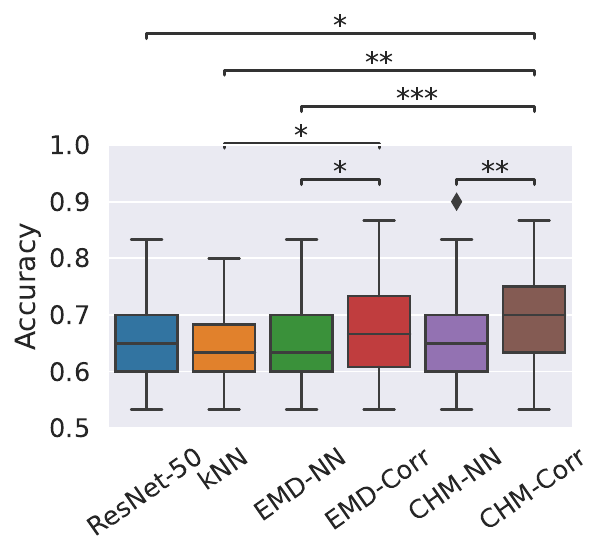}
    \subcaption{CUB}
    \label{fig:mannwhittney-utest-cub}
\end{minipage}
\caption{Mann-Whitney U test of the user accuracy scores of 6 methods.{\small
$^*$ = $p$-value < 0.05.
$^{**}$ = $p$-value < 0.01.
$^{***}$ = $p$-value < 0.001.
\vspace*{1cm}
}
}
\label{fig:mannwhittney-utest}
\end{wrapfigure}

\subsec{Experiment}
We perform a human study to assess the ImageNet-ReaL classification accuracy of \emph{users} of each classifier (described in \cref{sec:method_classifiers}) when they are provided with a classifier's predictions and explanations.
That is, we measure the AI-assisted classification accuracy of users following human-AI interaction model 1 (\cref{fig:interaction-model}a).
We compare the accuracy between user groups of four classifiers ResNet-50, kNN, EMD-Corr, and CHM-Corr (described in \cref{sec:method_explanations}).

Additionally, to thoroughly assess the impact of showing the correspondence boxes compared to showing only nearest neighbor images (\eg, CHM-Corr vs. CHM-NN in \cref{fig:ibex}), we test two more user groups of EMD-NN and CHM-NN, \ie the same explanations as those of the Corr classifiers but with the correspondence boxes \emph{hidden}.

\subsec{Results}
First, the mean accuracy of kNN users is consistently lower than that of the other models' users (\eg, 75.76\% vs. 78.87\% of EMD-Corr; \cref{table:humanaloneperformance}).
The EMD-Corr improvement over kNN is statistically significant ($p$ < 0.01 via Mann-Whitney U test; \cref{fig:mannwhittney-utest-imagenet})

Second, interestingly, we find the differences between EMD-, CHM-Corr and their respective baselines are small and not statistically significant (\cref{fig:mannwhittney-utest-imagenet}).
That is, on ImageNet-ReaL, quantitatively, showing the correspondence boxes on top of nearest neighbors is not more useful to users.
Third, surprisingly, the users of ResNet-50 (mean accuracy of 81.56\%; \cref{table:humanaloneperformance}) outperform all other methods' users, suggesting that on ImageNet, a task of many familiar classes to users, \emph{ante-hoc} explanations hurt user accuracy rather than help.
Note that in \citet{nguyen2021effectiveness}, post-hoc kNN explanations were found useful to humans compared to not showing any explanations.
Yet, here, each classifier's users are provided with a different set of images and AI decisions, which can also influence the user accuracy. 
When ResNet-50 is wrong, their users are substantially better in detecting such misclassifications compared to other models' users (\cref{fig:imagenet-ai-correctness}).

\subsection{On CUB fine-grained bird classification, correspondence-based explanations are the most useful to users, helping them to more accurately reject AI misclassifications}
\label{sec:result_finding4}

To assess whether the findings on ImageNet-ReaL in \cref{sec:result_finding3} generalize to a fine-grained classification task, we repeat the user study on CUB---which is considered much harder to lay users than ImageNet.

\subsec{Results} Interestingly, we find EMD-Corr and CHM-Corr users consistently outperform ResNet-50, kNN, EMD-NN, and CHM-NN users (\cref{table:humanaloneperformance}).
The differences between EMD-Corr (or CHM-Corr) and every other baseline are statistically significant ($p$ < 0.05 via Mann-Whitney U test; \cref{fig:mannwhittney-utest-cub}).
That is, on CUB, the visual correspondence boxes help users make more accurate decisions compared to (a) having no explanations at all (ResNet-50); (b) showing nearest neighbors sorted by image similarity only, not patch correspondences (\cref{fig:ibex}a; kNN); and (c) having patch-wise correspondence neighbors but not displaying the boxes (\cref{fig:placeholder_ibex}; CHM-NN and EMD-NN).

\begin{table}[t]
\small
\begin{minipage}{.495\linewidth}
\centering

\caption{\small{Human-only accuracy (\%)}}
\setlength\tabcolsep{4pt}
\resizebox{\columnwidth}{!}{%
\begin{tabular}{|l|cc|cc|}
\hline
\multicolumn{1}{|c|}{\multirow{2}{*}{{Method}}} & \multicolumn{2}{c|}{{ImageNet-ReaL}} & \multicolumn{2}{c|}{{CUB}} \\ \cline{2-5} 
\multicolumn{1}{|c|}{} & \multicolumn{1}{c|}{{Users}} & {Accuracy} & \multicolumn{1}{c|}{{Users}} & {Accuracy} \\ \hline
\rowcolor{ResNetBlue!7} \resnet & \multicolumn{1}{c|}{60} & \textbf{81.56} $\pm$ 5.54 & \multicolumn{1}{c|}{60} & 65.50 $\pm$ 7.46 \\ \hline
\rowcolor{kNNOrange!7} \knn & \multicolumn{1}{c|}{59} & 75.76 $\pm$ 8.55 & \multicolumn{1}{c|}{59} & 64.75 $\pm$ 7.14 \\ \hline
\rowcolor{EMDRed!7} \emdcorr & \multicolumn{1}{c|}{59} & \textbf{78.87} $\pm$ 6.57 & \multicolumn{1}{c|}{58} & \textbf{67.64} $\pm$ 7.44 \\ \hline
\rowcolor{CHMBrown!7} \chmcorr & \multicolumn{1}{c|}{59} & 77.23 $\pm$ 7.56 & \multicolumn{1}{c|}{59} & \textbf{69.72} $\pm$ 9.08 \\ \hline
EMD-NN & \multicolumn{1}{c|}{57} & 77.72 $\pm$ 8.27 & \multicolumn{1}{c|}{59} & 64.12 $\pm$ 7.07 \\ \hline
CHM-NN & \multicolumn{1}{c|}{60} & 77.56 $\pm$ 6.91 & \multicolumn{1}{c|}{60} & 65.72 $\pm$ 8.14 \\ \hline
\end{tabular}%
}
\label{table:humanaloneperformance}
\end{minipage}     
\begin{minipage}{.495\linewidth}
  \caption{\small{AI-only and Human-AI team accuracy (\%)}}
  \centering

\setlength\tabcolsep{2pt}
\resizebox{\columnwidth}{!}{%
\begin{tabular}{|l|cl|cl|}
\hline
\multicolumn{1}{|c|}{\multirow{2}{*}{{Method}}} & \multicolumn{2}{c|}{{ImageNet-ReaL}} & \multicolumn{2}{c|}{{CUB}} \\ \cline{2-5} 
\multicolumn{1}{|c|}{} & \multicolumn{1}{c|}{\small{{AI-only}}} & \multicolumn{1}{c|}{{\begin{tabular}[c]{@{}c@{}}Human-AI\\ \end{tabular}}} & \multicolumn{1}{c|}{\small{{AI-only}}} & \multicolumn{1}{c|}{{\begin{tabular}[c]{@{}c@{}}Human-AI\\ \end{tabular}}} \\ \hline
\rowcolor{ResNetBlue!7} \resnet & \multicolumn{1}{c|}{86.11} & 88.63	 \increase{2.52} & \multicolumn{1}{c|}{87.38} & 87.45 \increase{0.07} \\ \hline
\rowcolor{kNNOrange!7} \knn & \multicolumn{1}{c|}{85.95} & 87.24 \increase{1.29} & \multicolumn{1}{c|}{87.40} & 86.66 \decrease{0.74} \\ \hline
\rowcolor{EMDRed!7} \emdcorr & \multicolumn{1}{c|}{85.91} & 88.02 \increase{2.11} & \multicolumn{1}{c|}{86.88} & 86.86 \decrease{0.02} \\ \hline
\rowcolor{CHMBrown!7} \chmcorr & \multicolumn{1}{c|}{85.36} & 87.89 \increase{2.53} & \multicolumn{1}{c|}{85.48} & 86.25 \increase{0.77} \\ \hline \hline
\hfill \emph{mean} & \multicolumn{1}{c|}{85.83} & 87.94 \increase{2.11} & \multicolumn{1}{c|}{86.78	} & 86.80 \increase{0.02} \\ \hline
\end{tabular}%
}
\vspace{0.25cm} 
\label{table:aialone_and_humanteam}
\end{minipage}%
\end{table}

\paragraph{Corr explanations help users reject AI misclassifications while kNN is poorly trust-calibrated}
In an attempt to understand why the two Corr classifiers help users the most, we find that EMD-Corr and CHM-Corr users reject AI predictions at the highest rates (32.70\% and 33.73\%; \cref{tab:users-accept-reject-data}) while kNN users reject the least (18.47\%).

\begin{figure*}[h]
  \centering
    \includegraphics[width=1.0\textwidth]{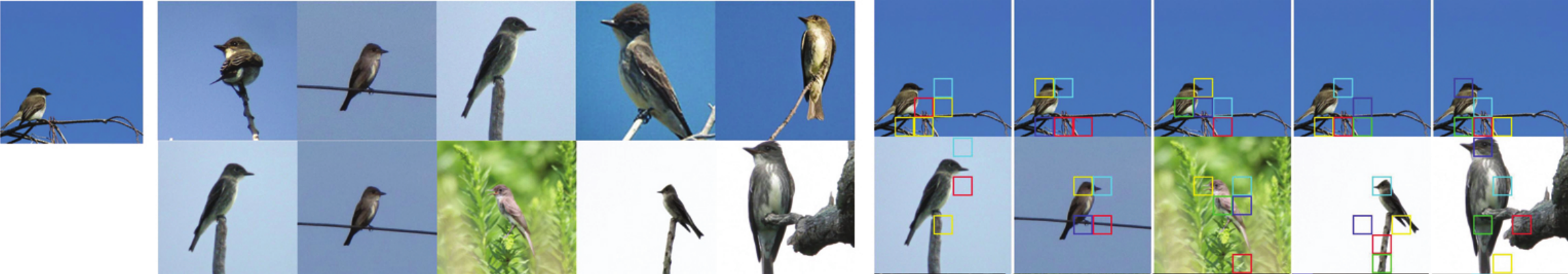}
    {
    \tiny	
		\begin{flushleft}
		\vspace{-1.3cm}
			groundtruth:\\
			\class{Sayornis}
		\end{flushleft}
	}
	\vspace{1cm}
	{\vspace{-0.6cm}
    \tiny	
		\begin{flushleft}
    		\hspace{2.3cm}
            (a) kNN (top) and CHM-NN (bottom) explanation
            \hspace{1.3cm}
            (b) CHM-Corr visual correspondence-based explanations
		\end{flushleft}
	}
    \caption{A \class{Sayornis} bird image is \textbf{misclassified} into \class{Olive Sided Flycatcher} by both kNN and CHM-Corr models. 
    Yet, all 3/3 CHM-Corr users \textcolor{OliveGreen}{correctly rejected} the AI prediction while 4/4 kNN users \textcolor{red}{wrongly accepted}. 
    CHM-Corr explanations (b) show users more diverse samples and more evidence against the AI's decision.
    More similar examples are in \cref{supp:when_xai_helps_group1}.}
    \label{fig:placeholder_bird}
\end{figure*}

This might have led to the substantially higher accuracy of CHM-Corr users, compared to all other models' user groups, when \emph{AI predictions are wrong} (\cref{fig:cub-ai-correctness}; \eg, 53.45\% of CHM-Corr vs. 41.22\% of ResNet-50).
That is, CHM-Corr users correctly reject 53.45\% of the images that the CHM-Corr classifier mislabels.
In contrast, kNN users reject the least, only 33.22\% of incorrect predictions (\cref{fig:cub-ai-correctness}). 
kNN explanations tend to \emph{fool} users into trusting the kNN's wrong decisions (\cref{fig:placeholder_bird})---the accuracy of kNN users is 33.22\%, much lower than the 41.22\% of ResNet-50 users who observe no explanations.
On ImageNet (\cref{fig:breakdown-ai-correctness}), kNN is also poorly ``trust-calibrated'' \cite{turnercalibrating,yang2017evaluating}.

We hypothesize that kNN explanations tend to fool users more as their nearest neighbors, by design, show images that are \emph{image-wise} similar to the query (regardless of whether the kNN prediction is correct or not) while EMD-Corr and CHM-Corr re-rank the images based on \emph{patch-wise} similarity.
Furthermore, we find the \textbf{images in kNN explanations are also less diverse} than those in Corr explanations in both LPIPS \cite{zhang2018unreasonable} and MS-SSIM \cite{wang2003multiscale} (\cref{sec:diversity_test}).
Corr explanations tend to include more diverse images (\cref{fig:placeholder_bird}a; top vs. bottom) and provide users with more contrastive evidence in order to reject AI's incorrect predictions.

Additionally, we also hypothesize that users are less confident about AI's decisions (and thus reject more) when Corr explanations show some background and uninformative patches used in the matching process (\eg, the 1st and 4th image in \cref{fig:placeholder_bird}b).
Yet, such boxes are not available in kNN explanations.

\textbf{When Corr explanations allow for more disagreement between AI and users, humans also tend to incorrectly reject AI's correct predictions more often} (\cref{fig:cub-ai-correctness}; Corr users are the least accurate among 6 methods when the AI is correct).
EMD-Corr and CHM-Corr users score $\sim$4 points below ResNet-50 users (84.94\% and 85.99\% vs. 89.87\% of ResNet-50).
When most users reject AI's correct predictions, we observe that some discriminative features (\eg, the belly stripes of the Field Sparrow;
Fig.~\ref{fig:group2_c}) are often occluded in the query, leading to human-AI disagreement.

\section{Related Work}

\subsec{Patch-wise similarity}
Calculating patch-wise similarity, either intra-image \cite{dosovitskiy2021an} or inter-image \cite{hai2022deepface,zhao2021towards,zhang2020deepemd}, has been useful in many tasks as the comparison enables machines to attend to fine-grained details and compute more accurate decisions.
Our EMD-Corr harnesses a similar approach to that in \cite{hai2022deepface,zhao2021towards}; which, however, was not tested on ImageNet classification as in our work.
Furthermore, we only compute the total patch-wise similarity over the top-5 patch pairs between the query and each exemplar instead of all pairs as in \cite{hai2022deepface,zhao2021towards}.
Compared to recent patch-wise similarity works that use either cross-attention in ViTs \cite{dosovitskiy2021an} or EMD \cite{zhao2021towards,zhang2020deepemd,hai2022deepface,kim2021vilt}, our work is the first to perform human evaluation of the correspondence-based explanations.

\subsec{Prototype-based XAI methods}
Our work is motivated by the recent finding that exemplar-based explanations are more effective than heatmap-based explanations in improving human classification accuracy \cite{nguyen2021effectiveness,jeyakumar2020can,kim2021hive,fel2021cannot}.
However, showing an entire image as an exemplar without any further localization may be confusing as it is unknown which parts of the image the AI is paying attention to \cite{nguyen2021effectiveness,jeyakumar2020can}.
Our EMD-Corr and CHM-Corr present a novel combination of heatmap-based and prototype-based XAI approaches.
None of the prior prototype-based XAI methods that operate at the patch level \cite{chen2019looks,nauta2021neural,zhao2021towards,donnelly2021deformable} (see Table 3 in \cite{donnelly2021deformable}) were tested on humans yet.
Also, in preliminary tests, we find their explanation formats too dense (i.e., showing over 10 prototypes \cite{chen2019looks}, 9 correspondence pairs per image \cite{donnelly2021deformable}, or an entire prototype tree to humans \cite{nauta2021looks,nauta2021neural}) to be useful for lay users.

Another major difference is that our Corr classifiers are nonparametric, allowing the training set to be adjusted or swapped with any external knowledgebase for debugging purposes.
In contrast, recent prototype-based classifiers \cite{chen2019looks,nauta2021neural,zhao2021towards,donnelly2021deformable} are parametric, using a set of learned prototypes and thus may not perform well on OOD datasets as EMD-Corr and CHM-Corr.


\subsec{Post-hoc prototype-based explanations} 
Some prototype-based methods are post-hoc \cite{kenny2019twin,nguyen2021effectiveness, crabbe2021explaining}, i.e., generating explanations to explain a decision after-the-fact, which could be highly unfaithful \cite{rudin2019stop,rudin2021interpretable}.
Instead, our approach is inherently interpretable \cite{rudin2021interpretable}, i.e., retrieving the patches first, and then using them to make classification decisions.
While our binary classification task is adopted from \cite{nguyen2021effectiveness}, our study compares 4 different classifiers while \citet{nguyen2021effectiveness} instead tested a single classifier with multiple post-hoc explanations.

\subsec{Human studies}
Our study has 709 users in total, \ie $\sim$60 users per method per dataset, which is substantially larger than that in most prior works.
That is, $\sim$30 and 40 users per method participated in \cite{nguyen2021effectiveness} and in \cite{mac2018teaching}, respectively while \citet{adebayo2020debugging} had 54 persons in total for the entire study of multiple methods.

\subsec{Human-AI teaming}
Human-AI collaboration is becoming more essential in the modern AI era \cite{hemmer2021human}. 
A large body of prior works has investigated such collaboration in other domains (e.g., NLP \cite{bansal2021does,zhang2022you}, healthcare \cite{caruana2015intelligible} and others \cite{hemmer2022effect, caruana2015intelligible, zhang2020effect, chu2020visual}); however, only few works investigated human-AI collaboration in the image classification setting \cite{nguyen2021effectiveness,kim2021hive,fel2021cannot}.

Some prior works predict when to defer the decision-making to humans \cite{raghu2019algorithmic, horvitz2007complementary, kamar2012combining}.
However, by simply offloading some inputs to humans based on confidence scores, we achieve complementary human-AI team performance in both ImageNet and CUB.
Previous works \cite{bansal2021does, schemmer2022should} found that algorithmic explanations benefit human decision-making in general, but did not find XAI methods to yield team complementary performance \cite{hemmer2021human}, which we report in this work.





\section{Discussion and Conclusion}
\label{sec:discussion}

\subsec{Limitations} Due to the limited amount of time and expensive cost of computation related to EMD-Corr or CHM-Corr, we did not experiment on a wider range of OOD datasets (e.g., adversarial poses \cite{alcorn2019strike}).
We tested our methods on ImageNet-A \cite{hendrycks2021natural}, ObjectNet \cite{barbu2019objectnet}, and ImageNet-C \cite{hendrycks2019benchmarking} as well, but on a small scale of 5K-image sets (see Table~\ref{tab:ablationstudy-otherdatasets}).
As using online crowdworkers for the XAI human evaluation, we share the same limitations with \cite{nguyen2021effectiveness,adebayo2020debugging,kim2021hive}.
That is, despite our best efforts to minimize biases, the human data quality can be improved in highly-controlled laboratory conditions like in \cite{geirhos2021partial}.
Algorithm-wise, EMD-Corr and CHM-Corr are re-ranking methods and therefore run substantially slower than ResNet-50 (see \cref{fig:time-performance-supp} for a speed comparison of all models).


Our work is the first attempt to: (1) study the effectiveness of patch-wise comparison in improving the robustness of deep image classifiers on ImageNet OOD benchmarks; 
(2) show the utility of visual correspondence-based explanations in helping users make more accurate image-classification decisions; 
(3) achieve human-AI complementary team performance in the image domain.

\subsubsection*{Acknowledgement}
The authors would like to thank Ken Stanley for the great idea of using correspondences as an explanation for image classification, leading to this work.
We also thank Thang Pham, Peijie Chen, and Hai Phan for feedback and discussions of the earlier results.
AN was supported by the NSF Grant No. 1850117 \& 2145767, and donations from NaphCare Foundation \& Adobe Research.

\newpage
{\small
\bibliographystyle{icml2021}
\bibliography{references.bib}
}

\clearpage
\newpage

\section*{Checklist}

The checklist follows the references.  Please
read the checklist guidelines carefully for information on how to answer these
questions.  For each question, change the default \answerTODO{} to \answerYes{},
\answerNo{}, or \answerNA{}.  You are strongly encouraged to include a {\bf
justification to your answer}, either by referencing the appropriate section of
your paper or providing a brief inline description.  For example:
\begin{itemize}
  \item Did you include the license to the code and datasets? \answerYes{See Section~\ref{gen_inst}.}
  \item Did you include the license to the code and datasets? \answerNo{The code and the data are proprietary.}
  \item Did you include the license to the code and datasets? \answerNA{}
\end{itemize}
Please do not modify the questions and only use the provided macros for your
answers.  Note that the Checklist section does not count towards the page
limit.  In your paper, please delete this instructions block and only keep the
Checklist section heading above along with the questions/answers below.

\begin{enumerate}

\item For all authors...
\begin{enumerate}
  \item Do the main claims made in the abstract and introduction accurately reflect the paper's contributions and scope?
    \textbf{\answerYes. Please see \cref{sec:result}}.
  \item Did you describe the limitations of your work?
    \textbf{\answerYes. Please see \cref{sec:discussion}}.
  \item Did you discuss any potential negative societal impacts of your work?
    \textbf{\answerNA}.
  \item Have you read the ethics review guidelines and ensured that your paper conforms to them?
    \textbf{\answerYes}.
\end{enumerate}

\item If you are including theoretical results...
\begin{enumerate}
  \item Did you state the full set of assumptions of all theoretical results?
    \textbf{\answerYes}.
        \item Did you include complete proofs of all theoretical results?
    \textbf{\answerNA}.
\end{enumerate}

\item If you ran experiments...
\begin{enumerate}
  \item Did you include the code, data, and instructions needed to reproduce the main experimental results (either in the supplemental material or as a URL)?
    \textbf{\answerNo}.
  \item Did you specify all the training details (e.g., data splits, hyperparameters, how they were chosen)?
    \textbf{\answerYes. Please see Sec.~\ref{sec:methods}}.
        \item Did you report error bars (e.g., with respect to the random seed after running experiments multiple times)?
    \textbf{\answerYes. We reported whenever possible. See $\mu$ and $\sigma$ in Table~\ref{table:humanaloneperformance}}.
        \item Did you include the total amount of compute and the type of resources used (e.g., type of GPUs, internal cluster, or cloud provider)?
    \textbf{\answerNo}.
\end{enumerate}

\item If you are using existing assets (e.g., code, data, models) or curating/releasing new assets...
\begin{enumerate}
  \item If your work uses existing assets, did you cite the creators?
    \textbf{\answerYes. We cited the authors and include the URLs}. 
  \item Did you mention the license of the assets?
    \textbf{\answerNA. We used publicly-available datasets and code}.
  \item Did you include any new assets either in the supplemental material or as a URL?
    \textbf{\answerNo}.
  \item Did you discuss whether and how consent was obtained from people whose data you're using/curating?
    \textbf{\answerYes. We explained to participants how their data will be used}.
  \item Did you discuss whether the data you are using/curating contains personally identifiable information or offensive content?
    \textbf{\answerYes. We checked and ensured that our data does not contain personally identifiable information or offensive content}.
\end{enumerate}

\item If you used crowdsourcing or conducted research with human subjects...
\begin{enumerate}
  \item Did you include the full text of instructions given to participants and screenshots, if applicable?
    \textbf{\answerYes. See screenshots in \cref{sec:human_study_setup}}.
  \item Did you describe any potential participant risks, with links to Institutional Review Board (IRB) approvals, if applicable?
    \textbf{\answerNA}.
  \item Did you include the estimated hourly wage paid to participants and the total amount spent on participant compensation?
    \textbf{\answerYes. See \cref{sec:method_participants}. Our rate was \$13.5/hr,
    higher than the Prolific recommended rate wage of \$9.60/hr}.
\end{enumerate}

\end{enumerate}



\newcommand{\beginsupplementary}{%
    \setcounter{table}{0}
    \renewcommand{\thetable}{A\arabic{table}}%
    
    \setcounter{figure}{0}
    \renewcommand{\thefigure}{A\arabic{figure}}%
    
    \setcounter{section}{0}
    \renewcommand{\thesection}{A\arabic{section}}
    \renewcommand{\thesubsection}{\thesection.\arabic{subsection}}
}

\beginsupplementary%
\appendix

\newcommand{\toptitlebar}{
    \hrule height 4pt
    \vskip 0.25in
    \vskip -\parskip%
}
\newcommand{\bottomtitlebar}{
    \vskip 0.29in
    \vskip -\parskip%
    \hrule height 1pt
    \vskip 0.09in%
}

\newcommand{\suptitle}{Appendix for:\\\papertitle}

\newcommand{\maketitlesupp}{
    \newpage
    \onecolumn
        \null
        \vskip .375in
        \begin{center}
            \toptitlebar
            {\Large \bf \suptitle\par}
            \bottomtitlebar
            \vspace*{24pt}
            {
                \large
                \lineskip=.5em
                \par
            }
            \vskip .5em
            \vspace*{12pt}
        \end{center}
}

\maketitlesupp%


\section{Implementation details}

\subsection{Fine-tuning iNaturalist-pretrained ResNet-50 for CUB}
\label{sec:app_resnet_inat}

To make a 200-way classifier using the ResNet-50 model from iNaturalist \cite{iNaturalist_model}, we remove the 5089-way classification head and add an average pooling layer followed by a linear feed-forward layer with 200 units. We keep all the initialization parameters unchanged and use the Adam optimizer \cite{kingma2014adam} without any hyperparameter tuning. We train the new layer using the CUB training set for 200 epochs. We do not train the intermediate layers since this backbone is shared among all methods (i.e., we freeze all the convolutional layers in the ResNet-50 model).
The iNaturalist-pretrained ResNet-50 model has a slight difference compared to the PyTorch reference implementation \cite{pretraineds2022pytorch}. This network has 18 extra layers in the last convolutional blocks, but the spatial dimension matches the original ResNet-50 (i.e., $2048 \times 7 \times 7$).


\subsection{Implementation details for kNN}
\label{appendix:method_classifier2}

We implement a vanilla kNN classifier that operates at the deep feature space of ResNet-50.
That is, given a query image $Q$, we sort all training-set images $\{ G_i \}$ based on their distance $D (Q, G_i)$, which is the cosine distance between the two corresponding image features $f(Q)$ and $f(G_i) \in \sR^{2048}$ at \layer{layer4} of ResNet-50, after \layer{avgpool} (see \href{https://github.com/pytorch/vision/blob/main/torchvision/models/resnet.py#L205}{code}):

\begin{equation}
\operatorname{D}\left(Q, G_i\right)=1-\frac{\left\langle f(Q), f(G_i)\right\rangle}{\left\|f(Q)\right\|\left\|f(G_i)\right\|}
\label{eq:cosine_distance}
\end{equation}

where $ \langle \cdot \rangle$ is the dot product, and $\left\| \cdot \right\|$ is the $L_2$ norm operator.

\subsection{Implementation details for EMD-Corr}
\label{appendix:method_classifier3}

We incorporate the Earth Mover's Distance (EMD) into a 2-stage hierarchical image retrieval, similar to \cite{zhao2021towards, hai2022deepface}. 
In the first stage, the kNN classifier selects the $N$ images with the lowest cosine distance -- $G_i$ -- to the query $Q$.
Then, we sort these $N$ images (a.k.a. re-ranking \cite{hai2022deepface}) using patch-wise similarity derived from EMD. 
The predicted label is finally determined by a majority vote of the labels of the top-k images, as in the kNN classifier, where $k \leq N$. In our classifier, we set $k=20$ and $N=50$.




Our patch-wise comparison algorithm in stage 2 (shown in Fig.~\ref{fig:emd_based_class}) is different from \cite{hai2022deepface, zhao2021towards, zhang2020deepemd} as the similarity of an image pair is not determined by all possible patches.
While the first stage retrieved images using global features, comparing only a few most similar patches by EMD offers benefits: (1) helping classifiers capture the distinctive image regions only (e.g., head-to-head comparison for birds); and (2) achieving human interpretability as looking at all possible pair-wise comparisons is impossible. We denote each patch-by-patch comparison as ``correspondence''.

The most similar patches between two images $Q$ and $G$ -- both divided into $M$ patches -- are found as a set of 2-D coordinates $L$ containing the \emph{highest} values in a \emph{flow} matrix $\boldsymbol{F}$.
Let $\mathcal{Q} = \{ (q_1, w_{q_1}), (q_2, w_{q_2}), \cdots. (q_{M}, w_{q_M}) \}$ and $\mathcal{G} = \{ g_1, w_{g_1}), (g_2, w_{g_2}), \cdots. (g_{M}, w_{g_M}) \}$ denote two sets of non-overlapping image patches, $g_i$ and $g_j$ are the patch embeddings; and $w_{q_i}$ and $w_{g_j}$ are the corresponding importance assigned by a feature weighting algorithm (e.g., Cross Correlation used in \cite{zhao2021towards}). 
We derive $\boldsymbol{F}= (f_{ij}) \in \mathbb{R}^{M \times M}$ by minimizing the \emph{transport plan cost} in Eq.~\ref{eq:transport_plan}.

\begin{equation}
\operatorname{Cost}(Q, G, \boldsymbol{F})=\sum_{i=1}^{M} \sum_{j=1}^{M} d_{ij} f_{ij}
\label{eq:transport_plan}
\end{equation}

where $f_{i j} \geq 0$ and $\sum_{j=1}^{M} \sum_{i=1}^{M} f_{i j} = 1$.
We use Eq.~\ref{eq:cosine_distance} to compute the ground distance $d_{ij}$ and run the Sinkhorn algorithm \cite{cuturi2013sinkhorn} for 100 iterations to seek the \textit{optimal transport plan} $\boldsymbol{F}$.
To assign importance weights (i.e., $w_{q_i}$ and  $w_{g_j}$), we use cross-correlation (CC) maps from \cite{stylianou2019visualizing}.

Finally, using $\boldsymbol{F}$ and D from Eq.~\ref{eq:cosine_distance}, the EMD distance between $Q$ and $G$ is computed by Eq.~\ref{eq:emd_dist}. Since we are interested in patch-wise comparison, the features used in stage 2 for $Q$ and $G$ are \layer{layer4} from \cite{pretraineds2022pytorch}.
Our EMD-Corr classifier's stage 2 solely relies on EMD distance for re-ranking instead of mixing EMD and cosine distance like in previous works \cite{zhao2021towards, hai2022deepface}.

\begin{equation}
\operatorname{d_{EMD}}(D, \boldsymbol{F}) = \sum_{(i,j) \in L} d_{ij} f_{ij}
\label{eq:emd_dist}
\end{equation}

\subsection{Implementation details for CHM-Corr classifier}
\label{supp:chm_details}

Similar to the EMD-Corr classifier, this classifier also consists of 2 stages -- selecting $N$ most similar images to the query $Q$ by the kNN classifier, followed by a correspondence-based re-ranking algorithm.
For re-ranking, we propose to use a Convolutional Hough Matching network (CHM) \cite{min2021convolutional} to first infer semantic correspondences between $Q$ and $G$, then calculate the similarity score between the two images based on a subset of these correspondences. 



The re-ranking algorithm starts with dividing both $Q$ and $G$ into $M$ patches, resulting in two set of   $\mathcal{Q} = \{ q_1, q_2, \cdots. q_{M} \}$ and $\mathcal{G} = \{ g_1, g_2, \cdots. g_{M} \}$ image patches.
To find the semantic correspondences between two images, we make use of the CHM network to transfer keypoints from the query image $Q$ to image $G$.


The CHM network finds correspondence between two given images in three stages: feature extraction and correlation computation, Hough matching, and keypoint transfer. In the first stage, the CHM network extracts features from multiple layers of a ResNet-101 network to construct a set of multi-scale features $\left\{\left(\mathbf{F}_{Q}, \mathbf{F}_{G}\right)\right\}_{s=1}^{S}$. The feature volume is then used to construct a correlation tensor by comparing all possible pairs in the feature space of two images. In the second stage, the correlation tensor is fed into a Convlutioanl Hough Matching (CHM) layer to perform Hough voting in the space of translation and scale to find candidate matches between two images. In the last stage, a kernel soft-argmax \cite{lee2019sfnet} is applied to the output of the CHM layer to create a dense flow field, and then correspondence keypoints are extracted using a soft sampler.

After finding visual correspondence between two images, we assign an importance weight $w_{i,j}$ for the pair $(q_i, g_j)$ using cross-correlation maps from \cite{stylianou2019visualizing}. 
Finally, the distance between $Q$ and $G$ is the average distance between 5 patch pairs with the lowest cosine distance.

We use the reference implementation of the Convolutional Hough Matching Network pretrained on the PF-Pascal Dataset \cite{ham2017proposal}. 
There are three variations of CHM networks depending on the parameter sharing strategy, i.e., \texttt{psi}, \texttt{iso}, and \texttt{full}. 
Our ablation study (\cref{supp:ablation_chm}) shows similar performance on a 5K subset of the ImageNet dataset. We select \texttt{psi} with a threshold of $T=0.55$ for the CHM-Corr classifier.

The CHM network requires a set of initial keypoints on the source image, i.e., a set of keypoint on the query image $Q$. Although some datasets come with this annotation information, generally, this information is not available. To have a comparable classifier with our EMD-Corr classifier, we discretize an image into a $7 \times 7$ grid, resulting in $49$ non-overlapping patches. For each patch, we pick a point at its center.

For assigning importance weight $w_{i, j}$ to $(q_i, g_j)$ pair, we first calculate the cross-correlation map between the two images $Q$ and $G$. 
Calculating a cross-correlation map using the last convolutional layer of the ResNet-50 model will result in two $7 \times 7$ maps for each  $Q$ and $G$.
For assigning importance weights, we binarize the cross-correlation for $Q$, using a threshold of $T=0.55$, i.e., we zero out all pairs in the non-salient part according to $Q$, by setting their importance weights to $0$, and for the remaining patches, we set the weights to $1$. 

After removing non-salient patch pairs in the last stage, we calculate the cosine similarity between pair $(q_i, g_j)$ using the corresponding feature volume in the last convolutional layer of the ResNet-50 model. The similarity score is the average similarity between top $5$ pairs with the highest cosine similarly.

\subsection{Generating Adversarial Patch dataset}
\label{supp:generating_adversarial_patches}

\citet{brown2017adversarial} generated a \textit{universal} adversarial patch to fool image classifiers into recognizing everything as \class{toaster}. 
This patch misleads the models' attention, by having them look only at the most salient item while ignoring the remaining pixels.
We apply this attack on ImageNet validation set, resulting in 50K Adversarial Patch images of $240\times240$ px. 
The patches are circles with a size of $5$\% the input image, targeting ResNet-50 \cite{he2016deep} classifying everything as \class{toaster} with a target confidence of 90\%.
The maximum attack iteration for each sample is 500.
We only train to optimize the adversarial patch on the ImageNet validation set for one epoch and save the immediate samples for the dataset.
We adapt the code from \cite{adv_patch_code} and make minor modifications. 

To obtain our Adversarial Patch dataset, from the \href{https://github.com/anguyen8/visual-correspondence-XAI}{main repository}, you can run the below command to generate the dataset or download the dataset \href{https://drive.google.com/file/d/1EALgEHzAJxS1oLo18-WOQ9jeHAOInTin/view?usp=sharing}{here}.

\begin{lstlisting}[language=bash]
cd datasets/adversarial-patch/
python make_patch.py --cuda --epochs 1 --patch_size 0.05 --max_count 500
--netClassifier resnet50 --patch_type circle  --train_size 50000 
--test_size 0 --image_size 240 --outf output_imgs
\end{lstlisting}





\clearpage

\section{Ablation study and small-scale experiments on ImageNet OODs}

\subsection{Different hyperparameters for EMD}
\label{supp:emd_diff_hypers}

\begin{table}[!htbp]
\centering
\caption{Accuracy of the EMD-Corr classifier with different EMD hyperparameters (\%) }
\label{tab:emd_diff_hypers}
\begin{tabular}{@{}lcccc@{}}
\toprule
\textbf{Datasets} & \begin{tabular}[c]{@{}c@{}}Number \\ of \\ Images\end{tabular} & \begin{tabular}[c]{@{}c@{}}Cross Correlation\\ Corrs-Num$= 5$\\ $k=20$\end{tabular} & \begin{tabular}[c]{@{}c@{}}Cross Correlation\\ Corrs-Num$= 49 \times 49$\\ $k=20$\end{tabular} & \begin{tabular}[c]{@{}c@{}}Uniform\\ Corrs-Num$=5$\\ $k=20$\end{tabular} \\ \midrule
ImageNet 2012 & 50,000 & 74.93 & 74.59 & 74.47 \\
\begin{tabular}[c]{@{}l@{}}CUB\\ (iNaturalist ResNet)\end{tabular} & 5,794 & 84.98 & 85.42 & 79.72 \\
\begin{tabular}[c]{@{}l@{}}CUB\\ (ImageNet ResNet)\end{tabular} &  5,794 & \na & 59.44 & 53.47 \\ \bottomrule
\end{tabular}
\end{table}


\subsection{Performance of classifiers on a 5K subset of different datasets}


\cref{tab:ablationstudy-otherdatasets} contains details about the performance of different classifiers on a 5K subset of various OOD datasets.  

\begin{table}[!htbp]
\centering
\caption{Performance of classifiers on 5K subsets of various OOD datasets
-- (Accuracy  \%) }
\label{tab:ablationstudy-otherdatasets}
\begin{tabular}{@{}lrrrr@{}}
\toprule
\textbf{Datasets} & \textbf{ResNet-50} & \textbf{kNN} & \textbf{EMD-Corr} & \textbf{CHM-Corr} \\ \midrule
ImageNet \cite{russakovsky2015imagenet} & 75.00 & 74.62 & 74.66 & 74.52 \\
ImageNet-R \cite{hendrycks2021many} & 35.68 & 34.60 & 35.66 & 36.18 \\
ObjectNet \cite{barbu2019objectnet} & 36.54 & 34.80 & 36.56 & 35.60 \\
ImageNet Sketch \cite{wang2019learning}  & 23.84 & 23.92 & 24.40 & 25.28 \\
ImageNet-A \cite{hendrycks2021natural} & 0.00 & 0.32 & 0.50 & 0.46 \\
DAmageNet \cite{chen2020universal} & 6.38 & 8.92 & 9.72 & 9.06 \\
ImageNet-C Gaussian noise (Level 1) \cite{hendrycks2019benchmarking} & 59.56 & 59.62 & 59.70 & 59.62 \\
ImageNet-C Gaussian blur (Level 1) \cite{hendrycks2019benchmarking} & 66.12 & 65.68 & 65.68 & 65.68 \\  \bottomrule
\end{tabular}
\end{table}

\subsection{Different weights for CHM}
\label{supp:ablation_chm}

\begin{table}[!htbp]
\centering
\caption{Accuracy of the CHM-Corr classifier on a 5K subset of ImageNet \cite{russakovsky2015imagenet} with different CHM parameters (\%) }
\label{tab:threshold_supp}
\begin{tabular}{@{}ccccccccc@{}}
\toprule
\multirow{2}{*}{Method} & \multicolumn{8}{c}{Threshold} \\ \cmidrule(l){2-9} 
 & 0.2 & 0.3 & 0.4 & 0.5 & 0.55 & 0.6 & 0.7 & 0.8 \\ \midrule
\texttt{PSI} & 74.26 & 74.36 & 74.36 & 74.26 & \textbf{74.52} & 74.38 & 74.44 & 73.78 \\
\texttt{ISO} & 74 & 74.04 & 74 & 74.18 & 74.24 & \textbf{74.28} & 74.1 & 73.76 \\
\texttt{FULL} & 74.62 & 74.62 & 74.48 & \textbf{74.64} & 74.4 & 74.44 & 74.56 & 74.02 \\ \bottomrule
\end{tabular}
\end{table}

\clearpage




\clearpage
\section{Runtime comparison between all methods}
\label{supp:runtime}

In this section, we provide a runtime analysis of all classifiers on a batch of 1000 random queries. For each classifier, we run the classification five times and report the average and standard deviation. We use a single NVIDIA V100 GPU with 16 gigabytes of memory to perform our benchmarks.

Here we also provide a FAISS \cite{johnson2019billion} implementation of the kNN classifier, which is significantly faster than the naive GPU implementation for the nearest neighbor search problem. The FAISS version of kNN requires one-time preprocessing to extract embeddings from the training set. This process takes just a few minutes for the CUB dataset, which has only 5.9K images. For ImageNet, which consists of 1.2 million images, we use a single NVIDIA A100 (40 GB) to extract and cache the embeddings on disk. This process takes less than 90 minutes, and the resulting cache file takes 9.8 GB of disk space. We also use the Linux's \texttt{time} tool to calculate the total memory usage of kNN using FAISS during the inference. The peak memory performance (\texttt{Maximum resident set size}) for the 1000 images is around 31 GB.

\begin{table}[!htbp]
\centering
\caption{Runtime (in seconds) for a set of 1,000 queries averaged over 5 runs -- kNN inference is fairly tractable using a FAISS implementation.
}
\label{tab:runtimeperformance}
\begin{tabular}{@{}lrr@{}}
\toprule
\multicolumn{1}{c}{\multirow{2}{*}{Method}} & \multicolumn{2}{c}{Dataset} \\
\multicolumn{1}{r}{} & ImageNet & CUB \\ \midrule
ResNet-50 & 9.17 $\pm$ 0.19 & 8.81 $\pm$ 0.14 \\
kNN (FAISS - CPU) & 17.35 $\pm$ 1.28 & 9.7 $\pm$ 0.32 \\
kNN (Naive - GPU) & 1,112.46 $\pm$ 0.86 & 23.88 $\pm$ 0.58 \\
EMD-Corr reranking step & 2,218.92 $\pm$ 99.14 & 1,927.69 $\pm$ 17.48 \\
CHM-Corr reranking step & 10,642.85 $\pm$ 1007.87 & 6,920.76 $\pm$ 67.58 \\ \bottomrule
\end{tabular}
\end{table}

\clearpage
\section{Sample explanations}
\label{supp:sample_xais}

This section contains sample visualizations for kNN, EMD-Corr, and CHM-Corr classifiers.



\subsection{kNN}
\label{supp:sample_knn_xai}

\begin{figure}[!htbp]
        \begin{subfigure}[b]{1\textwidth}
        \includegraphics[width=\linewidth]{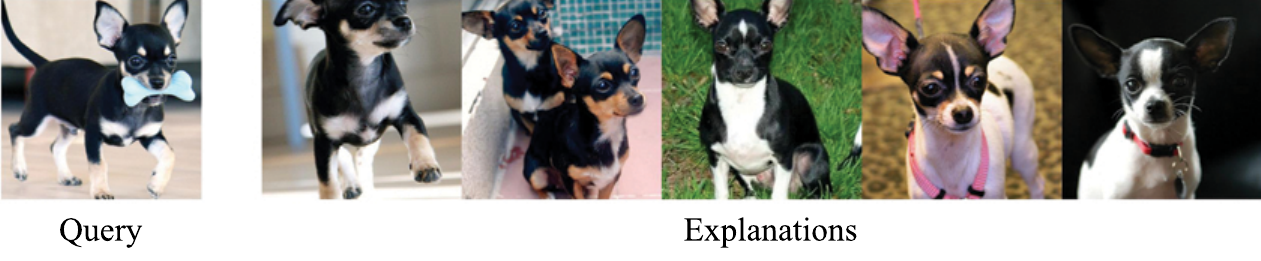}
        \end{subfigure}\par
        \caption{A sample explanation of the kNN classifier when classifying a \class{chihuahua} image.}
        \label{fig:sample_xai_knn}
\end{figure}

\subsection{EMD-NN}
\label{supp:sample_emd_nn_xai}

\begin{figure}[!htbp]
        \begin{subfigure}[b]{1\textwidth}
        \includegraphics[width=\linewidth]{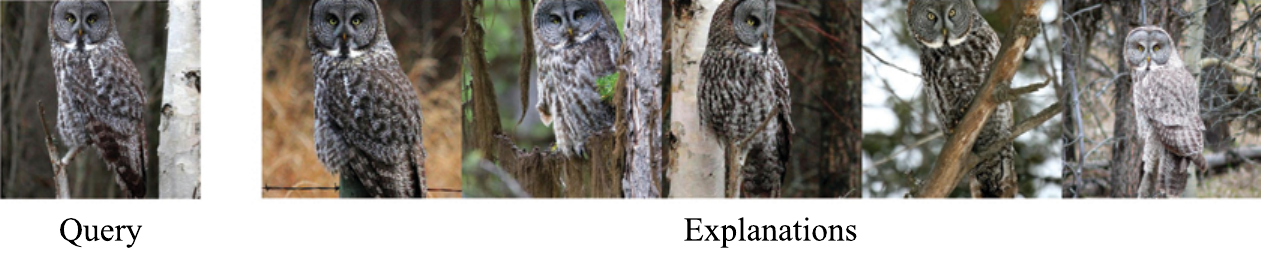}
        \end{subfigure}\par
        \caption{A sample EMD-NN explanation of the EMD-Corr classifier when classifying a \class{great grey owl} image.
        EMD-NN shows only the nearest neighbors after re-ranking.
        }
        \label{fig:sample_xai_emd}
\end{figure}

\subsection{EMD-Corr}
\label{supp:sample_emd_corr_xai}

\begin{figure}[!htbp]
        \begin{subfigure}[b]{1\textwidth}
        \includegraphics[width=\linewidth]{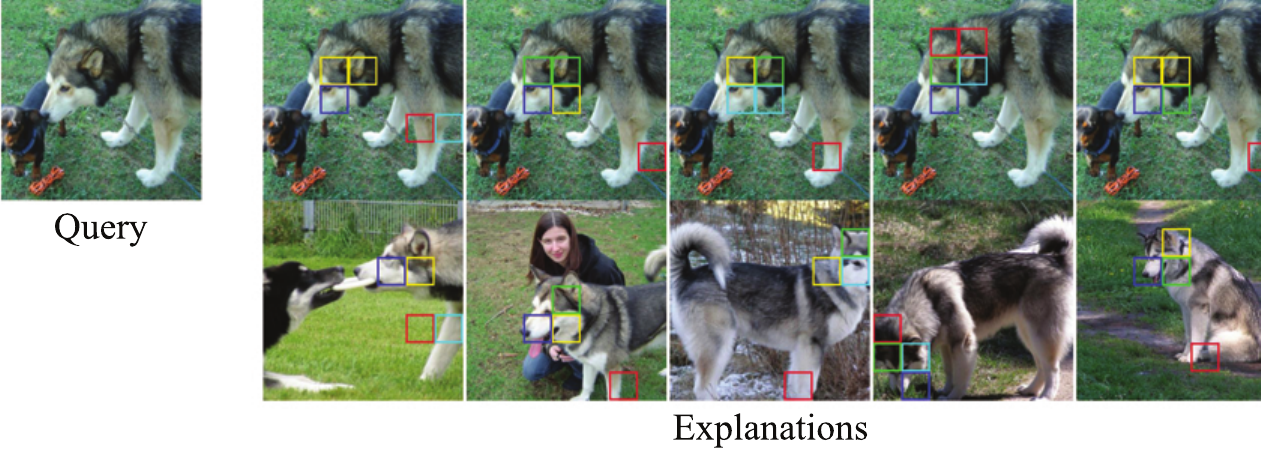}
        \end{subfigure}\par
        \caption{A sample explanation of the EMD-Corr classifier when classifying a \class{malamute} image.}
        \label{fig:sample_xai_emd_corr}
\end{figure}

\clearpage
\subsection{CHM-NN}
\label{supp:sample_chm_nn_xai}
\begin{figure}[!htbp]
        \begin{subfigure}[b]{1\textwidth}
        \includegraphics[width=\linewidth]{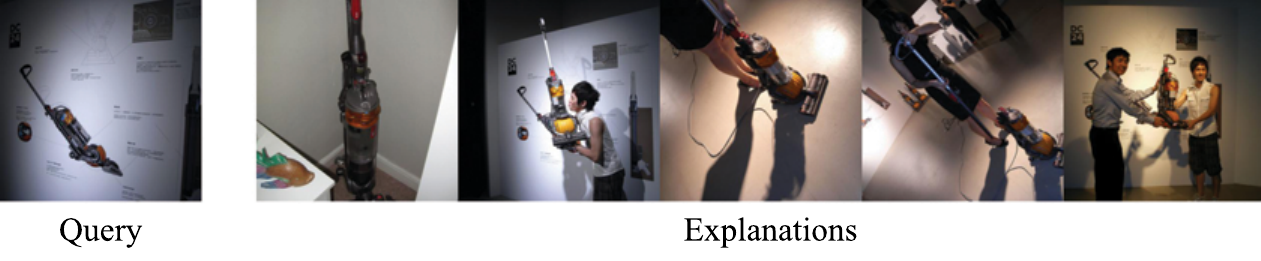}
        \end{subfigure}\par
        \caption{A sample CHM-NN explanation of the CHM-Corr classifier when classifying a \class{vacuum} image.
        CHM-NN only shows the nearest neighbors after re-ranking.
        }
        \label{fig:sample_xai_chm}
\end{figure}

\subsection{CHM-Corr}
\label{supp:sample_chm_corr_xai}

\begin{figure}[!htbp]
        \begin{subfigure}[b]{1\textwidth}
        \includegraphics[width=\linewidth]{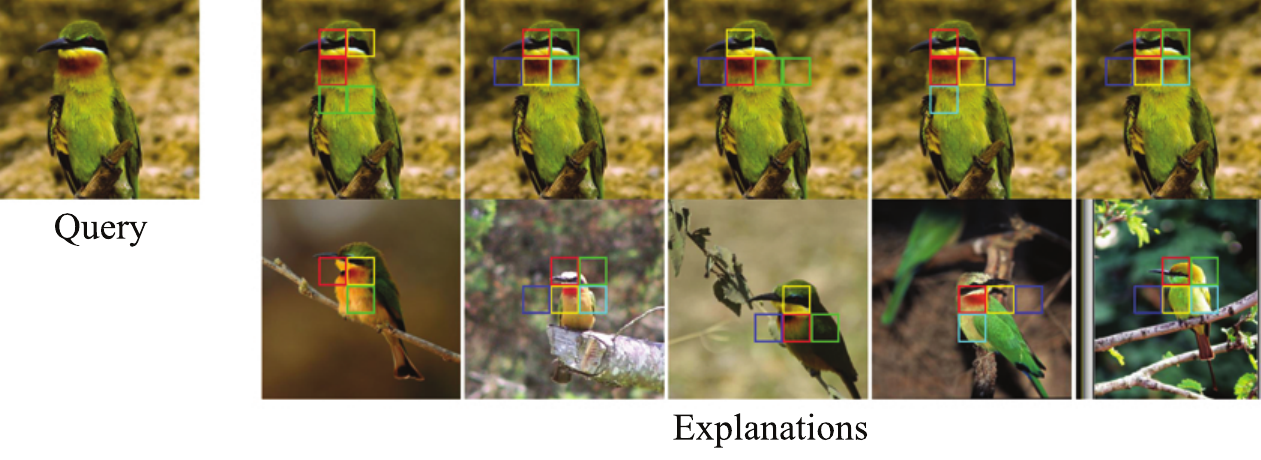}
        \end{subfigure}\par
        \caption{A sample explanation of the CHM-Corr classifier when classifying a \class{bee eater} image.}
        \label{fig:sample_xai_chm_corr}
\end{figure}

\clearpage

\section{Sample screens and training examples for human studies}
\label{sec:human_study_setup}

\subsection{Sample screens from human studies}
\label{supp:sample_images}

\begin{figure}[!hbt]
    \centering
    \begin{subfigure}[b]{\textwidth}
        \centering
        \includegraphics[width=0.9\textwidth]{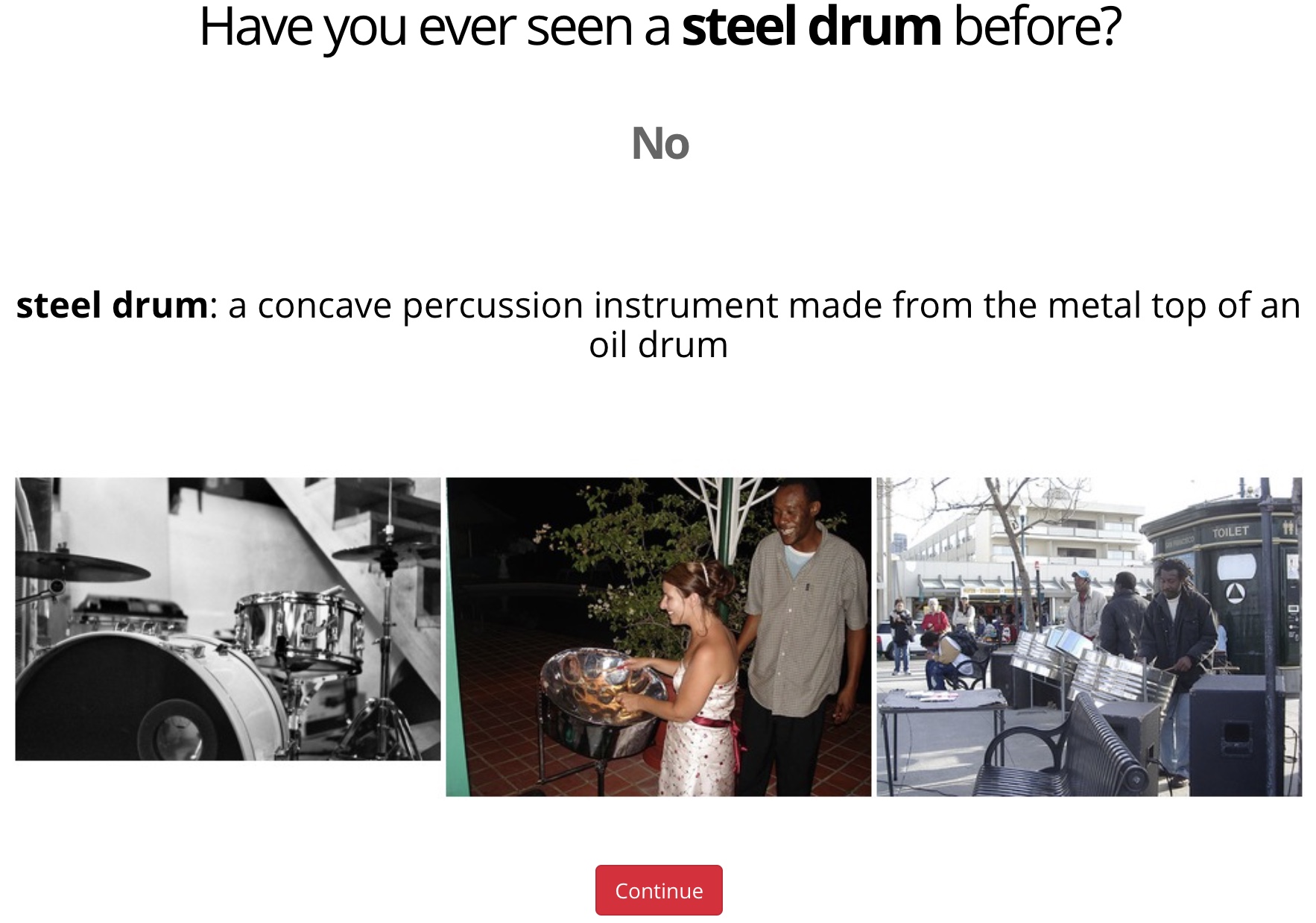}
        \caption{ImageNet studies}
    \end{subfigure}
    \hfill
    \begin{subfigure}[b]{\textwidth}
        \centering
        \includegraphics[width=0.7\textwidth]{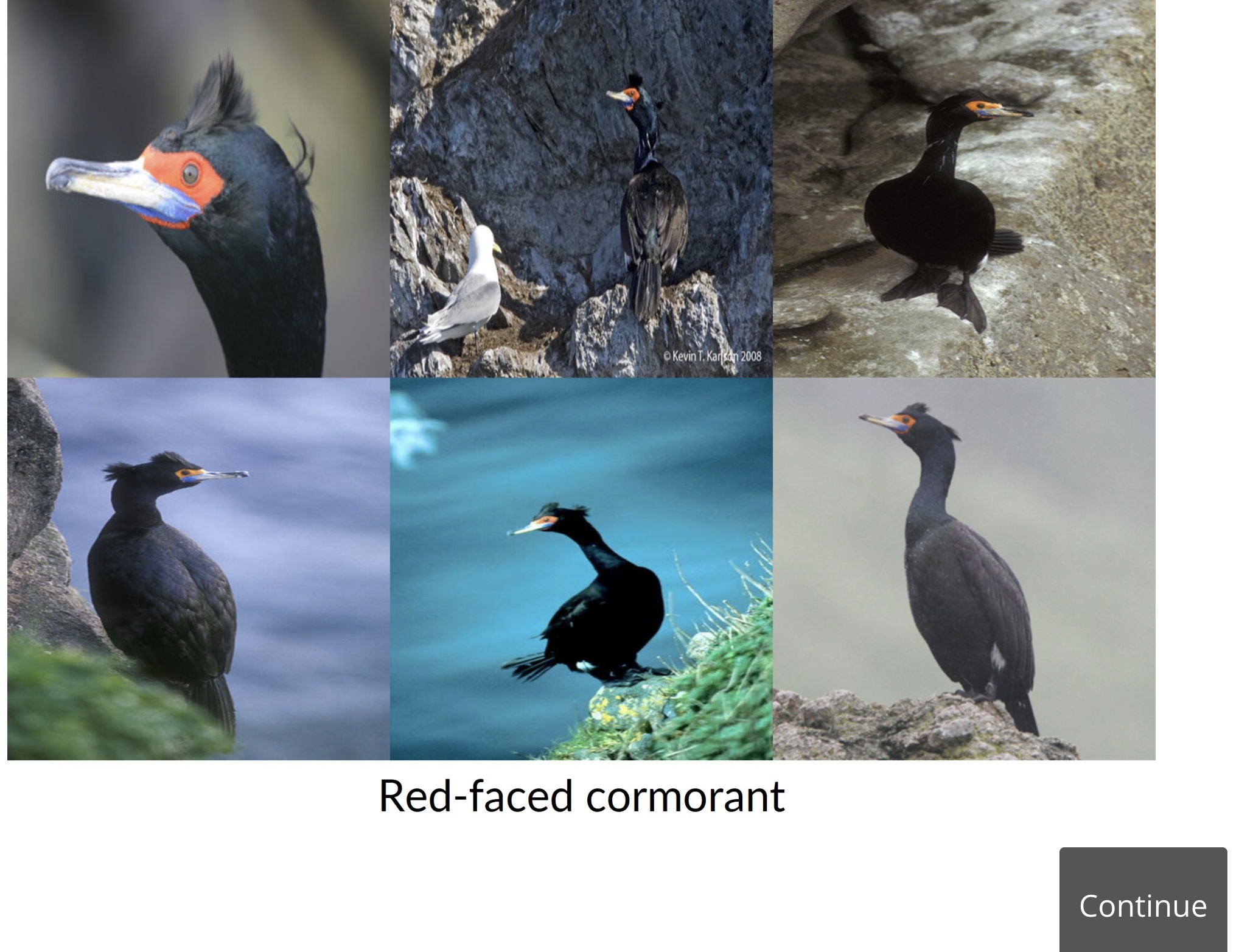}
        \caption{CUB studies}
    \end{subfigure}
    \caption{
    In ImageNet-ReaL experiments, before each trial, where users are asked if the query image belongs to the top-1 class $\vc$ (here, \class{steel drum}), we show three representative images from $\vc$ along with a 1-sentence WordNet description (a).
    Instead of showing 3 images, in CUB experiments, we offer 6 images from the top-1 class (here, \class{Red-faced Cormorant} to help users better recognize the distinct features of each bird.
    }
    \label{fig:sample_images}
\end{figure}

\begin{figure}[t]
        \centering
        \includegraphics[width=1.0\textwidth]{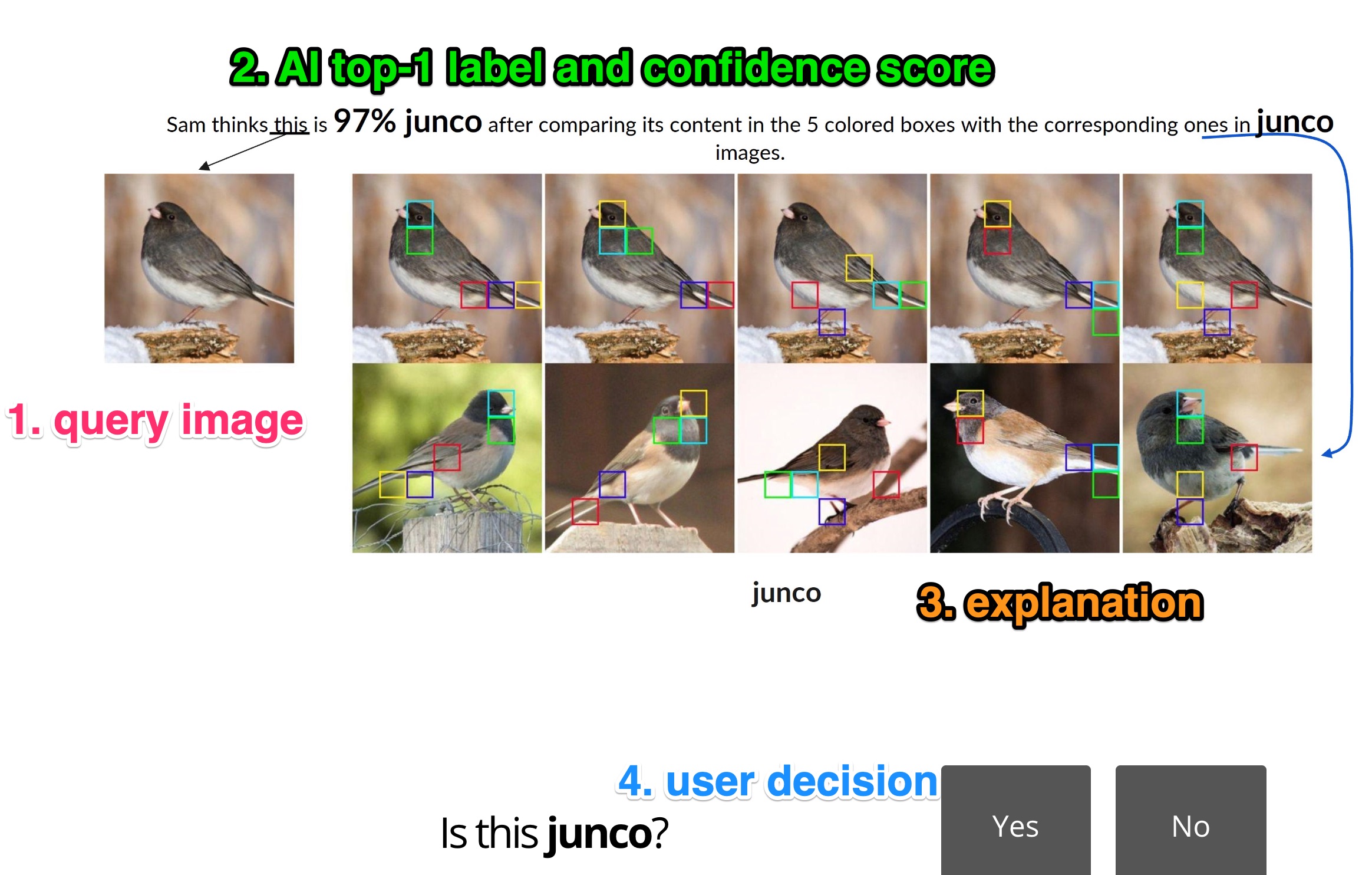}
    \caption{A sample screenshot from a human study of EMD-Corr users.
    Each user is provided with (1) the query image; (2) the AI top-1 predicted label and confidence score; and (3) an explanation, here the visual correspondence-based explanations of EMD-Corr.
    They are asked to provide a Yes/No answer to whether the query is an image of \class{junco}.
    }
    \label{fig:sample_trial_screen}
\end{figure}

\clearpage
\subsection{Sample groundtruth cases used in the Validation phase of our CUB human studies}
\label{supp:validation_images}

Below are example cases that we manually choose to be ``groundtruth'' in order to control for user quality during the validation phase.

\begin{figure*}[!hbt]
    \centering
    \begin{subfigure}[b]{\textwidth}
        \centering
        \includegraphics[width=0.2\textwidth]{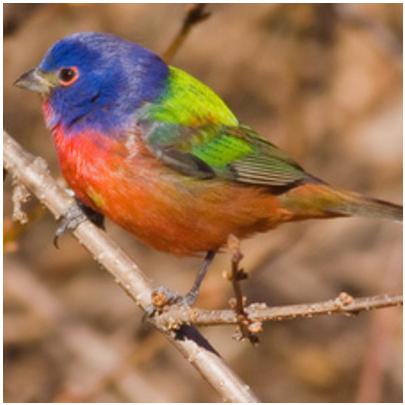}
        \caption{}
    \end{subfigure}
    
    \hfill
    \begin{subfigure}[b]{\textwidth}
        \centering
        \includegraphics[width=1.0\textwidth]{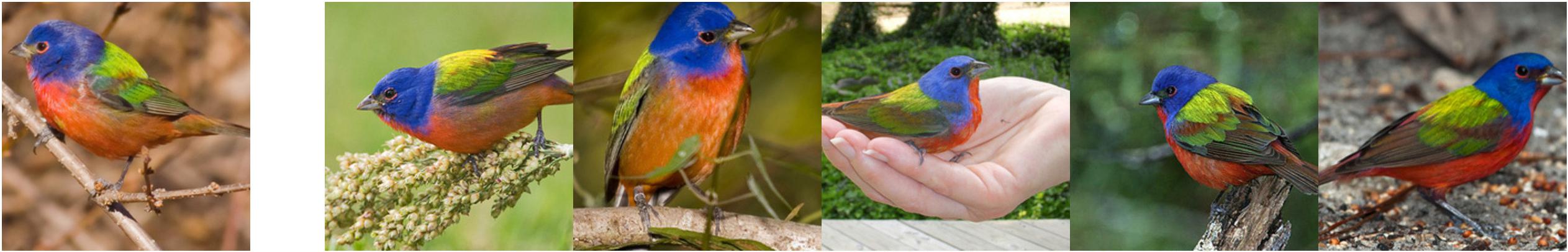}
        \caption{}
    \end{subfigure}
    
    \hfill
    \begin{subfigure}[b]{\textwidth}
        \centering
        \includegraphics[width=1.0\textwidth]{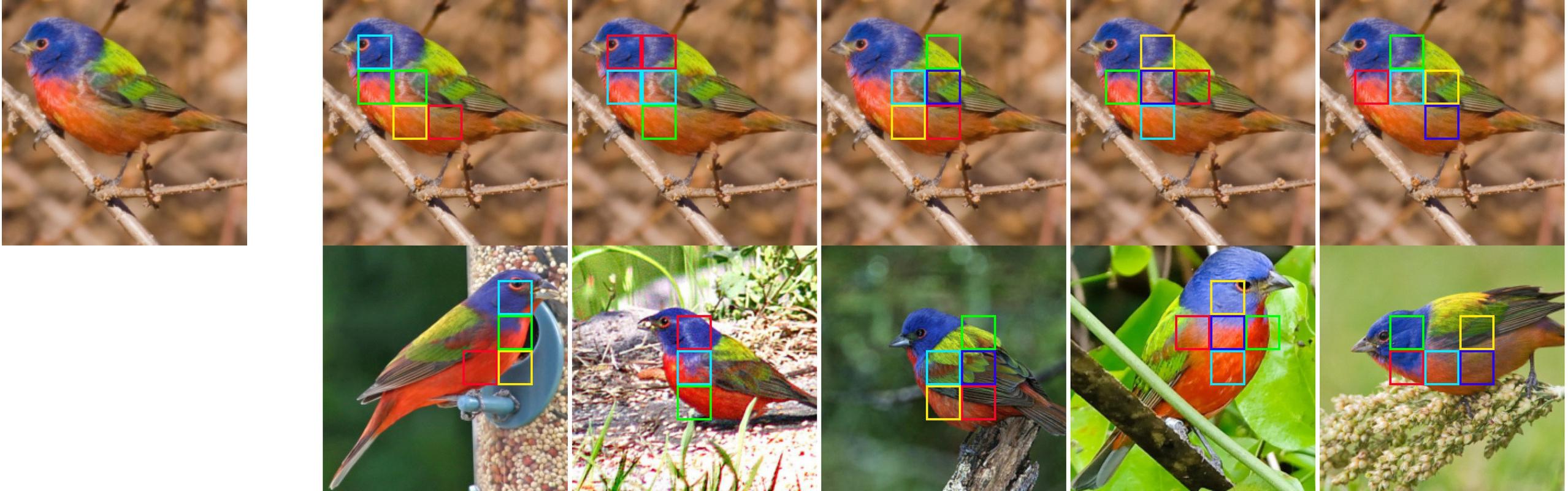}
        \caption{}
    \end{subfigure}
    
    \caption{A \textbf{groundtruth Yes} validation sample in a CUB study.
    That is, users are expected to select Yes when being presented with these explanations.
    The bird is \class{Painted Bunting}.\\
    (a) ResNet-50---no explanations provided.\\
    (b) kNN nearest-neighbor explanation.\\
    (c) EMD-Corr explanation.}
    \label{fig:cub_validation_images_correct}
\end{figure*}

\begin{figure*}[!hbt]
    \centering
    \begin{subfigure}[b]{\textwidth}
        \centering
        \includegraphics[width=1.0\textwidth]{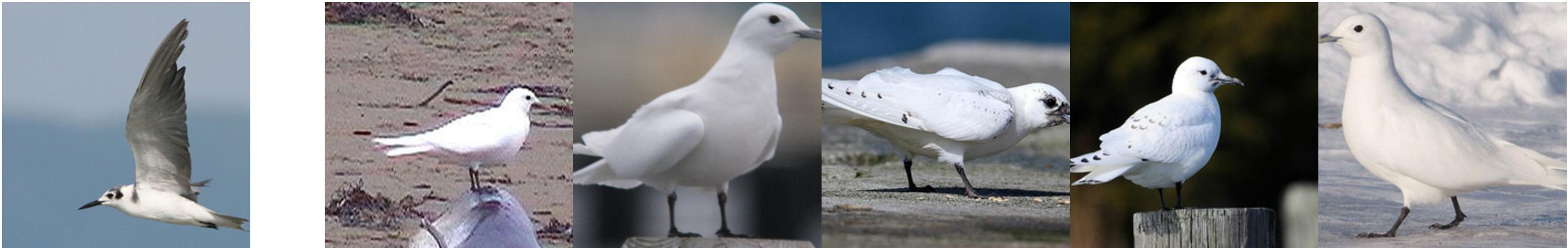}
        \caption{}
    \end{subfigure}
    
    \hfill
    \begin{subfigure}[b]{\textwidth}
        \centering
        \includegraphics[width=1.0\textwidth]{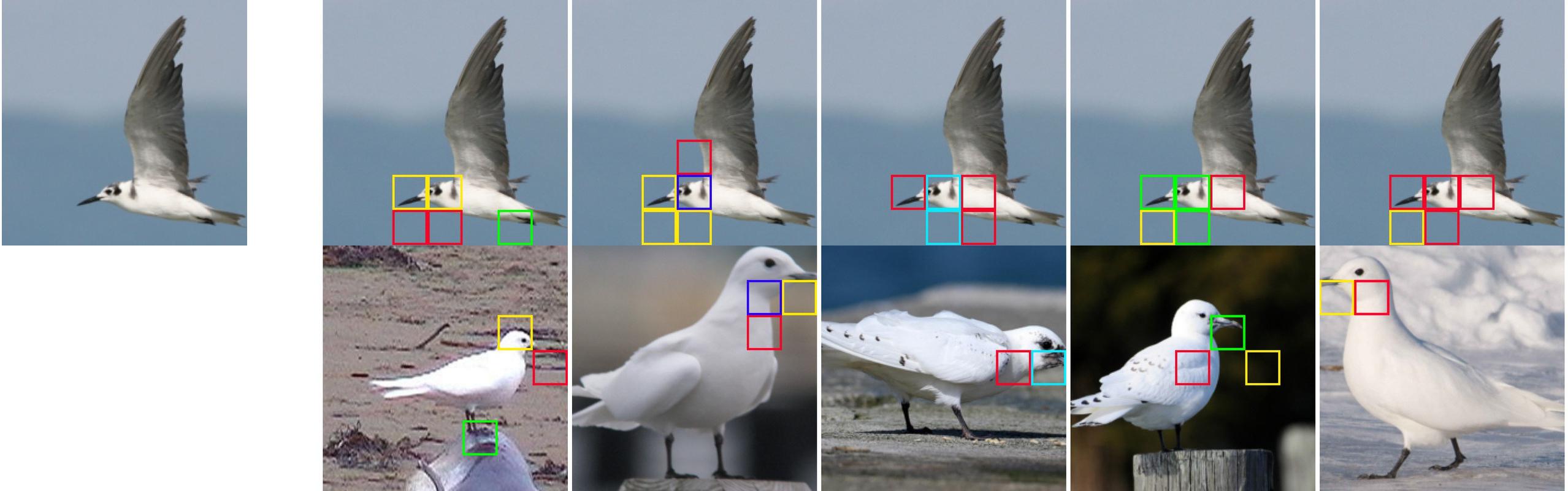}
        \caption{}
    \end{subfigure}
    
    
    \caption{A \textbf{groundtruth No} validation sample in a CUB study.
    That is, users are expected to select No when being presented with these explanations.
    The bird is \class{Black Tern}.\\
    (a) EMD-NN explanation.\\
    (b) EMD-Corr explanation.\\
    }    
    \label{fig:cub_validation_images_wrong}
\end{figure*}

\clearpage
\section{Human-AI team performance analysis}

This section provides more details about AI and team performance.




\subsection{Defining the difficulty level of queries}
\label{supp:defining-levels}

To further understand the performance of the classifiers in our study, we categorize each query image into Easy, Medium, and Hard categories based on the model's confidence score and the correctness of the top-1 label (see \cref{tab:diff-levels}). 
This breakdown allows us to analyze model and user behaviors at a specific level of difficulty.


\begin{table}[!htbp]
\centering
\caption{Difficulty levels}
\label{tab:diff-levels}
\begin{tabular}{@{}llll@{}}
\toprule
                       & \multicolumn{1}{c}{\textbf{Easy}} & \multicolumn{1}{c}{\textbf{Medium}} & \multicolumn{1}{c}{\textbf{Hard}} \\ \midrule
\textbf{AI is Correct}   & confidence $\in [0.75, 1)$        & confidence $\in [0.35, 0.75)$       & confidence $\in [0, 0.35)$        \\
\textbf{AI is Wrong} & confidence $\in [0, 0.35)$        & confidence $\in [0.35, 0.75)$       & confidence $\in [0.75, 1)$        \\
\bottomrule
\end{tabular}
\end{table}

\clearpage

\subsection{The acceptance and rejection ratios}
\label{sec:appendix_reject_ratio}

In Table \ref{tab:users-accept-reject-data} we provide details about whether users accepted or rejected AI's decisions for each type of classifier.

\begin{table}[!htbp]
\centering
\caption{The frequency of users accepting or rejecting AI's decision per classifier (\%).
}
\label{tab:users-accept-reject-data}
\begin{tabular}{@{}lcccc@{}}
\toprule
\multicolumn{1}{c}{\multirow{2}{*}{Method}} & \multicolumn{2}{c}{ImageNet-ReaL} & \multicolumn{2}{c}{CUB} \\ \cmidrule(l){2-5} 
\multicolumn{1}{c}{} & Accept & Reject & Accept & Reject \\ \midrule
\resnet & 60.44 & 39.56 & 74.28 & 25.72 \\
\knn & \textbf{69.60} & 30.40 & \textbf{81.53} & 18.47 \\
\emdcorr & 64.92 & 35.08 & 67.30 & \textbf{32.70} \\
\chmcorr & 67.51 & 32.49 & 66.27 & \textbf{33.73}\\ \hline
EMD-NN & 67.49 & 32.51 & 78.76 & 21.24 \\
CHM-NN & 68.94 & 31.06 & 76.94 & 23.06 \\
\bottomrule
\end{tabular}
\end{table}

Table \ref{tab:accept-reject-diff} shows the ratio of accepts and rejects based on the difficulty level described in Sec. \ref{supp:defining-levels}.

\begin{table}[!htbp]
\centering
\caption{The ratio of users accepting or rejecting AI's decision per difficulty level (\%)}
\label{tab:accept-reject-diff}
\begin{tabular}{@{}lcccc@{}}
\toprule
\multicolumn{1}{c}{\multirow{2}{*}{Difficulty Level}} & \multicolumn{2}{c}{ImageNet-ReaL} & \multicolumn{2}{c}{CUB} \\ \cmidrule(l){2-5} 
\multicolumn{1}{c}{} & Accept & Reject & Accept & Reject \\ \midrule
Easy & 72.7 & 27.3 & 82.75 & 17.25 \\
Medium & 58.42 & 41.58 & 66.43 & 33.57 \\ 
Hard & 62.38 & 37.62 & 78.34 & 21.66 \\ \bottomrule
\end{tabular}
\end{table}

\subsection{Time performance of users}

\begin{figure}[!htb]
        \begin{subfigure}[b]{0.5\textwidth}
                \includegraphics[width=\linewidth]{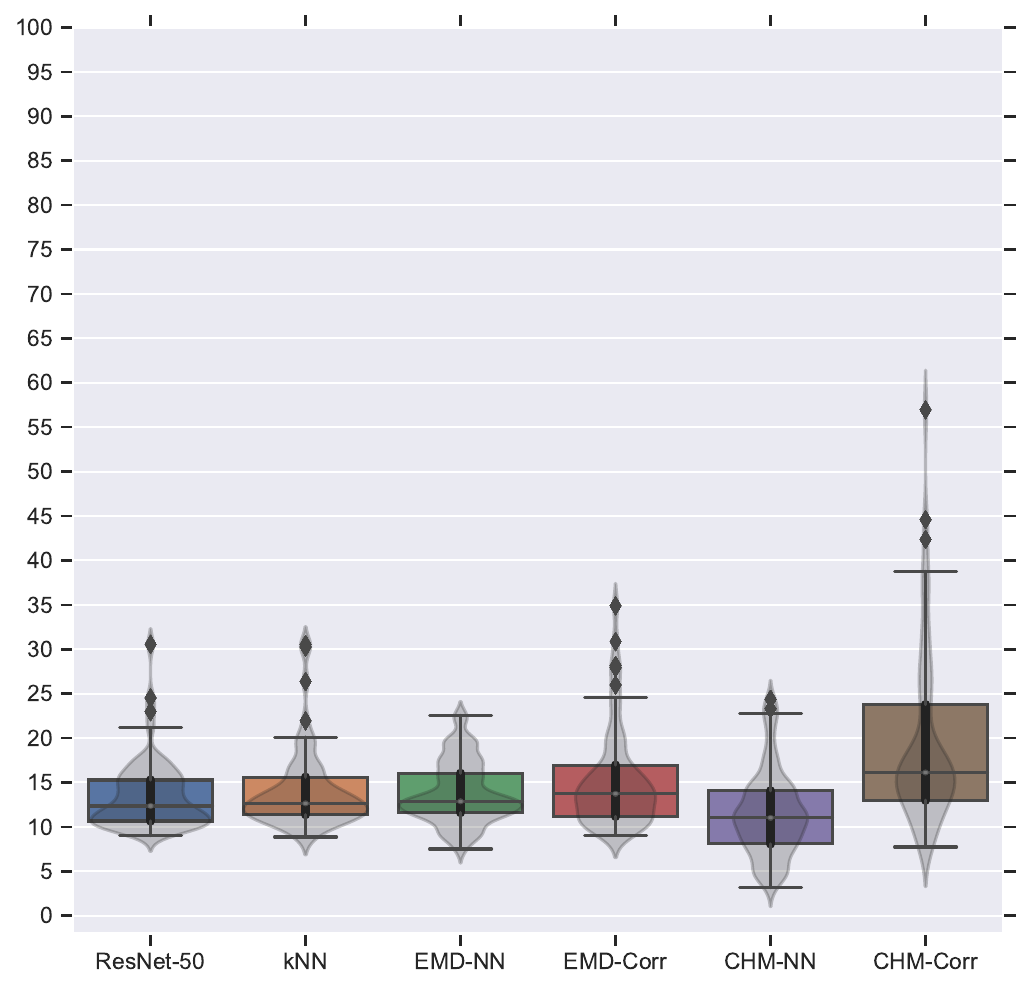}
                \caption{ImageNet}
                \label{fig:time-imagenet-supp}
        \end{subfigure}%
        \begin{subfigure}[b]{0.5\textwidth}
                \includegraphics[width=\linewidth]{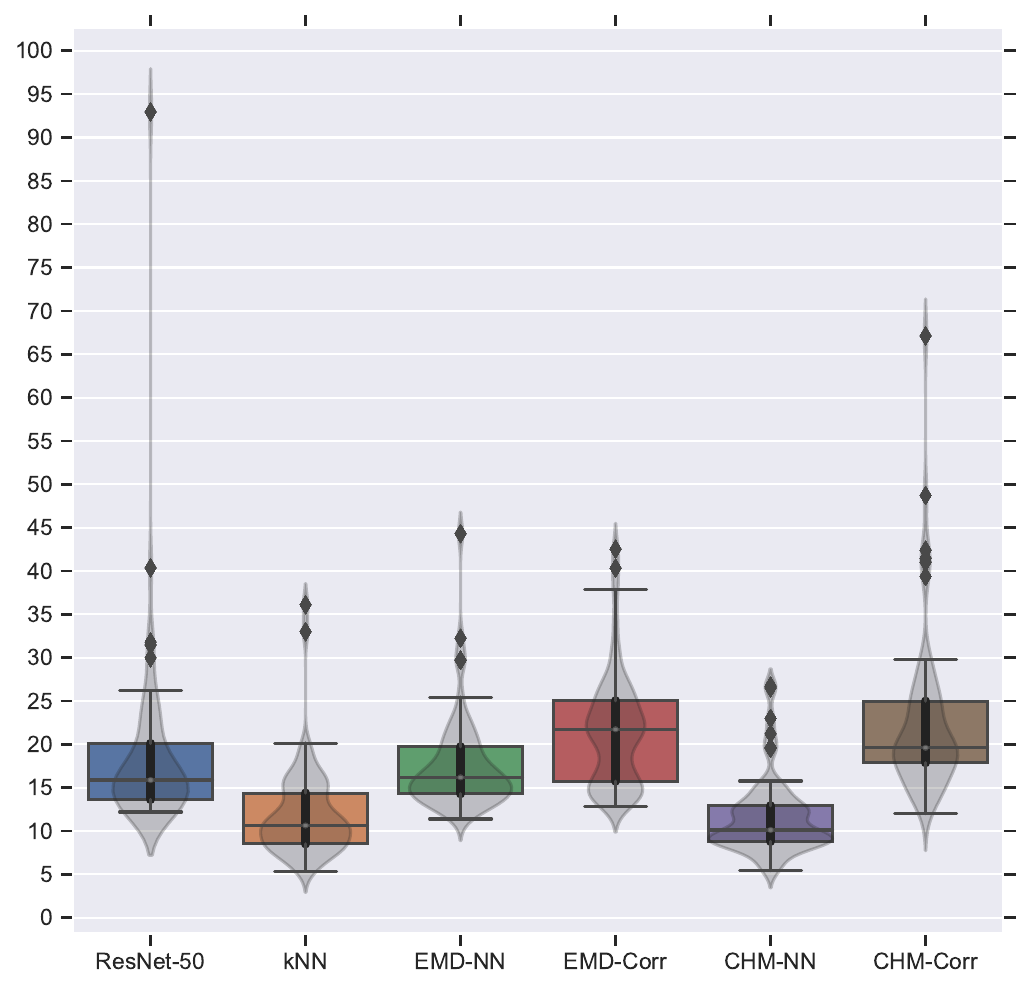}
                \caption{CUB}
                \label{fig:time-cub-supp}
        \end{subfigure}%
        \caption{Distribution of the average time taken for each trial (Seconds) }
        \label{fig:time-performance-supp}
\end{figure}

\cref{fig:time-performance-supp} shows the average time distribution to finish each trial per method.
 
\clearpage
\subsection{Human performance analysis based on AI correctness}
\label{supp:user-accuracy-on-ai-correctness}

\begin{figure}[!htbp]
        \begin{subfigure}[b]{1\textwidth}
        \includegraphics[width=\linewidth]{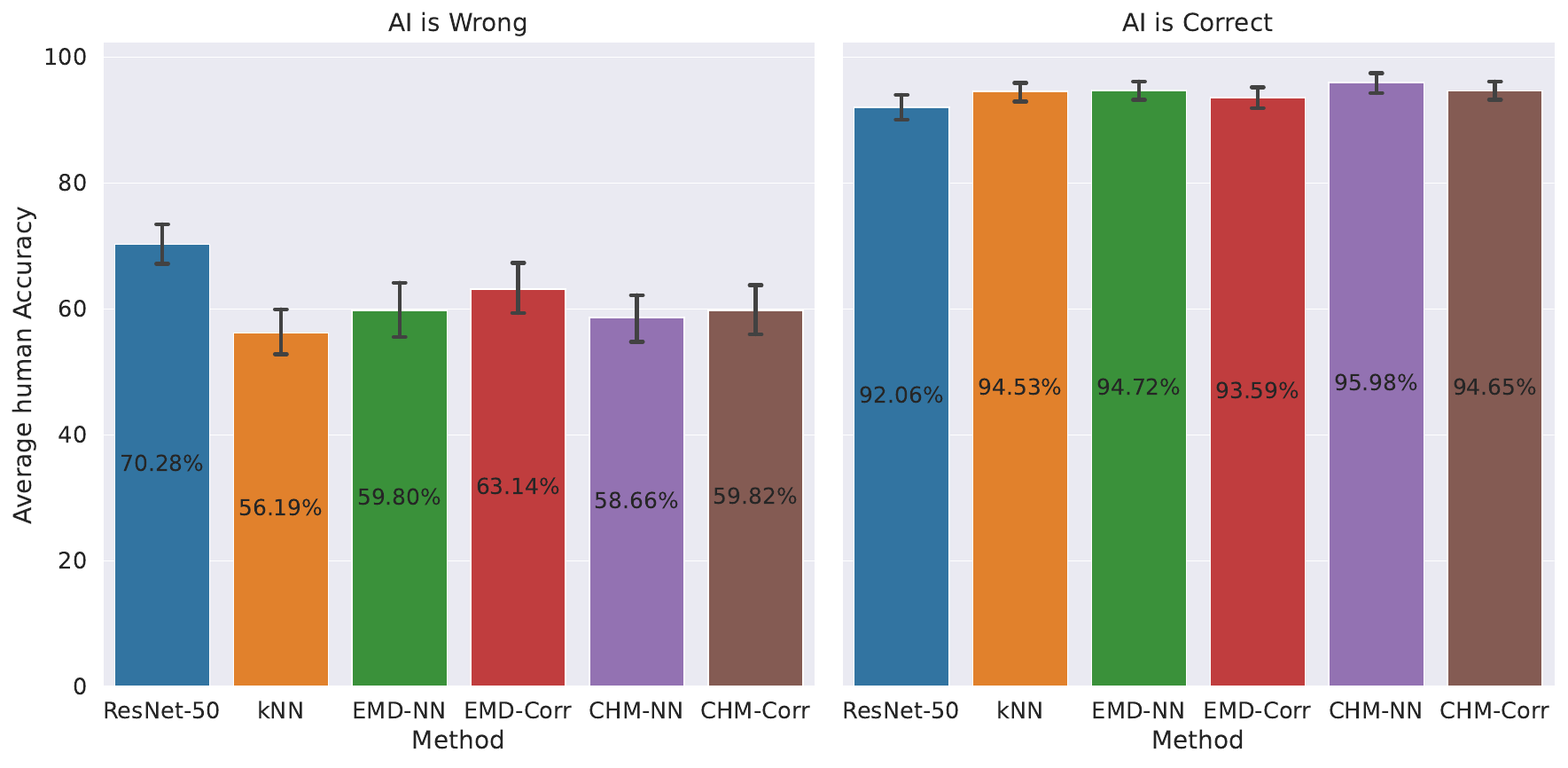}
        \caption{ImageNet-ReaL -- Mean user accuracy}
         \label{fig:imagenet-ai-correctness}
        \end{subfigure}\par
        \begin{subfigure}[b]{1\textwidth}
        \includegraphics[width=\linewidth]{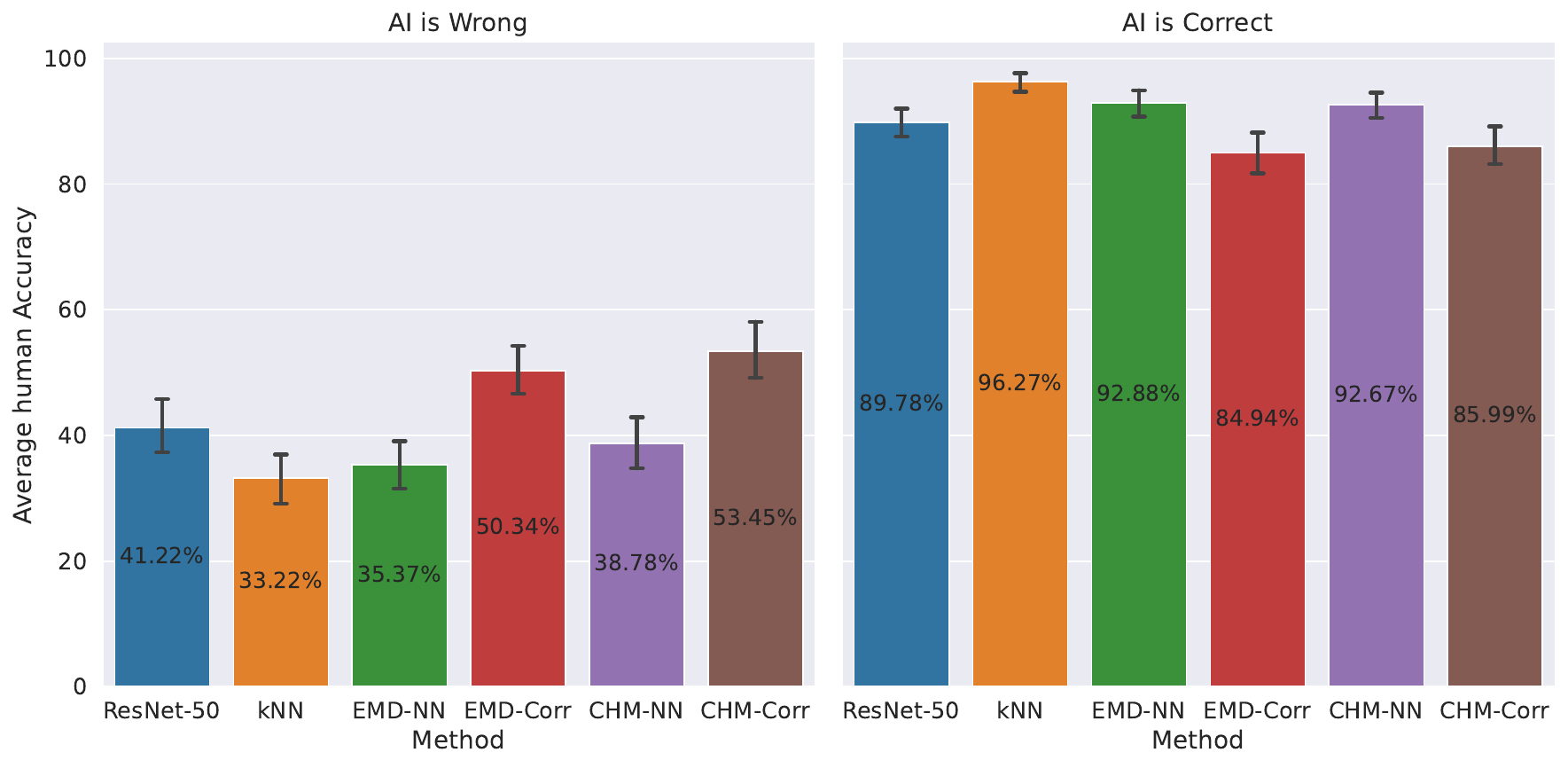}
        \caption{CUB -- Mean user accuracy}
         \label{fig:cub-ai-correctness}
        \end{subfigure}\par
        \caption{The breakdown of human performance by AI correctness}
        \label{fig:breakdown-ai-correctness}
\end{figure}

Figure \ref{fig:breakdown-ai-correctness} shows the breakdown of user accuracy based on the correctness of AI predictions on ImageNet-ReaL and CUB datasets.

\clearpage
\subsection{Human performance analysis based on the difficulty of the query}

In this section, we calculate the average user accuracy based on the difficulty level of the query (described in Sec \ref{supp:defining-levels}) and the correctness of AI's prediction.

\begin{figure}[!htbp]
        \begin{subfigure}[b]{1\textwidth}
                \includegraphics[width=\linewidth]{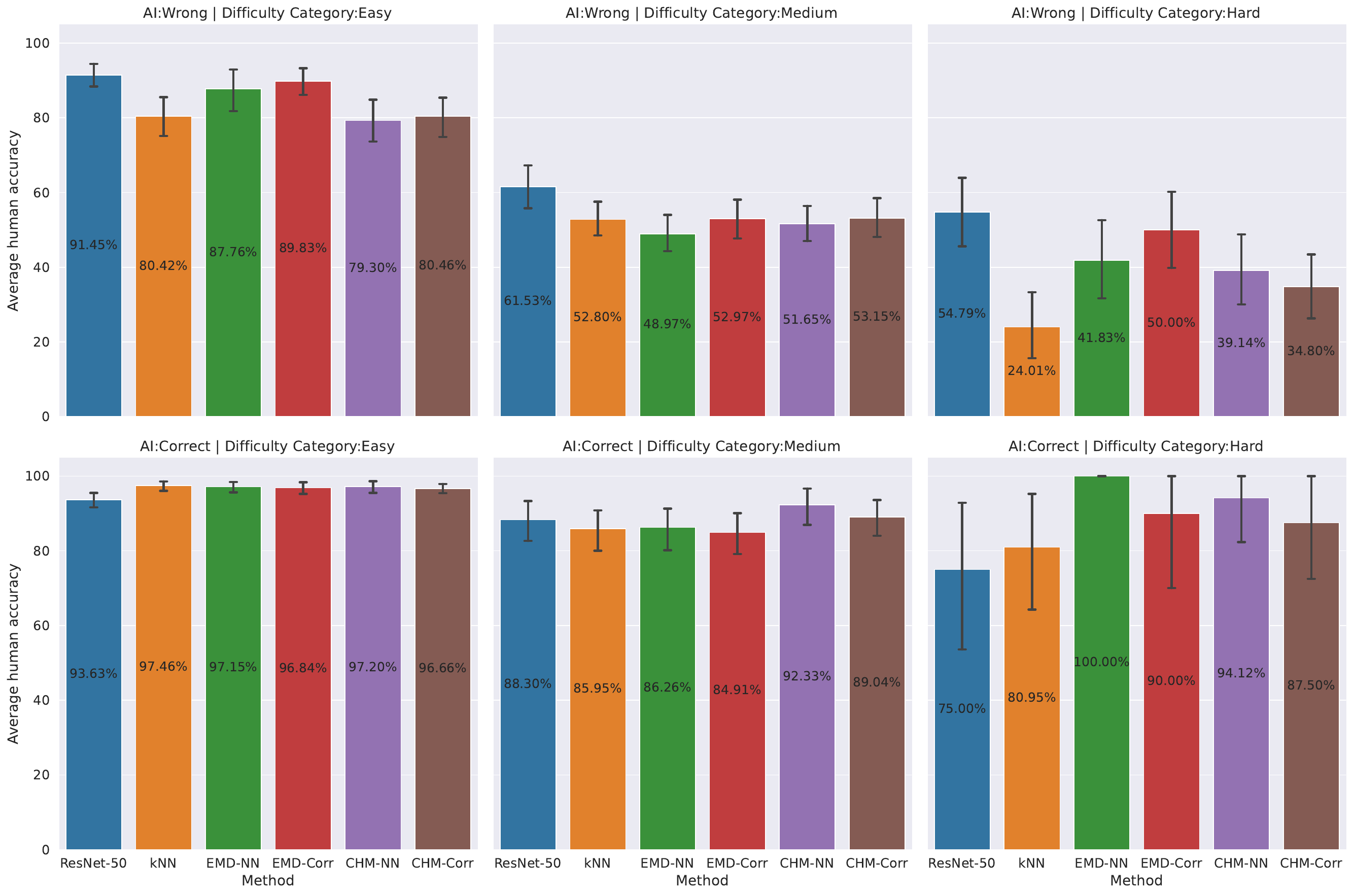}
                \caption{Mean user accuracy}
                \label{fig:imagenet-hardness-correctness}
        \end{subfigure}\par
        \begin{subfigure}[b]{1\textwidth}
                \includegraphics[width=\linewidth]{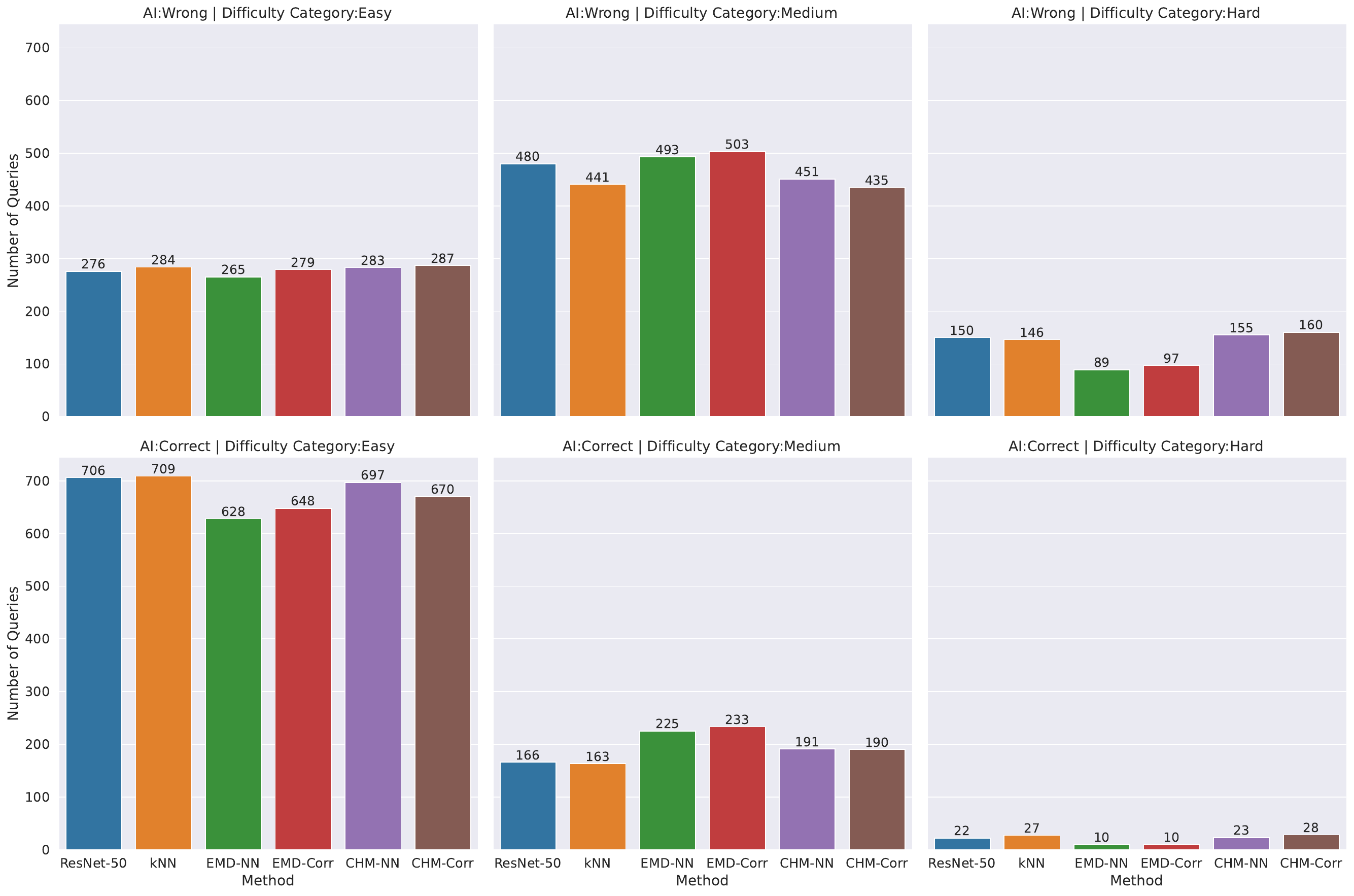}
                \caption{Number of queries}
                \label{fig:imagenet-hardness-correctness-counts}
        \end{subfigure}\par
        \caption{ImageNet -- The breakdown of the human performance by `Difficulty Level' and `AI Correctness'}
        \label{fig:imagenet-difflevel}
\end{figure}

\begin{figure}[!htbp]
        \begin{subfigure}[b]{1\textwidth}
                \includegraphics[width=\linewidth]{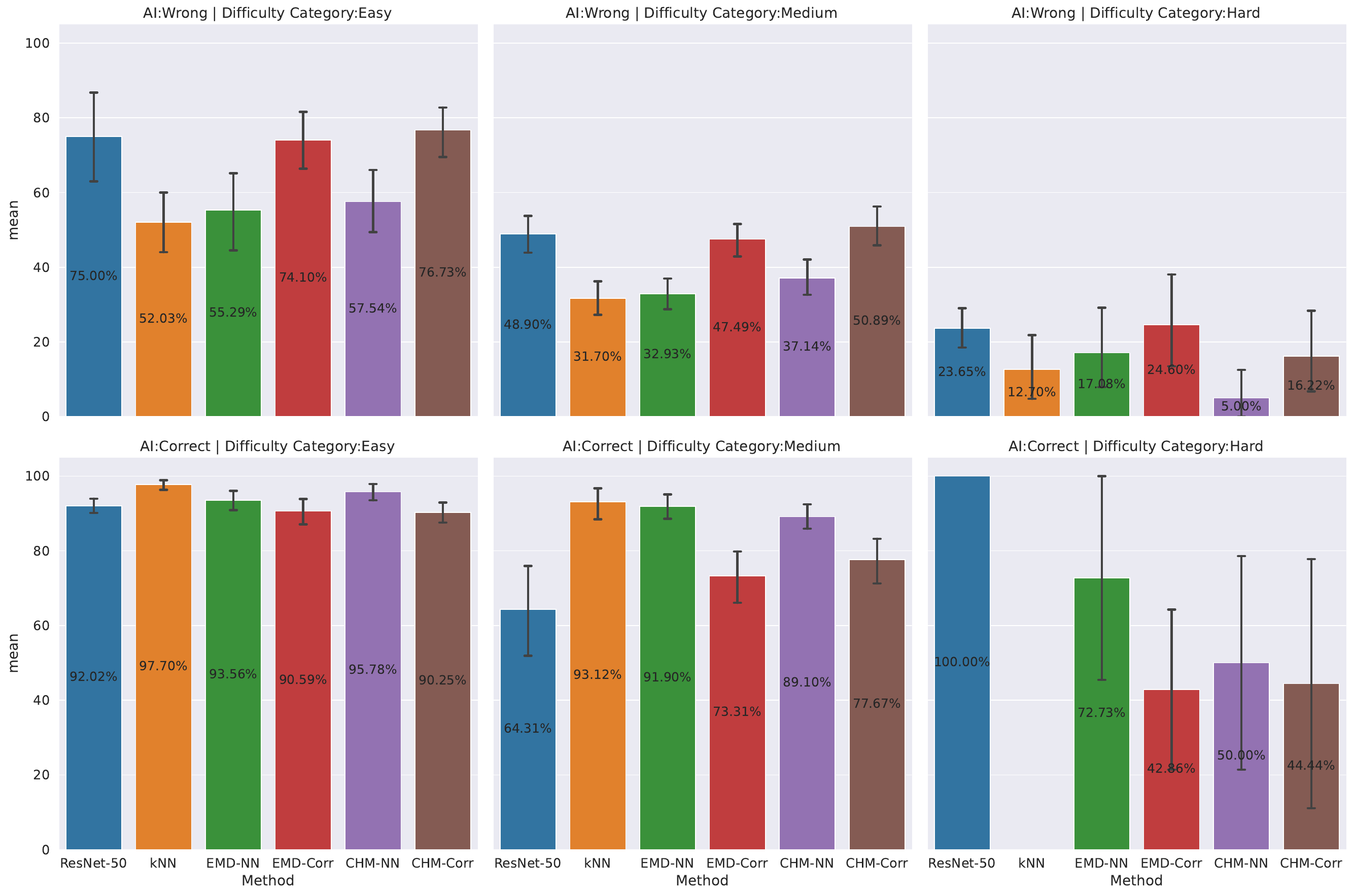}
                 \caption{Mean user accuracy}
                \label{fig:cub-hardness-correctness}
        \end{subfigure}\par
        \begin{subfigure}[b]{1\textwidth}
                \includegraphics[width=\linewidth]{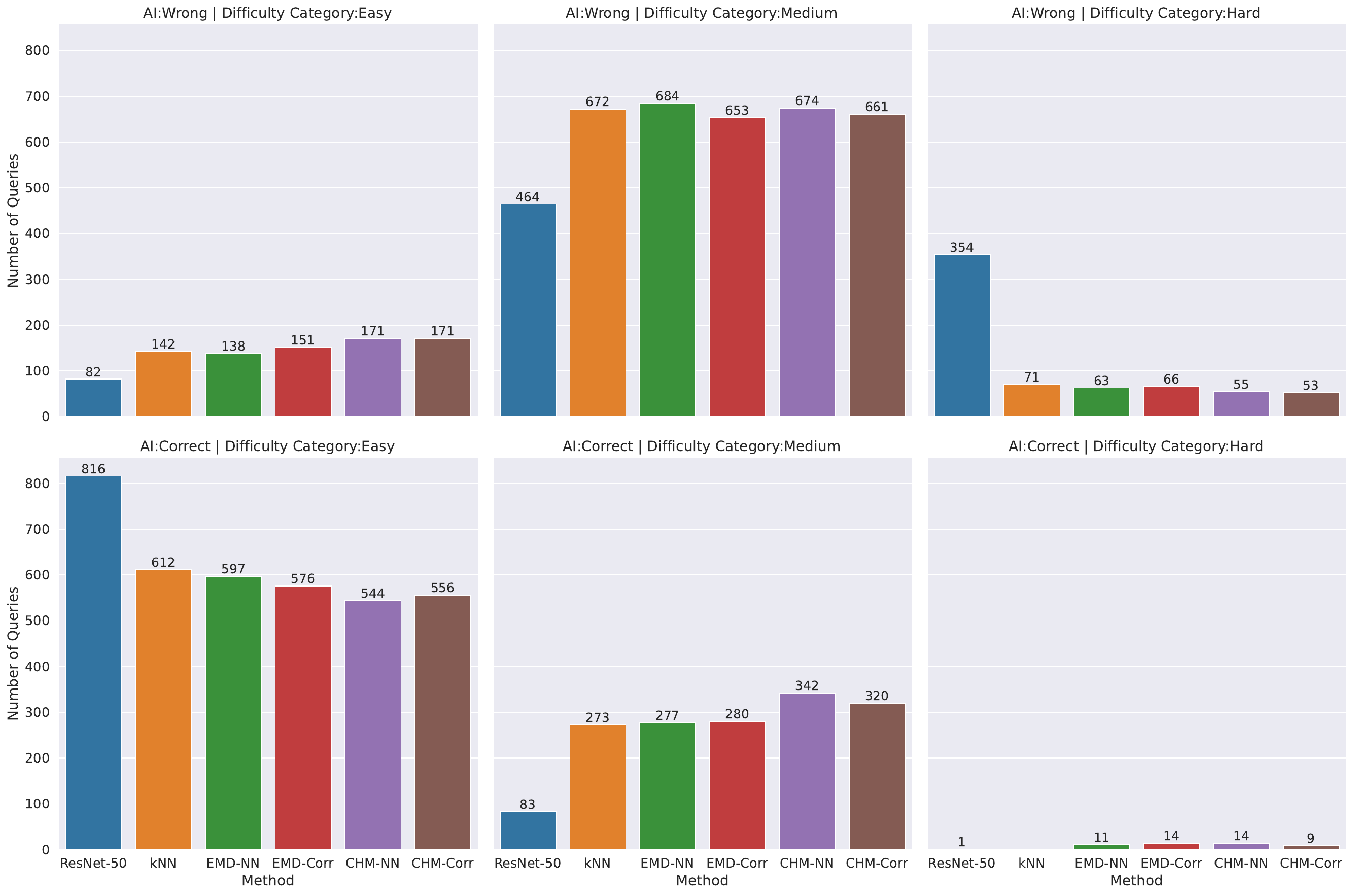}
                 \caption{Number of queries}
                \label{fig:cub-hardness-correctness-counts}
        \end{subfigure}\par
        \caption{CUB -- The breakdown of the human performance by `Difficulty Level' and `AI Correctness'}
        \label{fig:cub-difflevel}
\end{figure}

\clearpage
\subsection{Analysis of Hard images for humans in the ImageNet task}
\label{supp:imagenet_hard_for_human}

\begin{figure}[!htbp]
        \includegraphics[width=\linewidth, trim=2cm .0cm .0cm .0cm,clip]{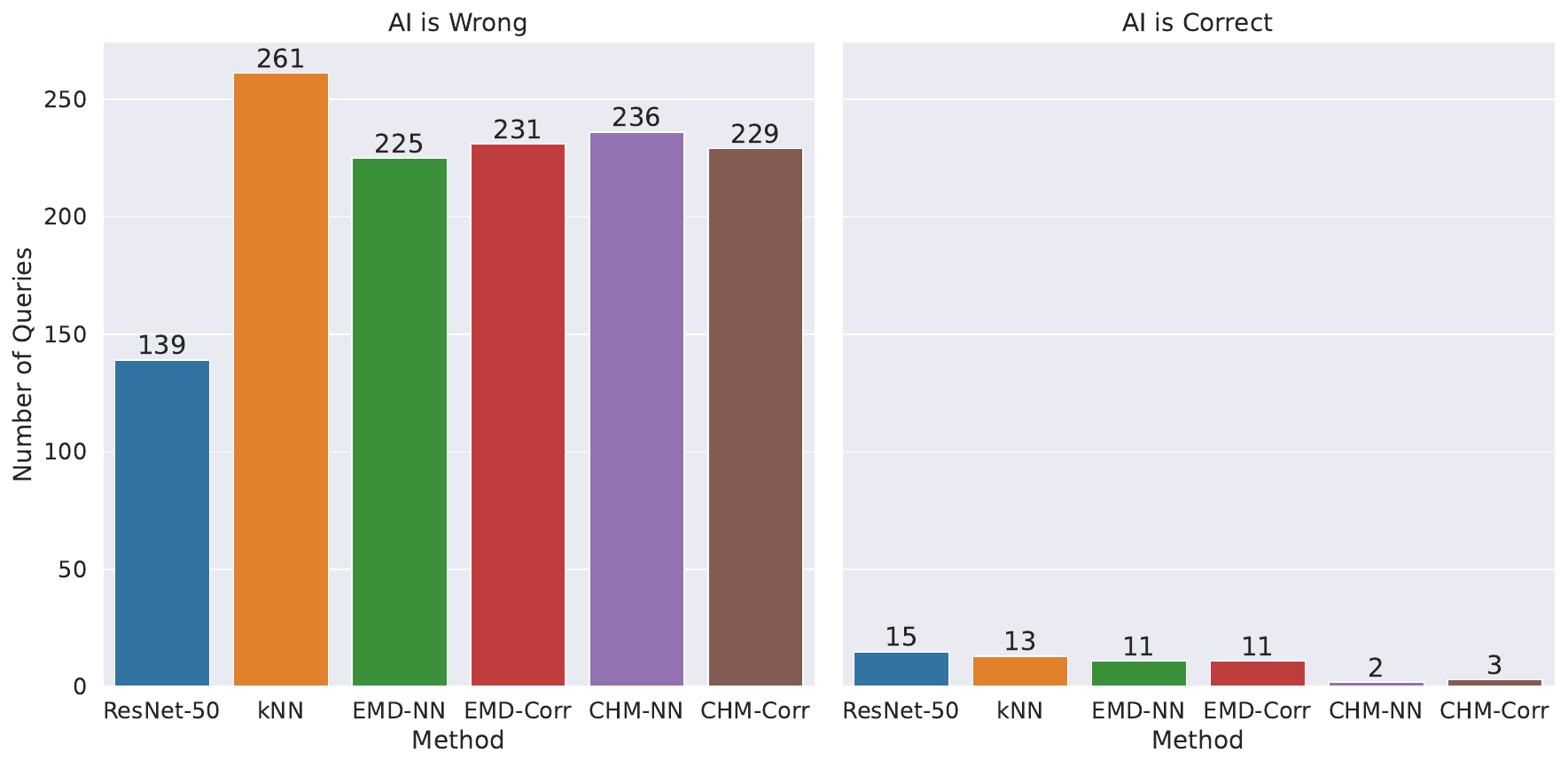}
        \caption{Number of confusing queries per classifier}
        \label{fig:confusing-queries}
\end{figure}

This section provides an analysis to understand what kinds of queries are hard for humans, i.e., for what types of images users cannot correctly accept or reject the AI's decision. To this end, we filter the queries with a mean user's accuracy of $0.25$ or below. Figure \ref{fig:confusing-queries} shows the distribution of Hard images for humans based on the classifier and the classifier's correctness.

To better understand what types of images are more challenging for humans, we automatically create supergroups for ImageNet class members. All 1,000 classes of ImageNet are a subgroup of class \texttt{entity -- n00001740} in the WordNet glossary \cite{miller1995wordnet}. To create groups with uniform sizes, we start from the \texttt{entity} root node and break up each class into its sub-classes recursively; in each iteration, we pick the supergroup with the largest number of classes. Here we report the parent class of queries after $12$ iterations and for queries with only one ImageNet-ReaL label. Using this automated procedure, we can see that the majority of hard images for humans fall into the \texttt{carnivore} category, which is a supergroup for cats, lions, dogs, wolves, etc. Details about each parent class and its ImageNet class members can be found in Table \ref{tab:supergroups}.

\begin{figure}[!htbp]
        \includegraphics[width=\linewidth, trim=0cm .0cm .0cm .0cm,clip]{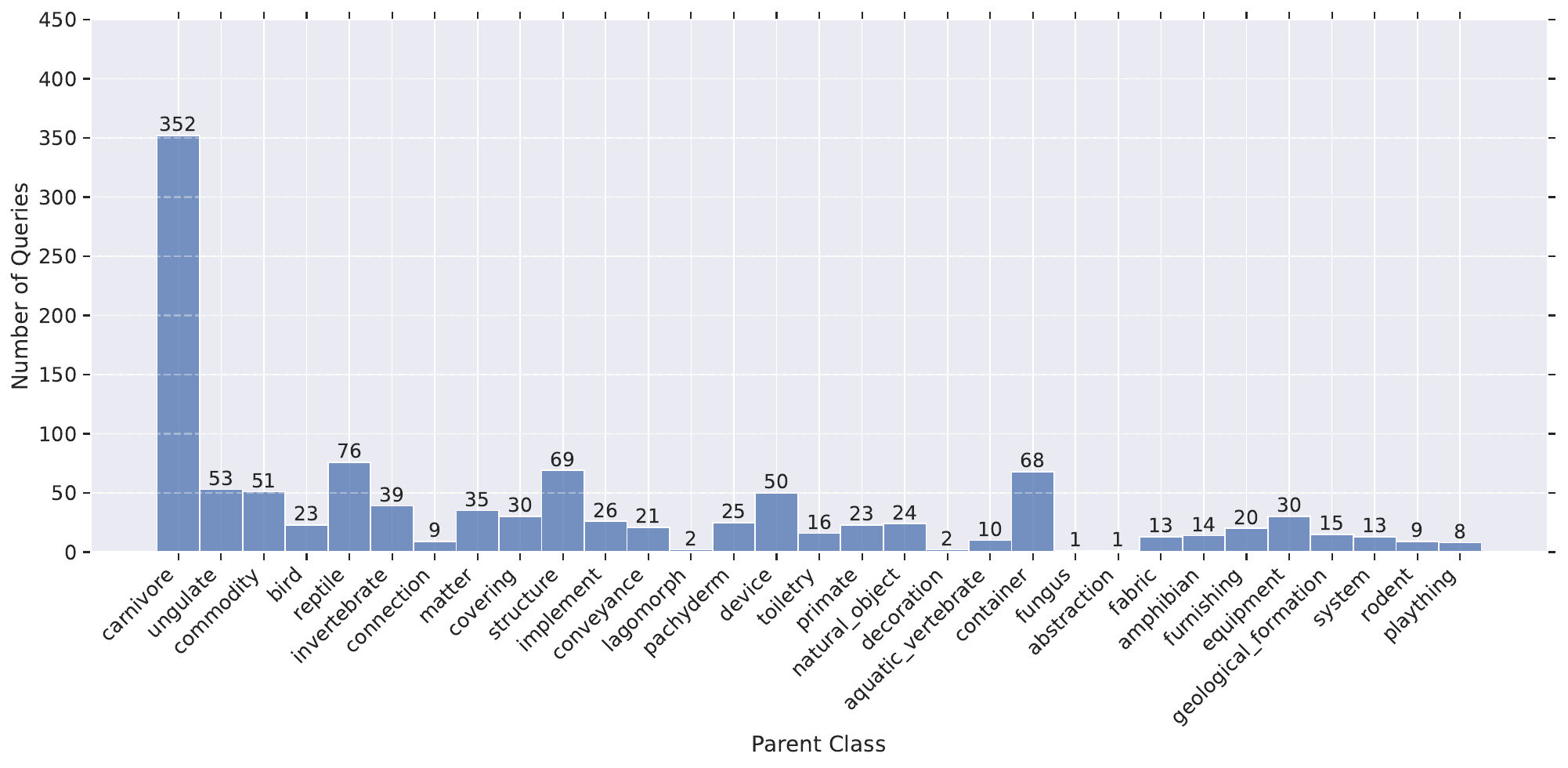}
        \caption{Parent class of confusing queries}
        \label{fig:confusing--queries-parent}
\end{figure}

\begin{table}[]
\centering
\caption{Parent classes of Hard queries for humans and their ImageNet class members.}
\label{tab:supergroups}
\resizebox{\columnwidth}{!}{%
\begin{tabular}{@{}lll@{}}
\toprule
\textbf{Parent Class} & \textbf{ImageNet Class Members} &  \\ \midrule
Abstraction & Street Sign &  \\ \cmidrule(lr){2-2}
Amphibian & Axolotl, European Fire Salamander, Tree Frog &  \\ \cmidrule(lr){2-2}
Aquatic Vertebrate & Goldfish &  \\ \cmidrule(lr){2-2}
Bird & \begin{tabular}[c]{@{}l@{}}American Egret, Bulbul, Cock, Oystercatcher, \\ Red-Backed Sandpiper, Ruffed Grouse\end{tabular} &  \\ \cmidrule(lr){2-2}
Carnivore & \begin{tabular}[c]{@{}l@{}}Beagle, Black-And-Tan Coonhound, Black-Footed Ferret, Bloodhound, \\ Bluetick, Border Terrier, Bouvier Des Flandres, Bull Mastiff, Collie, \\ Coyote, Curly-Coated Retriever, English Foxhound, English Springer, \\ Entlebucher, Eskimo Dog, Flat-Coated Retriever, French Bulldog, \\ German Short-Haired Pointer, Great Dane, Great Pyrenees, \\ Greater Swiss Mountain Dog, Irish Water Spaniel, Kelpie, Lakeland Terrier, \\ Lhasa, Malamute, Miniature Poodle, Norfolk Terrier, Otter, Pembroke, \\ Polecat, Redbone, Rhodesian Ridgeback, Rottweiler, Schipperke, \\ Scotch Terrier, Scottish Deerhound, Sealyham Terrier, Shetland Sheepdog, \\ Siberian Husky, Staffordshire Bullterrier, Standard Poodle,  Standard Schnauzer, \\ Tabby, Tibetan Terrier, Tiger Cat, Toy Poodle, Vizsla, Welsh Springer Spaniel, Yorkshire Terrier\end{tabular} &  \\ \cmidrule(lr){2-2}
Commodity & \begin{tabular}[c]{@{}l@{}}Abaya, Academic Gown, Dishwasher, Dutch Oven, Espresso Maker, \\ Microwave, Military Uniform, Miniskirt, Washer\end{tabular} &  \\ \cmidrule(lr){2-2}
Connection & Chain &  \\ \cmidrule(lr){2-2}
Container & \begin{tabular}[c]{@{}l@{}}Ambulance, Cassette, Envelope, Pitcher, Purse, Shopping Basket, \\ Soap Dispenser, Soup Bowl, Tank, Wallet, Washbasin, Whiskey Jug\end{tabular} &  \\ \cmidrule(lr){2-2}
Conveyance & Dogsled, Schooner, Stretcher, Trolleybus, Yawl &  \\ \cmidrule(lr){2-2}
Covering & Dome, Doormat, Pickelhaube, Prayer Rug, Scabbard, Shower Curtain &  \\ \cmidrule(lr){2-2}
Decoration & Necklace &  \\ \cmidrule(lr){2-2}
Device & \begin{tabular}[c]{@{}l@{}}Analog Clock, Car Wheel, Cello, Combination Lock, \\ Padlock, Projectile, Radiator, Upright, Wall Clock\end{tabular} &  \\ \cmidrule(lr){2-2}
Equipment & Balance Beam, Cd Player, Dumbbell, Horizontal Bar, Monitor, Polaroid Camera &  \\ \cmidrule(lr){2-2}
Fabric & Wool &  \\ \cmidrule(lr){2-2}
Fungus & Hen-Of-The-Woods &  \\ \cmidrule(lr){2-2}
Furnishing & Cradle, Crib, Desk, Entertainment Center &  \\ \cmidrule(lr){2-2}
Geological Formation & Coral Reef, Lakeside, Seashore &  \\ \cmidrule(lr){2-2}
Implement & Ballpoint, Plow, Plunger, Quill, Teapot &  \\ \cmidrule(lr){2-2}
Invertebrate & \begin{tabular}[c]{@{}l@{}}Barn Spider, Bee, Cricket, Damselfly, Dragonfly, Dungeness Crab, \\ Long-Horned Beetle, Mantis, Sea Slug, Snail, Sulphur Butterfly\end{tabular} &  \\ \cmidrule(lr){2-2}
Lagomorph & Wood Rabbit &  \\ \cmidrule(lr){2-2}
Matter & Artichoke, Bell Pepper, Cheeseburger, Cucumber, Hay, Plate, Pretzel &  \\ \cmidrule(lr){2-2}
Natural Object & Banana, Corn, Lemon, Sandbar &  \\ \cmidrule(lr){2-2}
Pachyderm & African Elephant, Indian Elephant &  \\ \cmidrule(lr){2-2}
Plaything & Teddy &  \\ \cmidrule(lr){2-2}
Primate & Gibbon, Gorilla, Langur, Siamang, Titi &  \\ \cmidrule(lr){2-2}
Reptile & \begin{tabular}[c]{@{}l@{}}African Crocodile, Alligator Lizard, American Chameleon, \\ Banded Gecko, Boa Constrictor, Frilled Lizard, Green Mamba,\\ Green Snake, Mud Turtle, Night Snake, Rock Python, Terrapin\end{tabular} &  \\ \cmidrule(lr){2-2}
Rodent & Beaver &  \\ \cmidrule(lr){2-2}
Structure & \begin{tabular}[c]{@{}l@{}}Bakery, Bannister, Church, Cliff Dwelling, Dam, Dock, \\ Grocery Store, Megalith, Plate Rack, Stupa, Totem Pole\end{tabular} &  \\ \cmidrule(lr){2-2}
System & Radio &  \\ \cmidrule(lr){2-2}
Toiletry & Hair Spray, Lotion, Sunscreen &  \\ \cmidrule(lr){2-2}
Ungulate & Bighorn, Bison, Hog, Impala, Llama, Water Buffalo, Wild Boar &  \\ \bottomrule
\end{tabular}%
}
\end{table}

\clearpage
\subsection{Analysis of Hard images for humans in the bird classification task}
\label{supp:cub_hard_for_human}

\begin{figure}[!htbp]
        \includegraphics[width=\linewidth]{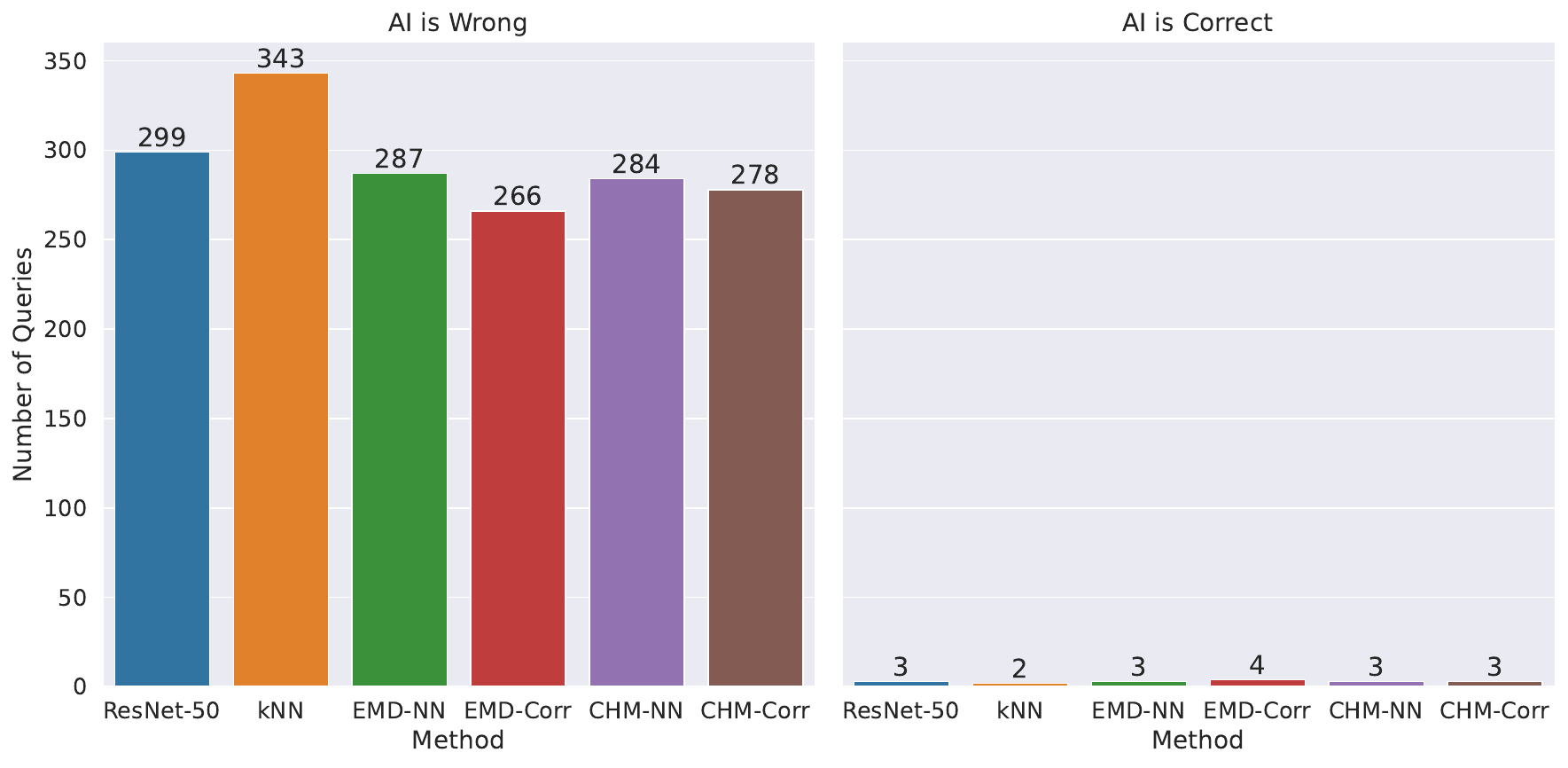}
        \caption{Number of confusing queries per classifier}
        \label{fig:cub-accuracy-correctness}
\end{figure}

Similar to the analysis we conducted for ImageNet in Sec.\ref{supp:imagenet_hard_for_human}, here we analyze the confusing bird types for humans. We filter queries with a mean user accuracy of less than $0.25$. Figure \ref{fig:cub-accuracy-correctness} shows the distribution of the most challenging samples for humans based on different classifiers' correctness.

Table \ref{tab:confusionmatrix-birds} shows the top 5 confusing bird types for human users. Each row in this table shows how many users failed to reject AI's prediction when providing different kinds of explanations.

\begin{table}[]
\centering
\caption{Top-5 confusing bird types for humans per classifier}
\label{tab:confusionmatrix-birds}
\begin{tabular}{@{}cllc@{}}
\toprule
\multicolumn{1}{l}{\textbf{Classifier}} & \multicolumn{1}{c}{\textbf{Ground Truth}} & \multicolumn{1}{c}{\textbf{Confused with}} & \textbf{Users} \\ \midrule
\multirow{5}{*}{ResNet-50} & Forsters Tern & Common Tern & 17 \\
 & Great Grey Shrike & Loggerhead Shrike & 10 \\
 & Nelson Sharp Tailed Sparrow & Le Conte Sparrow & 8 \\
 & Acadian Flycatcher & Least Flycatcher & 7 \\
 & American Crow & Common Raven & 7 \\ \midrule
\multirow{5}{*}{kNN} & California Gull & Western Gull & 21 \\
 & Elegant Tern & Caspian Tern & 13 \\
 & Fish Crow & American Crow & 12 \\
 & Rusty Blackbird & Brewer Blackbird & 11 \\
 & Acadian Flycatcher & Yellow Bellied Flycatcher & 10 \\ \midrule
\multirow{5}{*}{EMD-NN} & California Gull & Western Gull & 15 \\
 & Common Tern & Artic Tern & 12 \\
 & Nelson Sharp Tailed Sparrow & Savannah Sparrow & 8 \\
 & Acadian Flycatcher & Yellow Bellied Flycatcher & 8 \\
 & Yellow Bellied Flycatcher & Acadian Flycatcher & 8 \\ \midrule
\multirow{5}{*}{EMD-Corr} & California Gull & Western Gull & 15 \\
 & Common Tern & Artic Tern & 12 \\
 & Nelson Sharp Tailed Sparrow & Savannah Sparrow & 8 \\
 & Acadian Flycatcher & Yellow Bellied Flycatcher & 8 \\
 & Yellow Bellied Flycatcher & Acadian Flycatcher & 8 \\ \midrule
\multirow{5}{*}{CHM-NN} & Great Grey Shrike & Loggerhead Shrike & 19 \\
 & Le Conte Sparrow & Nelson Sharp Tailed Sparrow & 15 \\
 & California Gull & Western Gull & 15 \\
 & Louisiana Waterthrush & Northern Waterthrush & 12 \\
 & Horned Grebe & Eared Grebe & 9 \\ \midrule
\multirow{5}{*}{CHM-Corr} & Great Grey Shrike & Loggerhead Shrike & 20 \\
 & Horned Grebe & Eared Grebe & 15 \\
 & California Gull & Western Gull & 15 \\
 & Louisiana Waterthrush & Northern Waterthrush & 13 \\
 & Le Conte Sparrow & Nelson Sharp Tailed Sparrow & 13 \\ \bottomrule
\end{tabular}%
\end{table}

\clearpage
\subsection{Accepting AI's wrong decision}

This section shows samples for which users incorrectly accepted the incorrect AI prediction.

\subsubsection{Accepting the wrong kNN Classifier's prediction}
\begin{figure}[!htbp]
        \begin{subfigure}[b]{1\textwidth}
                \includegraphics[width=\linewidth, trim=1cm .0cm .0cm .0cm,clip]{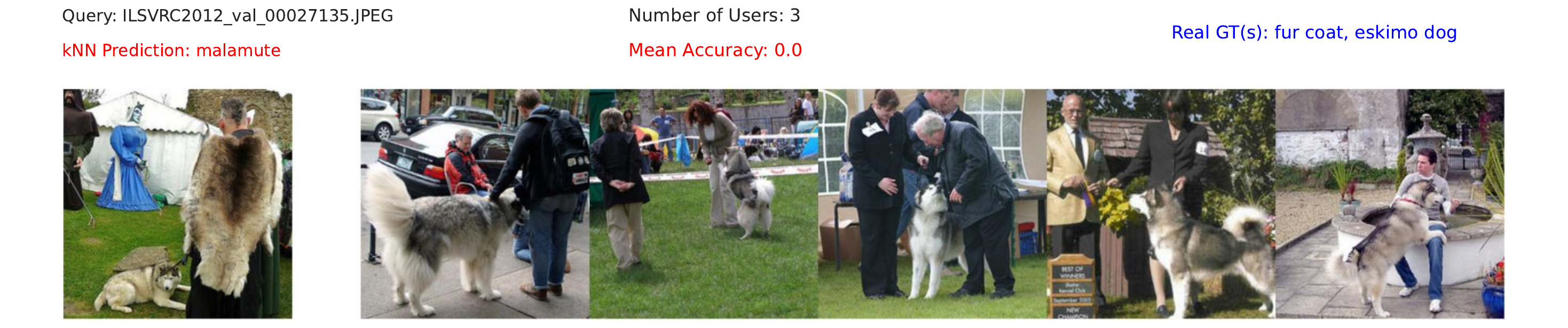}
        \end{subfigure}\par
        \begin{subfigure}[b]{1\textwidth}
                \includegraphics[width=\linewidth, trim=1cm .0cm .0cm .0cm,clip]{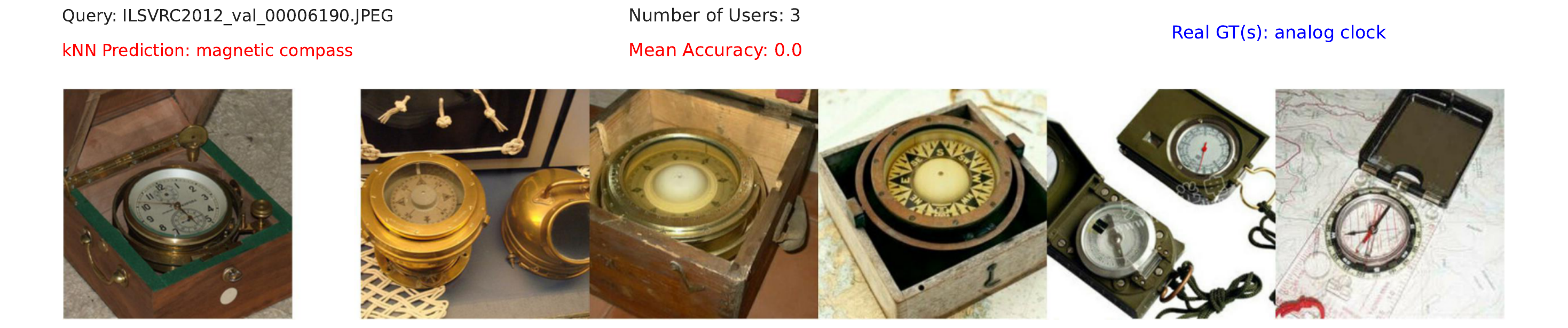}
        \end{subfigure}\par
         \begin{subfigure}[b]{1\textwidth}
                \includegraphics[width=\linewidth, trim=1cm .0cm .0cm .0cm,clip]{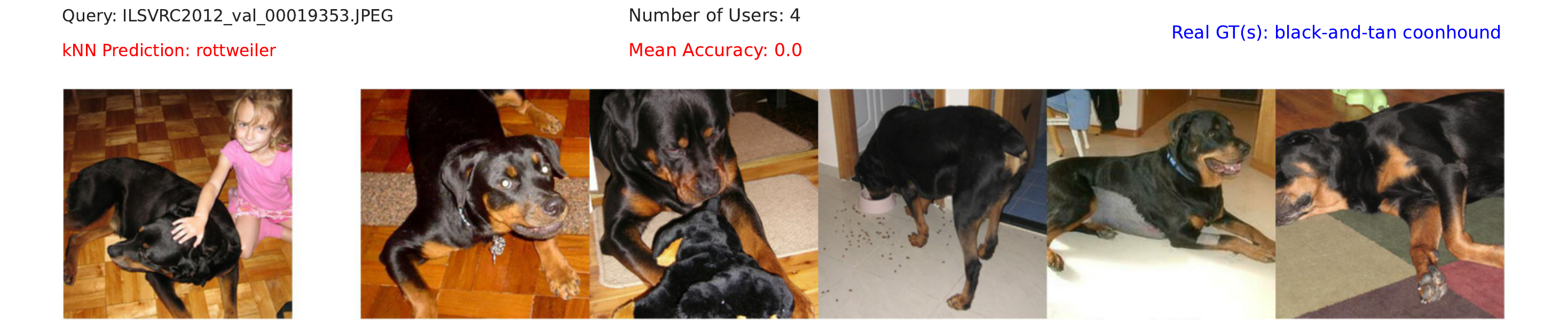}
        \end{subfigure}\par
        \caption{Accepting the wrong kNN prediction due to confusing explanations}
        \label{fig:accepting_wrong_ai_knn}
\end{figure}

\begin{figure}[!htbp]

    \begin{subfigure}[b]{1\textwidth}
        \includegraphics[width=\linewidth, trim=1cm .0cm .0cm .0cm,clip]{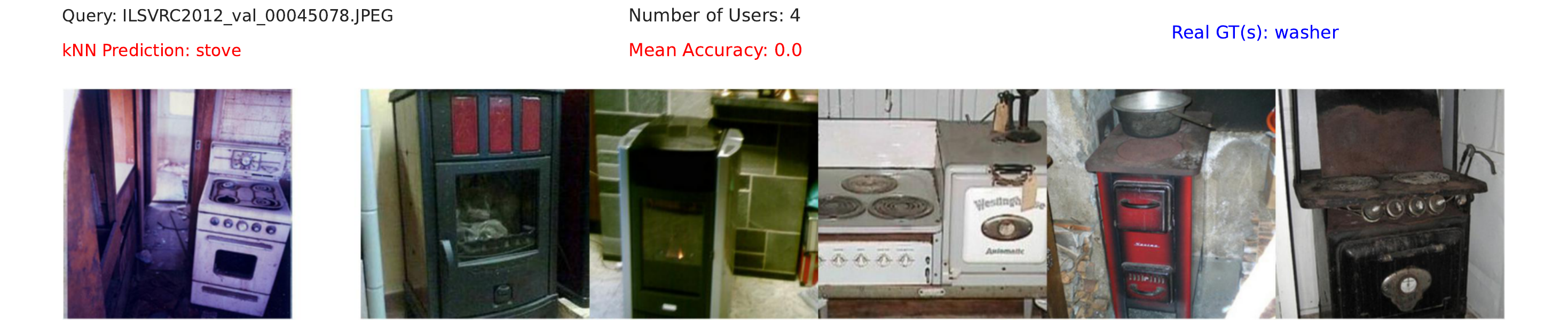}
    \end{subfigure}\par
    \begin{subfigure}[b]{1\textwidth}
        \includegraphics[width=\linewidth, trim=1cm .0cm .0cm .0cm,clip]{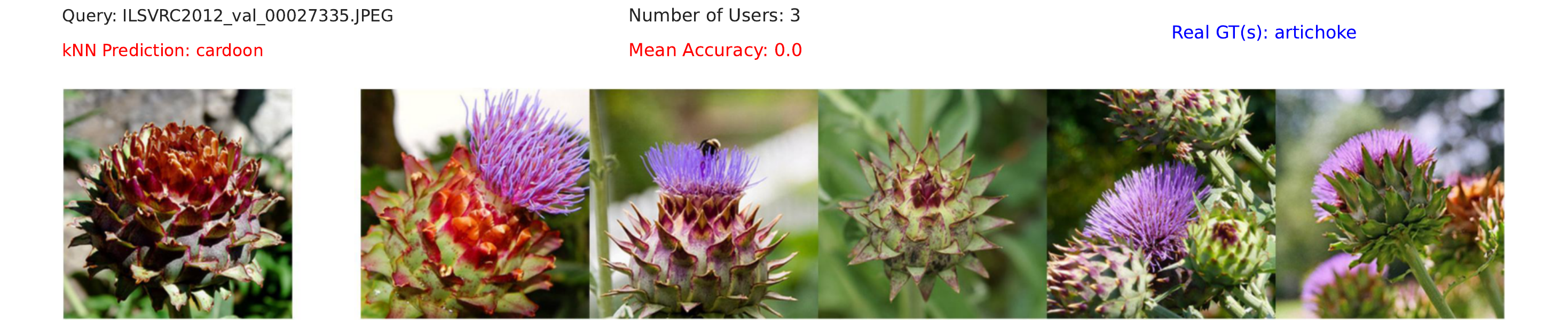}
    \end{subfigure}\par
    \begin{subfigure}[b]{1\textwidth}
        \includegraphics[width=\linewidth, trim=1cm .0cm .0cm .0cm,clip]{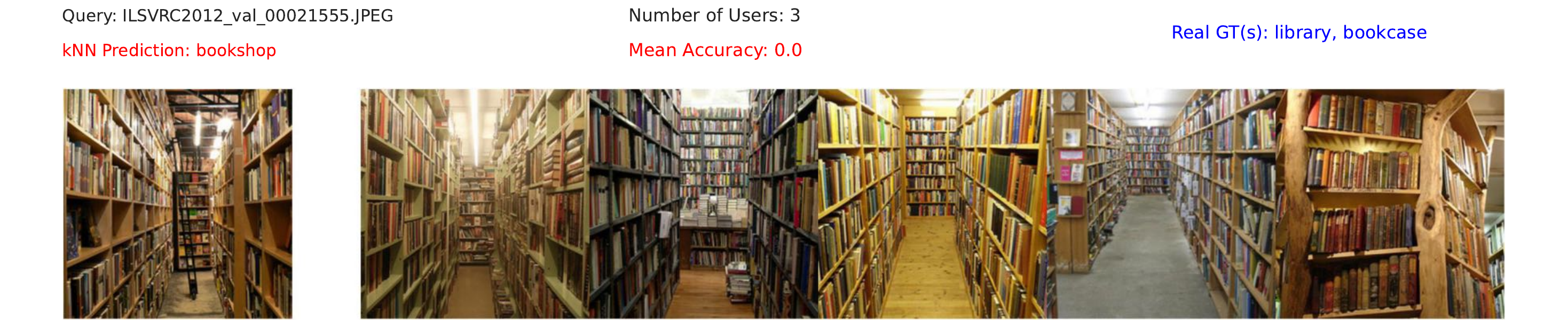}
    \end{subfigure}\par
        
        \caption{Accepting the wrong kNN prediction due to poor ImageNet-ReaL labeling}
        \label{fig:accepting_wrong_ai_knn_bad_labels}
\end{figure}

\clearpage
\subsubsection{Accepting the wrong EMD-NN Classifier's prediction}

\begin{figure}[!htbp]
        \begin{subfigure}[b]{1\textwidth}
        \includegraphics[width=\linewidth, trim=1cm .0cm .0cm .0cm,clip]{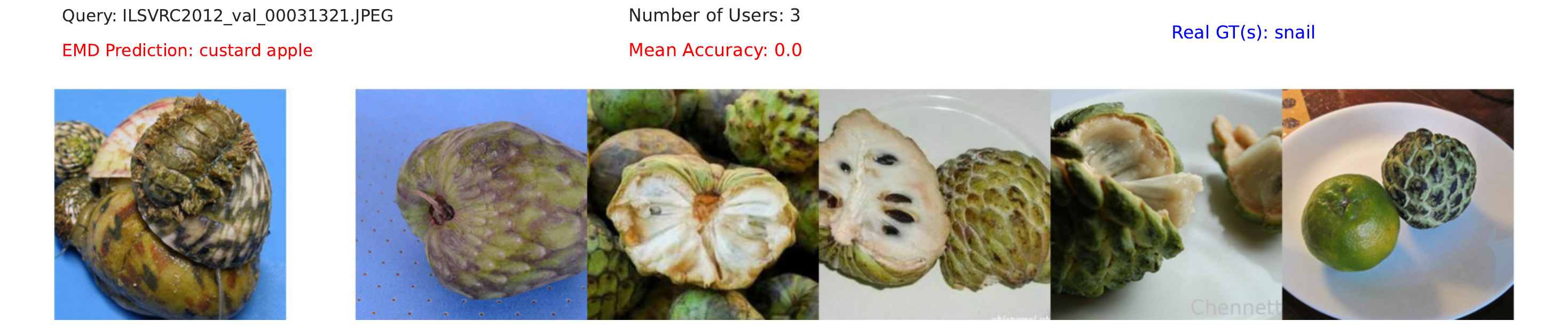}
        \end{subfigure}\par
         \begin{subfigure}[b]{1\textwidth}
        \includegraphics[width=\linewidth, trim=1cm .0cm .0cm .0cm,clip]{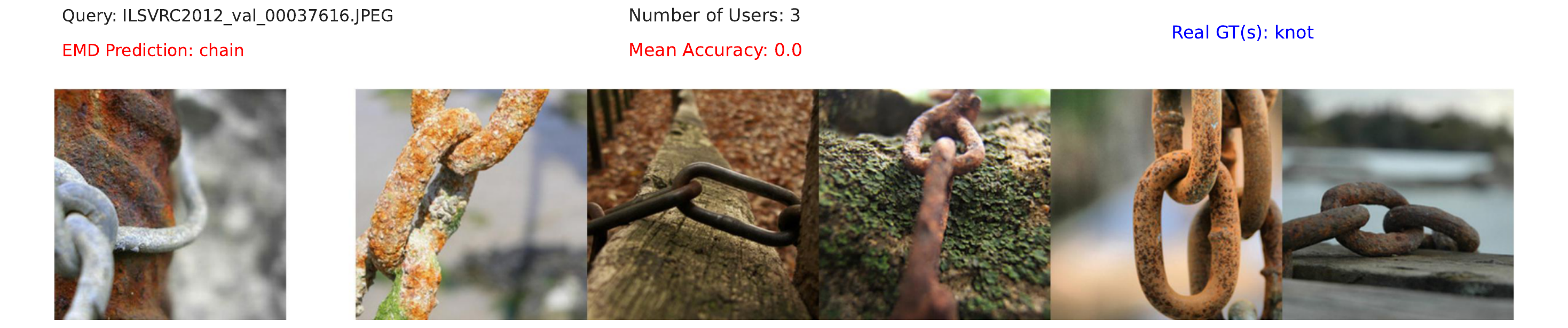}
        \end{subfigure}\par
        \begin{subfigure}[b]{1\textwidth}
        \includegraphics[width=\linewidth, trim=1cm .0cm .0cm .0cm,clip]{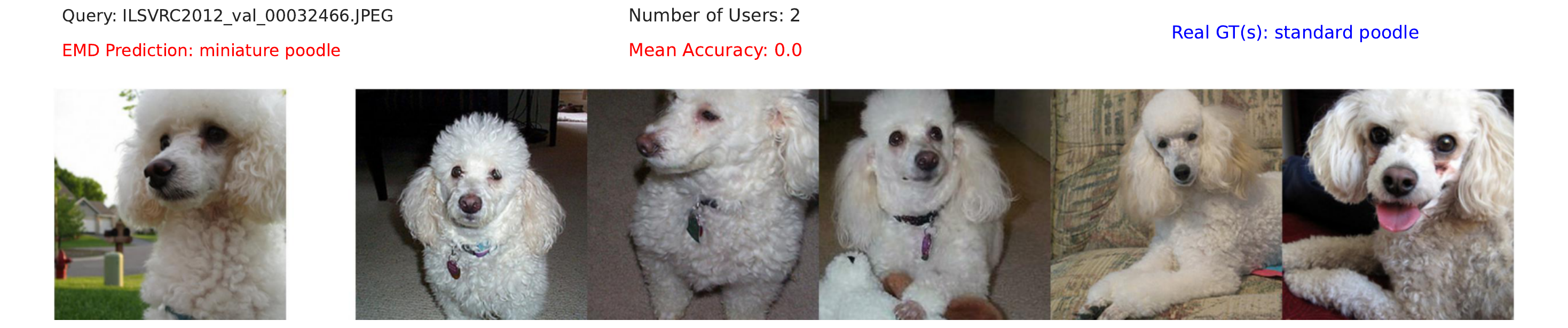}
        \end{subfigure}\par
        \caption{Accepting the wrong EMD-NN prediction due to confusing explanations}
        \label{fig:accepting_wrong_ai_EMD_nn}
\end{figure}

\begin{figure}[!htbp]
        \begin{subfigure}[b]{1\textwidth}
        \includegraphics[width=\linewidth, trim=1cm .0cm .0cm .0cm,clip]{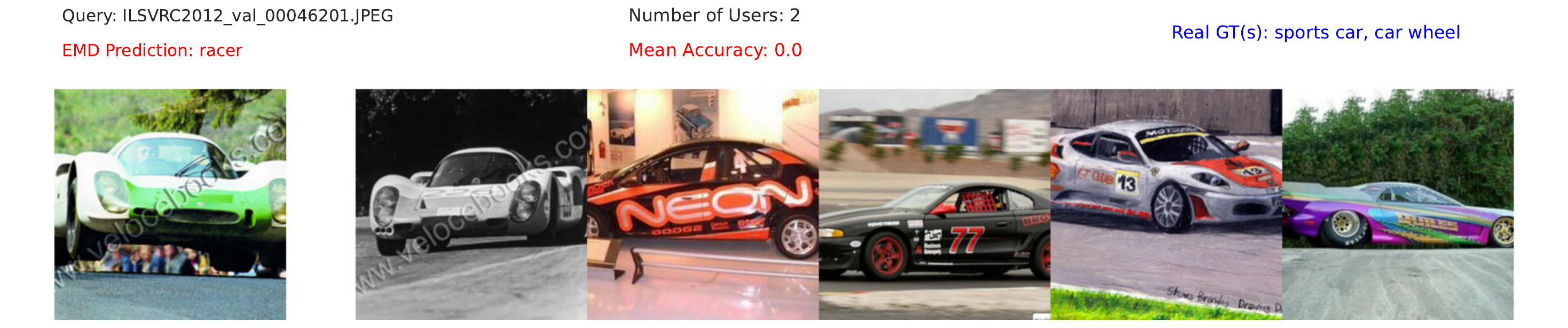} 
        \end{subfigure}\par
        \begin{subfigure}[b]{1\textwidth}
        \includegraphics[width=\linewidth, trim=1cm .0cm .0cm .0cm,clip]{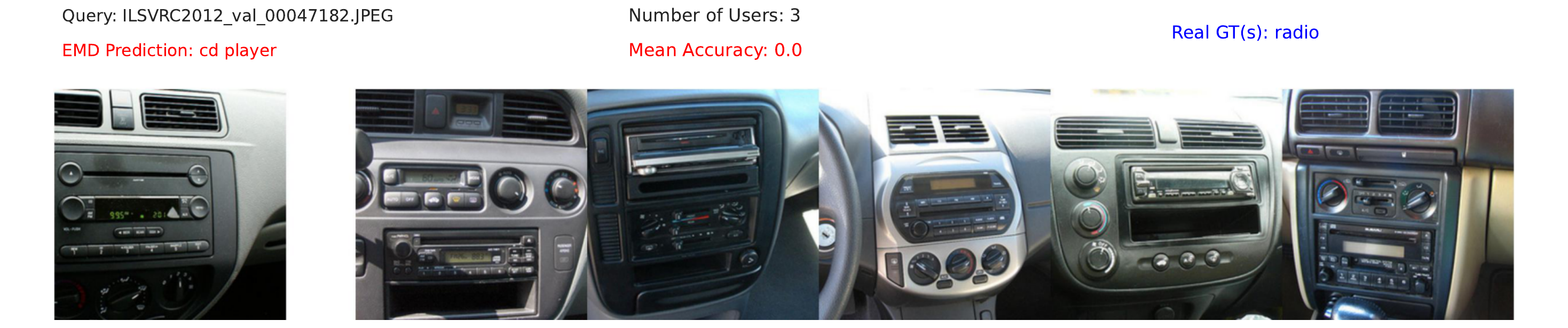} 
        \end{subfigure}\par
        
        \begin{subfigure}[b]{1\textwidth}
        \includegraphics[width=\linewidth, trim=1cm .0cm .0cm .0cm,clip]{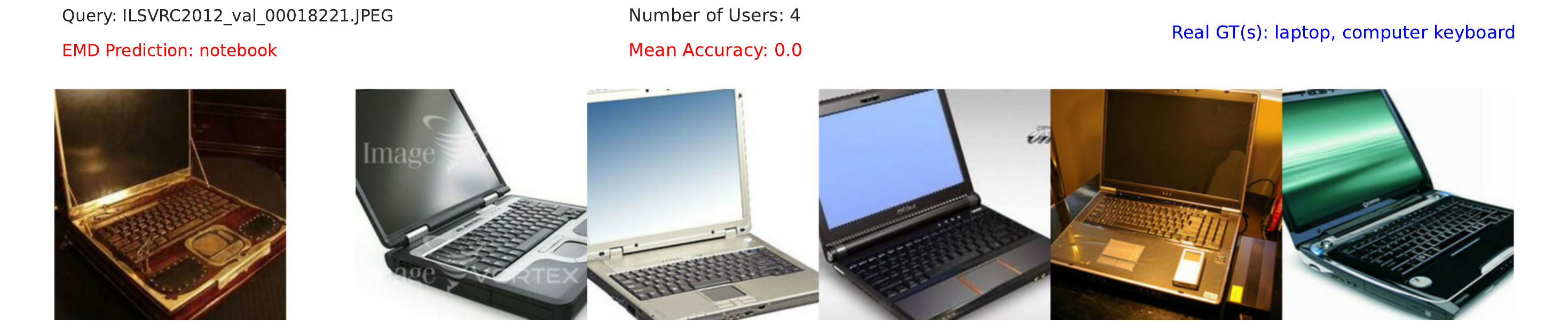} 
        \end{subfigure}\par

        
        \caption{Accepting wrong EMD-NN prediction due to 'Bad Labels'}
        \label{fig:accepting_wrong_ai_EMD_nn_bad_labels}
\end{figure}

\clearpage
\subsubsection{Accepting the wrong EMD-Corr Classifier's prediction}

\begin{figure}[!htbp]
    \begin{subfigure}[b]{1\textwidth}
        \includegraphics[width=\linewidth]{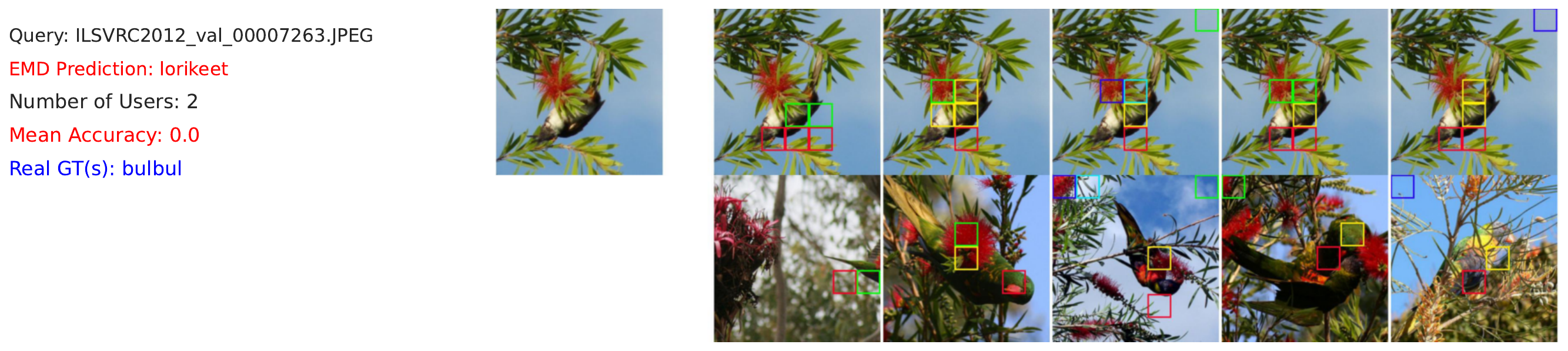}
    \end{subfigure}\par
    \begin{subfigure}[b]{1\textwidth}
        \includegraphics[width=\linewidth]{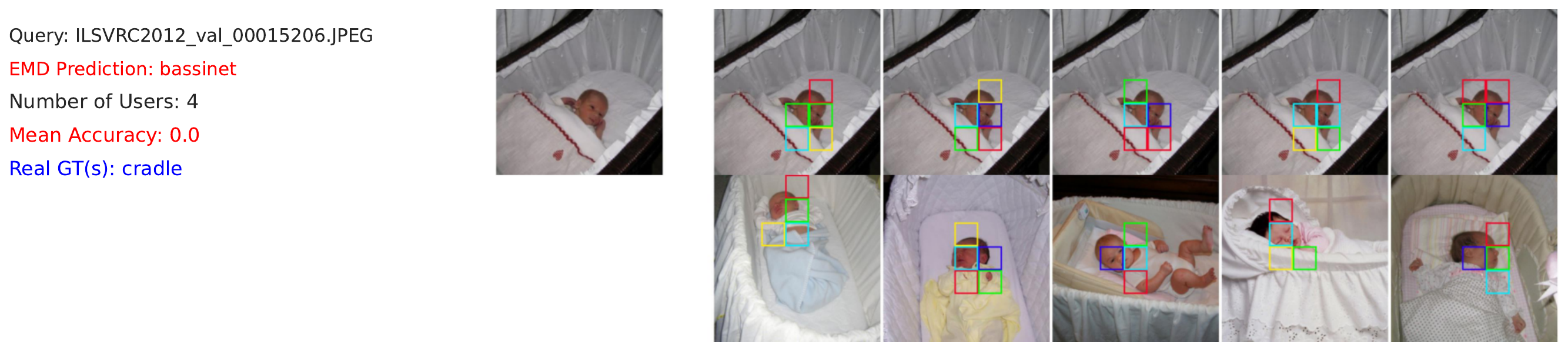}
    \end{subfigure}\par
    \begin{subfigure}[b]{1\textwidth}
        \includegraphics[width=\linewidth]{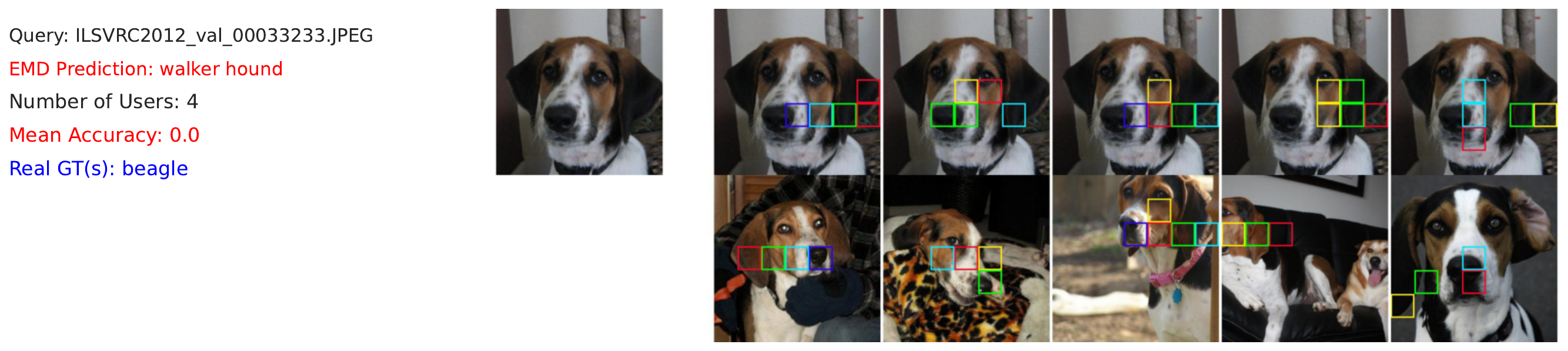}
    \end{subfigure}\par
    \caption{Accepting wrong EMD-Corr prediction due to confusing explanations}
    \label{fig:accepting_wrong_ai_emd_corr}
\end{figure}

\begin{figure}[!htbp]
    
    \begin{subfigure}[b]{1\textwidth}
        \includegraphics[width=\linewidth]{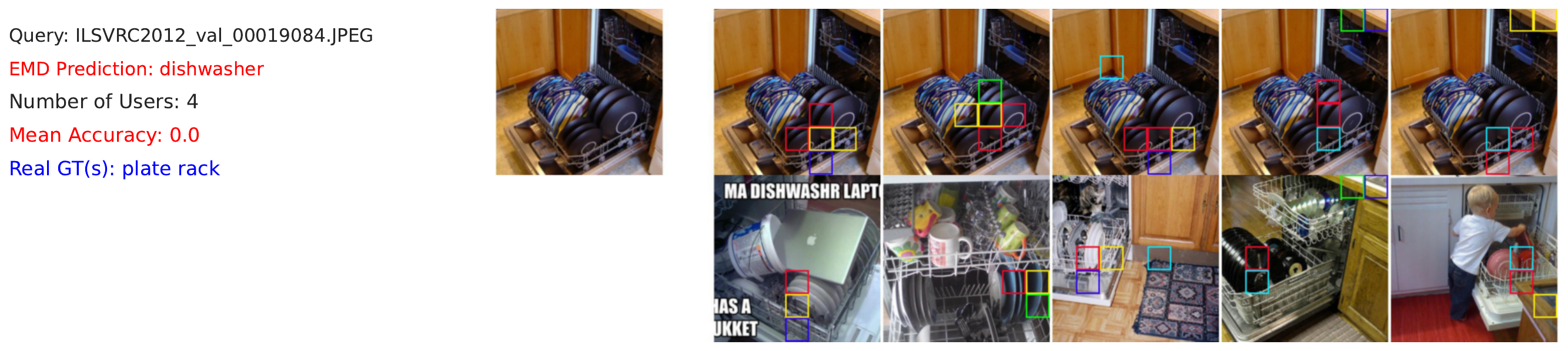}
    \end{subfigure}\par
    \begin{subfigure}[b]{1\textwidth}
        \includegraphics[width=\linewidth]{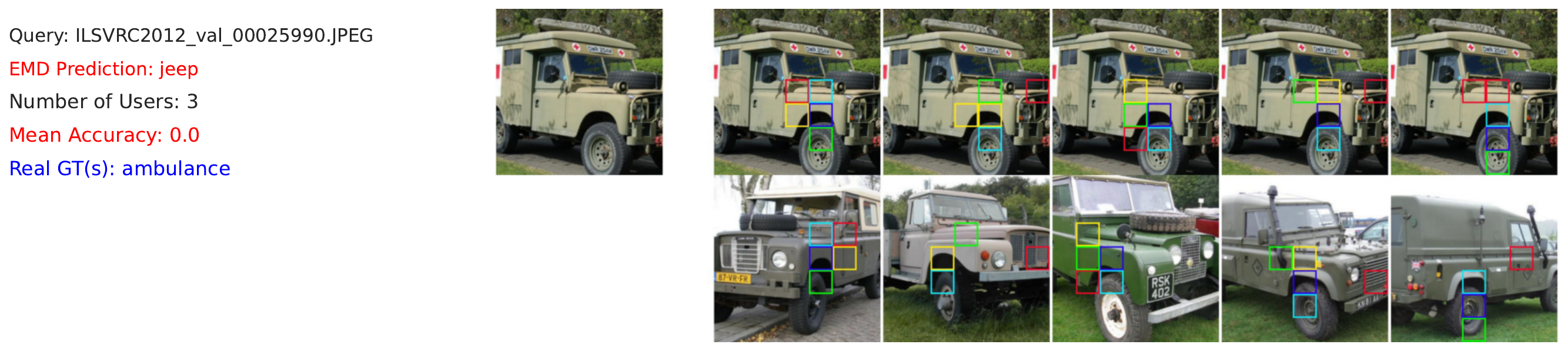}
    \end{subfigure}\par
   \begin{subfigure}[b]{1\textwidth}
        \includegraphics[width=\linewidth]{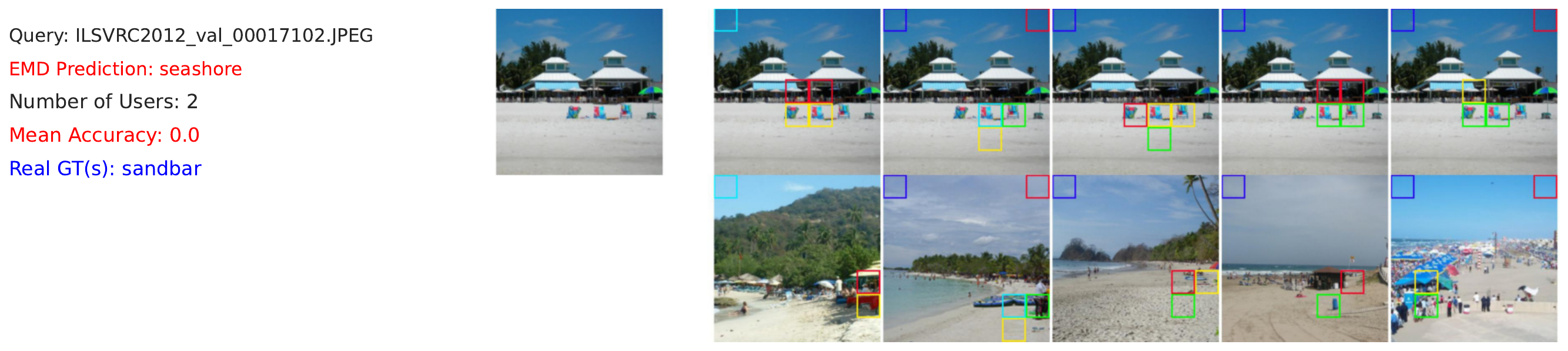}
    \end{subfigure}\par

    
    \caption{Accepting wrong EMD-Corr prediction due to poor ImageNet-ReaL labeling}
    \label{fig:accepting_wrong_ai_emd_corr_badlables}
\end{figure}

\clearpage
\subsubsection{Accepting the wrong CHM-NN Classifier's prediction}

\begin{figure}[!htbp]

\begin{subfigure}[b]{1\textwidth}
\includegraphics[width=\linewidth, trim=1cm .0cm .0cm .0cm,clip]{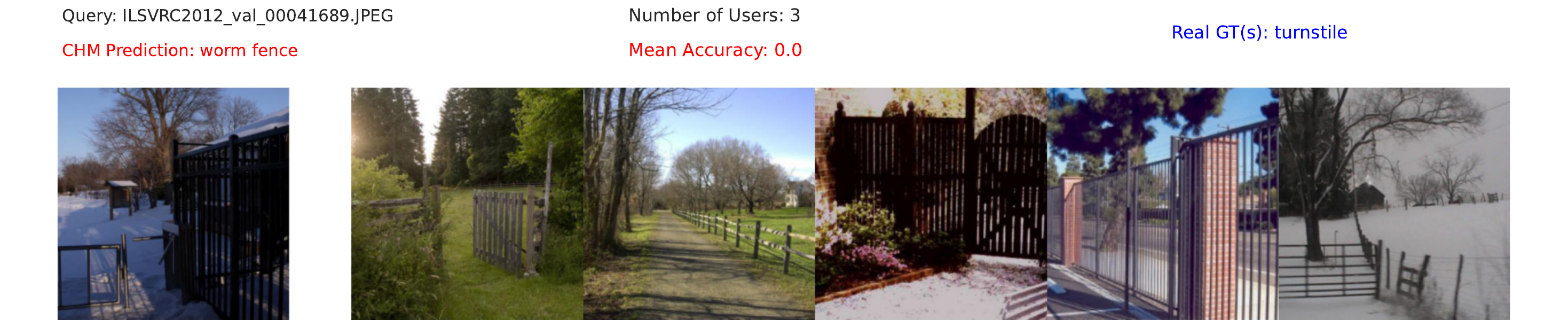}
\end{subfigure}\par

\begin{subfigure}[b]{1\textwidth}
\includegraphics[width=\linewidth, trim=1cm .0cm .0cm .0cm,clip]{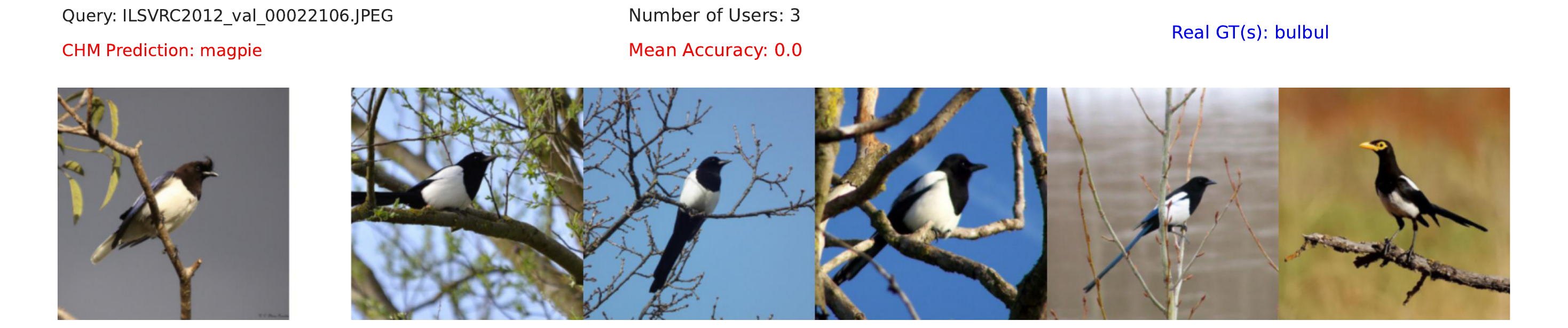}
\end{subfigure}\par

\begin{subfigure}[b]{1\textwidth}
\includegraphics[width=\linewidth, trim=1cm .0cm .0cm .0cm,clip]{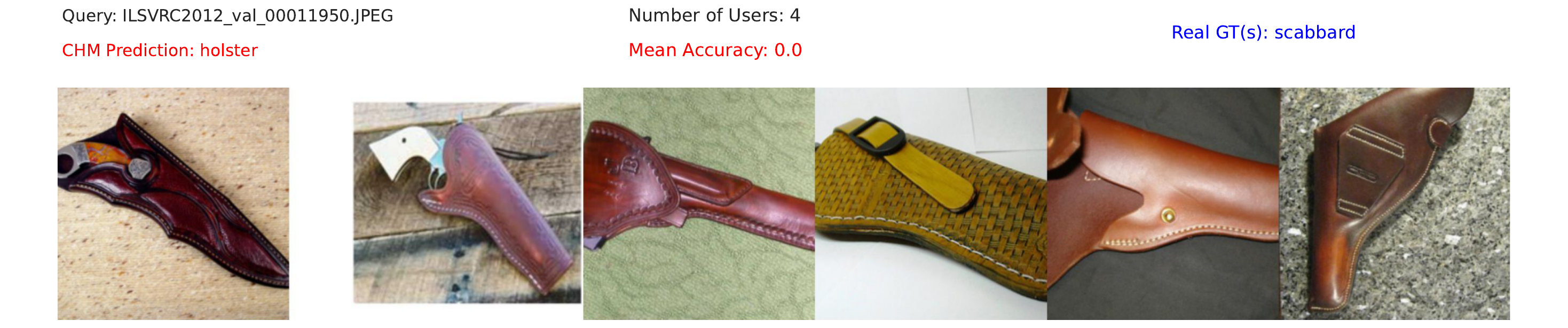}
\end{subfigure}\par


\caption{Accepting wrong CHM-NN prediction due to confusing explanations}
\label{fig:accepting_wrong_ai_chm_nn}
\end{figure}

\begin{figure}[!htbp]

\begin{subfigure}[b]{1\textwidth}
\includegraphics[width=\linewidth, trim=1cm .0cm .0cm .0cm,clip]{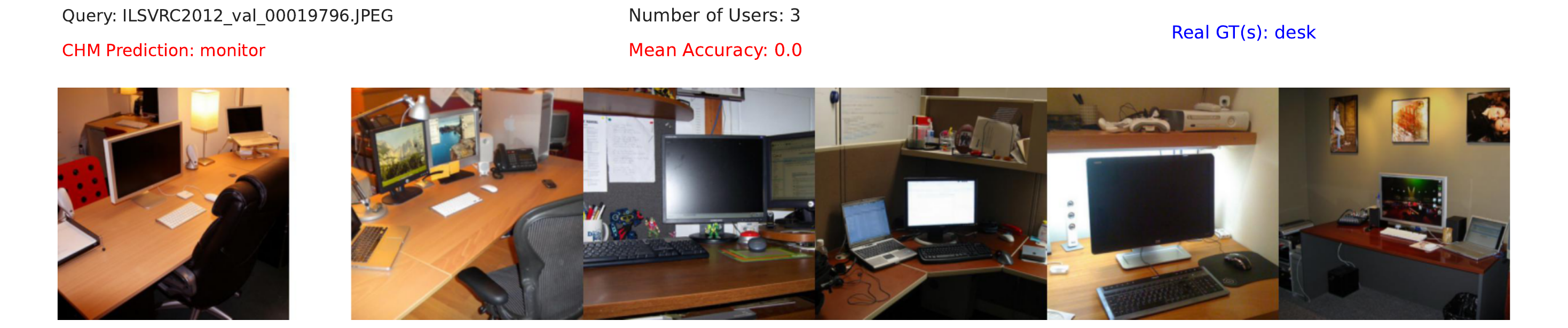}
\end{subfigure}\par

\begin{subfigure}[b]{1\textwidth}
\includegraphics[width=\linewidth, trim=1cm .0cm .0cm .0cm,clip]{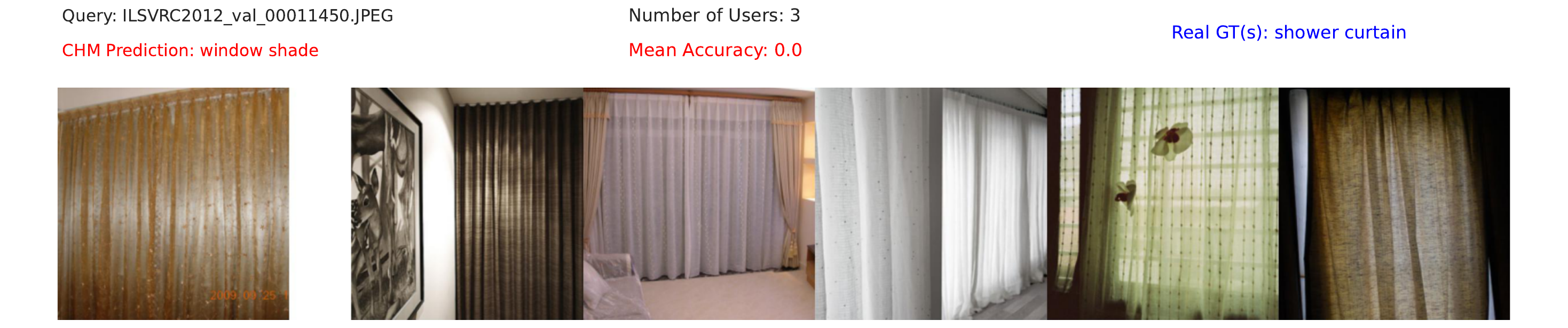}
\end{subfigure}\par


\begin{subfigure}[b]{1\textwidth}
\includegraphics[width=\linewidth, trim=1cm .0cm .0cm .0cm,clip]{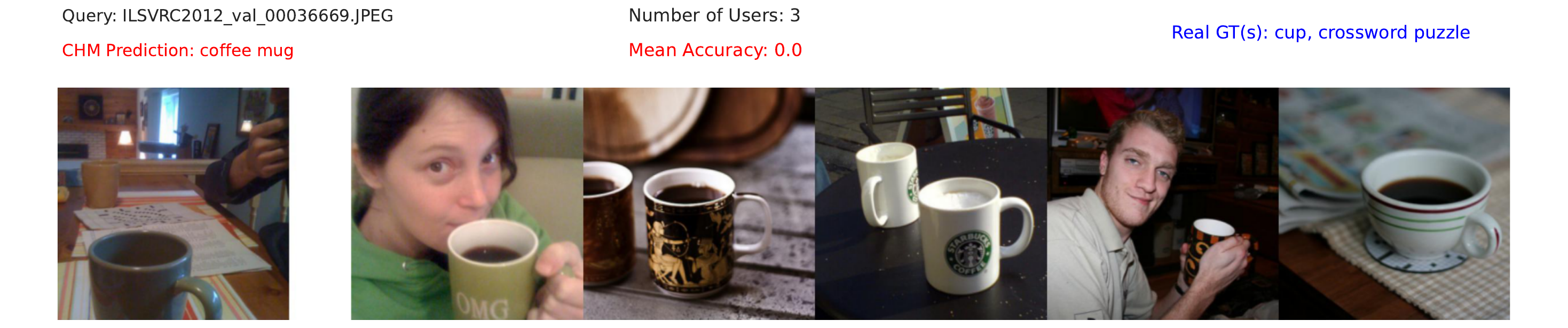}
\end{subfigure}\par


\caption{Accepting wrong CHM-NN prediction due to poor ImageNet-ReaL labeling}
\label{fig:accepting_wrong_ai_chm_nn_badlabels}
\end{figure}

\clearpage
\subsubsection{Accepting the wrong CHM-Corr Classifier's prediction}

\begin{figure}[!htbp]

    \begin{subfigure}[b]{1\textwidth}
        \includegraphics[width=\linewidth, trim=0cm .0cm .0cm .0cm,clip]{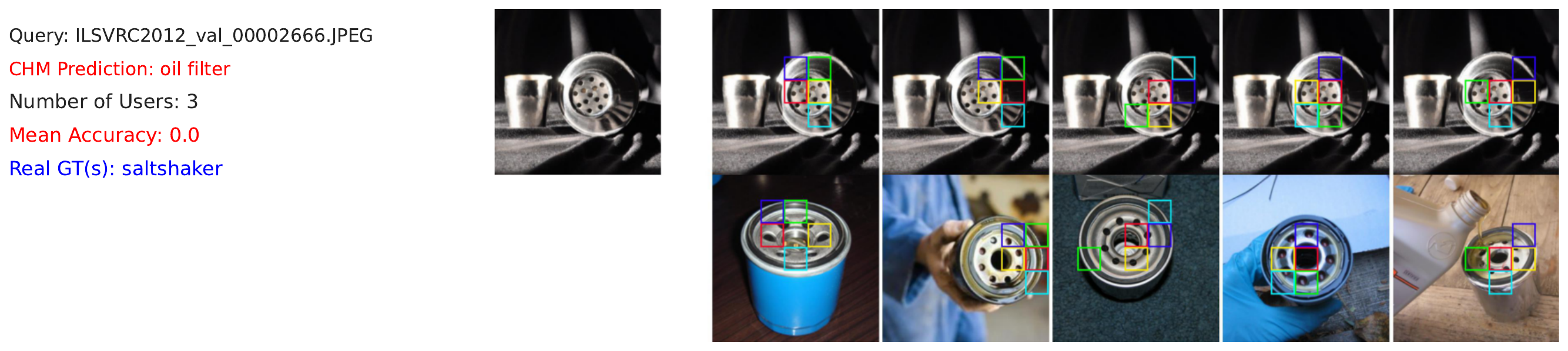}
    \end{subfigure}\par
    \begin{subfigure}[b]{1\textwidth}
        \includegraphics[width=\linewidth, trim=0cm .0cm .0cm .0cm,clip]{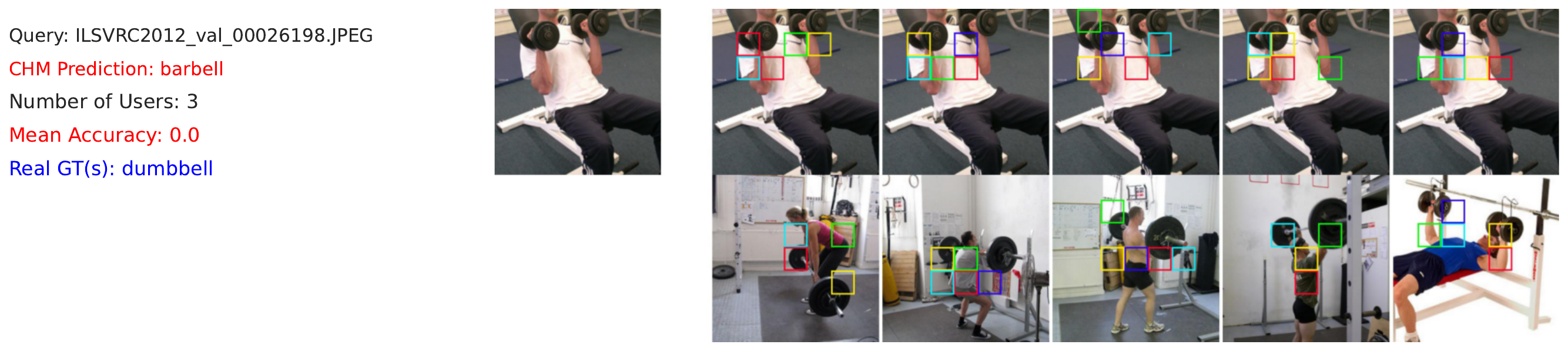}
    \end{subfigure}\par
    \begin{subfigure}[b]{1\textwidth}
        \includegraphics[width=\linewidth, trim=0cm .0cm .0cm .0cm,clip]{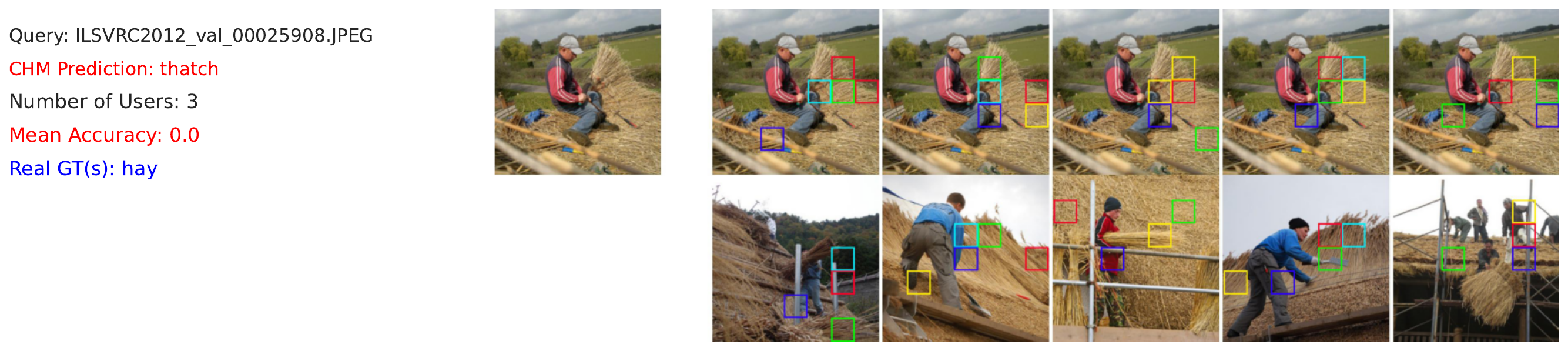}
    \end{subfigure}\par
        
        \caption{Accepting wrong CHM-Corr prediction due to confusing explanations}
        \label{fig:accepting_wrong_ai_chm_corr}
\end{figure}

\begin{figure}[!htbp]
        \begin{subfigure}[b]{1\textwidth}
                \includegraphics[width=\linewidth, trim=0cm .0cm .0cm .0cm,clip]{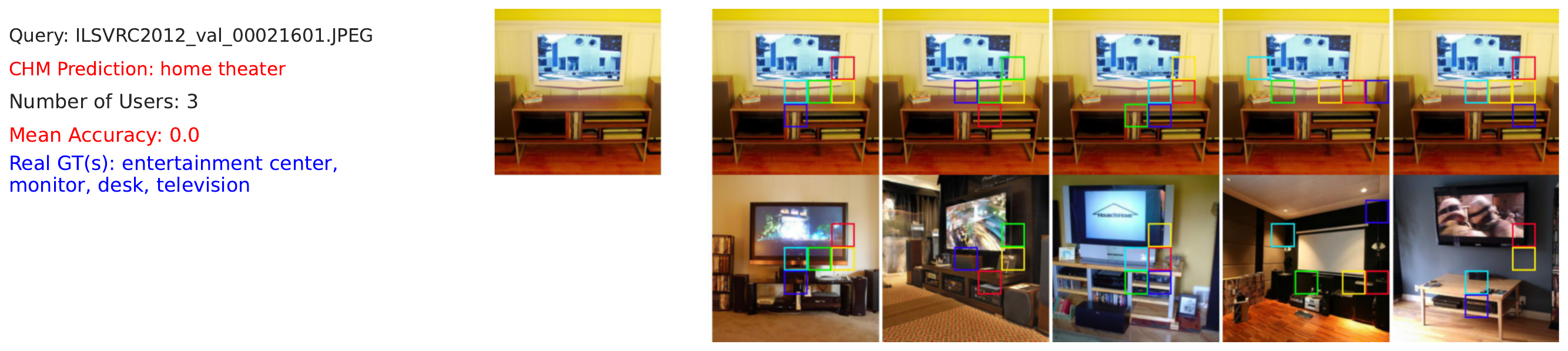}
        \end{subfigure}\par
         \begin{subfigure}[b]{1\textwidth}
                \includegraphics[width=\linewidth, trim=0cm .0cm .0cm .0cm,clip]{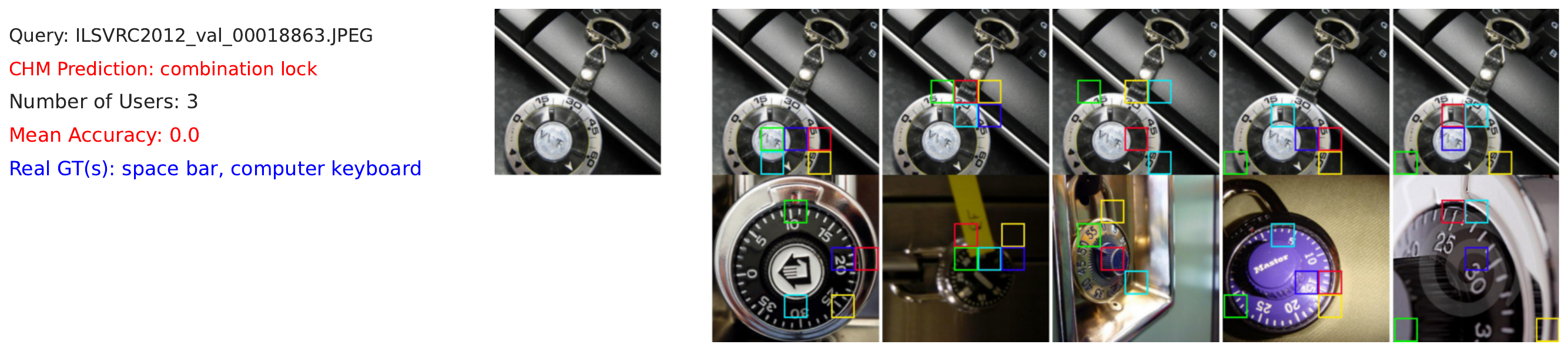}
        \end{subfigure}\par
        \begin{subfigure}[b]{1\textwidth}
        \includegraphics[width=\linewidth, trim=0cm .0cm .0cm .0cm,clip]{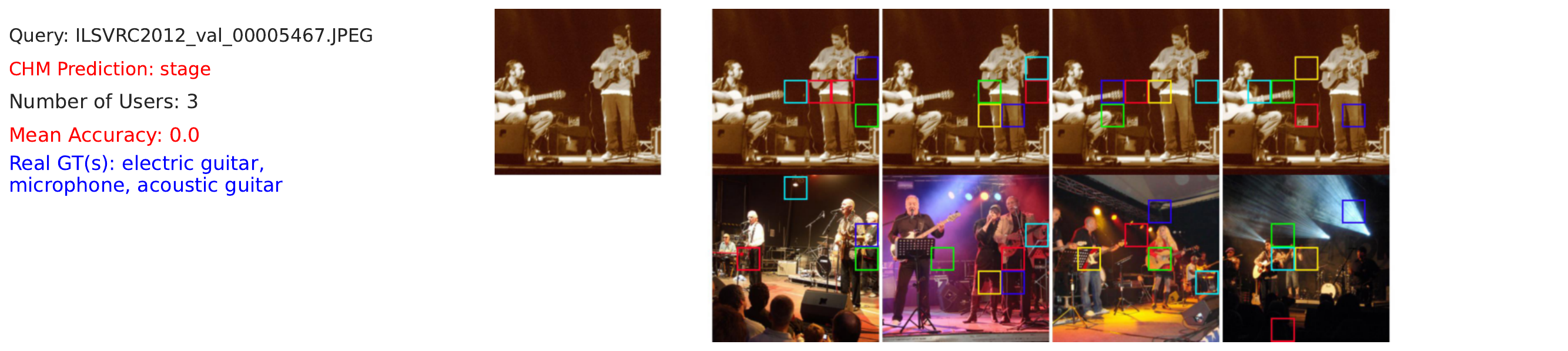}
        \end{subfigure}\par

        \caption{Accepting wrong CHM-Corr prediction due to poor ImageNet-ReaL labeling}
        \label{fig:accepting_wrong_ai_chm_corr_Badlabels}
\end{figure}

\clearpage
\subsection{When explanations fool users}
\label{supp:fooling_users}



This section provides clear evidence that explanations have the potential to fool human users. Both ResNet-50 and EMD-Corr misclassified an image of \class{tow truck} into \class{cab}. When asking a user to accept or reject this particular misclassification, they act differently based on the provided explanation. A total of 6 users who saw the query without any visual explanation were able to correctly reject AI's decision, while 3 out of 6 (50\%) users who received either an EMD-NN or EMD-Corr explanation incorrectly accepted the decision.


\begin{figure}[!htbp]
        \begin{subfigure}[b]{1\textwidth}
                \includegraphics[width=\linewidth, trim=3cm .0cm .0cm .0cm,clip]{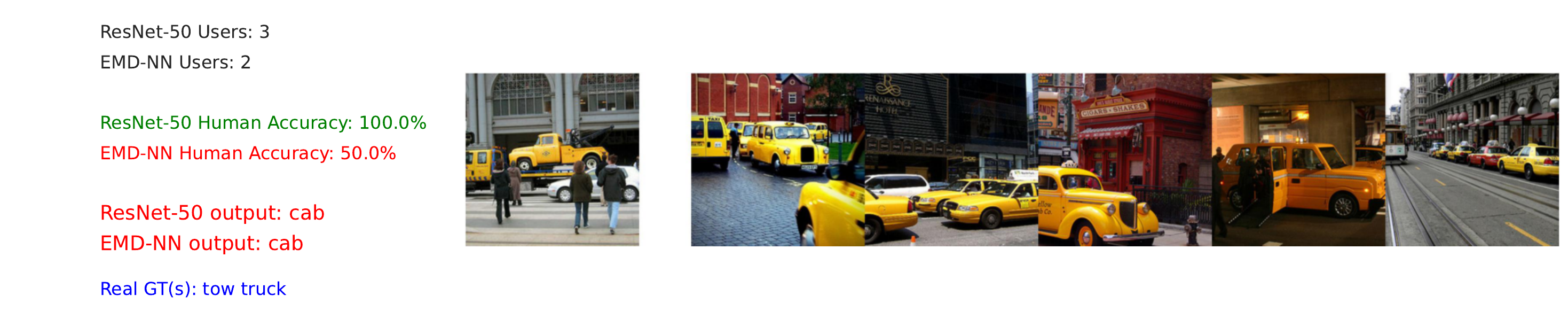}
        \end{subfigure}\par
        \begin{subfigure}[b]{1\textwidth}
                \includegraphics[width=\linewidth, trim=3cm .0cm .0cm .0cm,clip]{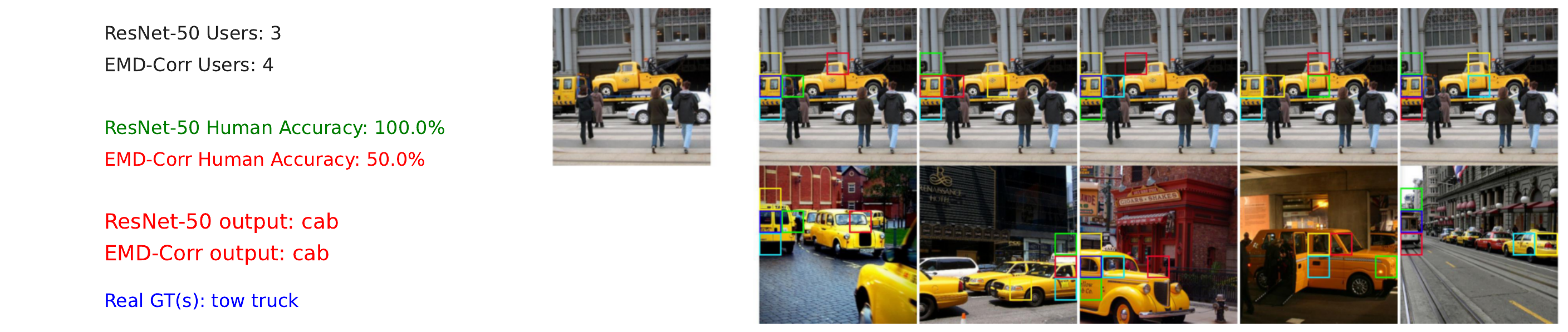}
        \end{subfigure}\par
        \caption{Samples for human users failing to reject wrong AI decisions---a tow truck misclassified as a \class{cab} by EMD-Corr classifier. }
        \label{fig:fooling_users_sample_1}
\end{figure}



\clearpage
\section{Classification accuracy of Human-AI teams}
\label{supp:complementary_perf}

In this section, we provide a detailed breakdown of human-AI team accuracy at different confidence thresholds.

We divide the set of images into two groups for each confidence threshold $T$: (1) images in which the AI's confidence equals or exceeds $T$, and (2) images in which the AI's confidence is less than $ T$.
In the first group, we only consider AI's decision, while for the second group, we ask a human user to judge AI's predicted label, i.e., whether users accept or reject AI's classification.
The aggregate accuracy of the human-AI team is the weighted average of the accuracy obtained from both groups.
To determine the best threshold, we first determine the value of $T$ that results in the best AI-alone accuracy on a small subset of the ImageNet-ReaL (2K images) and CUB (1K images) datasets, and then we evaluate the AI-alone accuracy on the held-out set for each dataset (42K images on ImageNet-ReaL and 4K on CUB).

\begin{table}[!htbp]
\caption{ResNet-50 - Aggregating Human and AI (\%) -- Bold numbers represent human-AI team performance at the optimal threshold}
\setlength\tabcolsep{2pt}
\label{tab:table-c-resnet-t}
\resizebox{\textwidth}{!}{%
\begin{tabular}{@{}rrrrrrrrr@{}}
\toprule
\multicolumn{1}{c}{} & \multicolumn{4}{c}{ImageNet}                          & \multicolumn{4}{c}{CUB}         \\ \midrule
\multicolumn{1}{c}{$T$} &
  \multicolumn{1}{c}{\begin{tabular}[c]{@{}c@{}}$\%$ of \\ images\\ handled \\ by AI\end{tabular}} &
  \multicolumn{1}{c}{\begin{tabular}[c]{@{}c@{}}AI-alone\\ accuracy\\ \small{(confidence >= $T$)}\end{tabular}} &
  \multicolumn{1}{c}{\begin{tabular}[c]{@{}c@{}}human\\ accuracy\\ \small{(confidence < $T$)}\end{tabular}} &
  \multicolumn{1}{c|}{\begin{tabular}[c]{@{}c@{}}Aggregated\\ human-AI\\ accuracy\end{tabular}} &
  \multicolumn{1}{c}{\begin{tabular}[c]{@{}c@{}}$\%$ of \\ images\\ handled \\ by AI\end{tabular}} &
  \multicolumn{1}{c}{\begin{tabular}[c]{@{}c@{}}AI-alone\\ accuracy\\ \small{(confidence >= $T$)}\end{tabular}} &
  \multicolumn{1}{c}{\begin{tabular}[c]{@{}c@{}}human\\ accuracy\\ \small{(confidence < $T$)}\end{tabular}} &
  \multicolumn{1}{c}{\begin{tabular}[c]{@{}c@{}}Aggregated\\ human-AI\\ accuracy\end{tabular}} \\ \midrule
0.00                 & 100.00 & 83.14  & n/a    & \multicolumn{1}{r|}{n/a}   & 100.00 & 85.83 & n/a    & n/a   \\
0.05                 & 99.98  & 83.16  & 100.00 & \multicolumn{1}{r|}{83.16} & 100.00 & 85.83 & n/a    & n/a   \\
0.10                 & 99.71  & 83.34  & 100.00 & \multicolumn{1}{r|}{83.39} & 100.00 & 85.83 & n/a    & n/a   \\
0.15                 & 98.74  & 83.96  & 89.09  & \multicolumn{1}{r|}{84.03} & 99.91  & 85.87 & 100.00 & 85.88 \\
0.20                 & 97.86  & 84.48  & 85.98  & \multicolumn{1}{r|}{84.51} & 99.71  & 86.01 & 76.47  & 85.99 \\
0.25                 & 96.43  & 85.29  & 89.82  & \multicolumn{1}{r|}{85.45} & 99.40  & 86.18 & 79.49  & 86.14 \\
0.30                 & 94.39  & 86.35  & 92.41  & \multicolumn{1}{r|}{86.69} & 98.88  & 86.47 & 83.93  & 86.44 \\
0.35                 & 92.47  & 87.32  & 89.14  & \multicolumn{1}{r|}{87.46} & 98.19  & 86.89 & 76.40  & 86.70 \\
0.40                 & 90.84  & 88.13  & 86.73  & \multicolumn{1}{r|}{88.00} & 97.45  & 87.32 & 72.17  & 86.93 \\
0.45                 & 88.50  & 89.15  & 84.62  & \multicolumn{1}{r|}{\cellcolor[HTML]{\thresholdcolor}\textbf{88.63}} & 96.01  & 87.99 & 69.36  & 87.25 \\
0.50                 & 85.88  & 90.20  & 83.79  & \multicolumn{1}{r|}{89.29} & 94.32  & 88.82 & 65.38  & 87.49 \\
0.55                 & 82.65  & 91.35  & 81.52  & \multicolumn{1}{r|}{89.64} & 92.16  & 89.85 & 59.27  & \cellcolor[HTML]{\thresholdcolor}\textbf{87.45} \\
0.60                 & 78.96  & 92.59  & 80.80  & \multicolumn{1}{r|}{90.11} & 89.47  & 90.91 & 60.78  & 87.74 \\
0.65                 & 76.57  & 93.36  & 80.50  & \multicolumn{1}{r|}{90.35} & 87.68  & 91.81 & 57.23  & 87.55 \\
0.70                 & 72.85  & 94.50  & 77.83  & \multicolumn{1}{r|}{89.98} & 84.69  & 92.81 & 54.56  & 86.95 \\
0.75                 & 70.17  & 95.24  & 76.06  & \multicolumn{1}{r|}{89.52} & 82.52  & 93.45 & 54.60  & 86.66 \\
0.80                 & 66.77  & 96.04  & 76.10  & \multicolumn{1}{r|}{89.41} & 79.41  & 94.48 & 52.55  & 85.85 \\
0.85                 & 61.89  & 96.99  & 75.65  & \multicolumn{1}{r|}{88.86} & 75.51  & 95.52 & 51.72  & 84.79 \\
0.90                 & 57.63  & 97.65  & 75.63  & \multicolumn{1}{r|}{88.32} & 71.88  & 96.37 & 51.91  & 83.87 \\
0.95                 & 47.42  & 98.67  & 76.08  & \multicolumn{1}{r|}{86.79} & 61.08  & 97.68 & 54.55  & 80.89 \\
1.00                 & 0.47   & 100.00 & 81.52  & 81.61                      & 0.00   & n/a   & 65.50  & n/a   \\ \bottomrule
\end{tabular}%
}
\end{table}

\begin{table}[!htbp]
\caption{kNN - Aggregating Human and AI (\%) -- Bold numbers represent human-AI team performance at the optimal threshold}
\label{tab:table-c-knn}
\setlength\tabcolsep{2pt}
\resizebox{\textwidth}{!}{%
\begin{tabular}{@{}rrrrrrrrr@{}}
\toprule
\multicolumn{1}{c}{} & \multicolumn{4}{c}{ImageNet}                         & \multicolumn{4}{c}{CUB}         \\ \midrule
\multicolumn{1}{c}{$T$} &
  \multicolumn{1}{c}{\begin{tabular}[c]{@{}c@{}}$\%$ of \\ images\\ handled \\ by AI\end{tabular}} &
  \multicolumn{1}{c}{\begin{tabular}[c]{@{}c@{}}AI-alone\\ accuracy\\ \small{(confidence >= $T$)}\end{tabular}} &
  \multicolumn{1}{c}{\begin{tabular}[c]{@{}c@{}}human\\ accuracy\\ \small{(confidence < $T$)}\end{tabular}} &
  \multicolumn{1}{c|}{\begin{tabular}[c]{@{}c@{}}Aggregated\\ human-AI\\ accuracy\end{tabular}} &
  \multicolumn{1}{c}{\begin{tabular}[c]{@{}c@{}}$\%$ of \\ images\\ handled \\ by AI\end{tabular}} &
  \multicolumn{1}{c}{\begin{tabular}[c]{@{}c@{}}AI-alone\\ accuracy\\ \small{(confidence >= $T$)}\end{tabular}} &
  \multicolumn{1}{c}{\begin{tabular}[c]{@{}c@{}}human\\ accuracy\\ \small{(confidence < $T$)}\end{tabular}} &
  \multicolumn{1}{c}{\begin{tabular}[c]{@{}c@{}}Aggregated\\ human-AI\\ accuracy\end{tabular}} \\ \midrule
0.00                 & 100.00 & 82.14 & n/a    & \multicolumn{1}{r|}{n/a}   & 100.00 & 85.47 & n/a    & n/a   \\
0.05                 & 100.00 & 82.14 & n/a    & \multicolumn{1}{r|}{n/a}   & 100.00 & 85.47 & n/a    & n/a   \\
0.10                 & 99.99  & 82.16 & 100.00 & \multicolumn{1}{r|}{82.16} & 100.00 & 85.47 & n/a    & n/a   \\
0.15                 & 98.26  & 83.34 & 97.14  & \multicolumn{1}{r|}{83.58} & 99.86  & 85.59 & 100.00 & 85.61 \\
0.20                 & 98.26  & 83.34 & 97.14  & \multicolumn{1}{r|}{83.58} & 99.86  & 85.59 & 100.00 & 85.61 \\
0.25                 & 96.52  & 84.36 & 90.23  & \multicolumn{1}{r|}{84.57} & 99.62  & 85.76 & 68.18  & 85.69 \\
0.30                 & 91.89  & 86.85 & 80.06  & \multicolumn{1}{r|}{86.30} & 97.20  & 87.18 & 50.70  & 86.16 \\
0.35                 & 89.25  & 88.10 & 77.86  & \multicolumn{1}{r|}{87.00} & 94.55  & 88.68 & 47.83  & 86.45 \\
0.40                 & 89.25  & 88.10 & 77.86  & \multicolumn{1}{r|}{87.00} & 94.55  & 88.68 & 47.83  & 86.45 \\
0.45                 & 86.34  & 89.44 & 73.40  & \multicolumn{1}{r|}{\cellcolor[HTML]{\thresholdcolor}\textbf{87.24}} & 91.15  & 90.13 & 50.85  & \cellcolor[HTML]{\thresholdcolor}\textbf{86.66} \\
0.50                 & 83.25  & 90.67 & 70.78  & \multicolumn{1}{r|}{87.34} & 86.73  & 92.02 & 50.70  & 86.54 \\
0.55                 & 79.74  & 91.91 & 67.99  & \multicolumn{1}{r|}{87.06} & 81.01  & 93.97 & 47.31  & 85.11 \\
0.60                 & 72.51  & 94.24 & 67.87  & \multicolumn{1}{r|}{86.99} & 71.52  & 96.67 & 47.39  & 82.63 \\
0.65                 & 72.51  & 94.24 & 67.87  & \multicolumn{1}{r|}{86.99} & 71.52  & 96.67 & 47.39  & 82.63 \\
0.70                 & 65.44  & 96.15 & 67.43  & \multicolumn{1}{r|}{86.23} & 62.15  & 97.81 & 49.68  & 79.59 \\
0.75                 & 65.44  & 96.15 & 67.43  & \multicolumn{1}{r|}{86.23} & 62.15  & 97.81 & 49.68  & 79.59 \\
0.80                 & 61.82  & 96.91 & 66.50  & \multicolumn{1}{r|}{85.30} & 56.87  & 98.12 & 51.14  & 77.86 \\
0.85                 & 53.34  & 98.10 & 66.93  & \multicolumn{1}{r|}{83.55} & 45.50  & 99.01 & 52.95  & 73.91 \\
0.90                 & 53.34  & 98.10 & 66.93  & \multicolumn{1}{r|}{83.55} & 45.50  & 99.01 & 52.95  & 73.91 \\
0.95                 & 36.77  & 99.19 & 70.42  & \multicolumn{1}{r|}{81.00} & 28.58  & 99.28 & 58.60  & 70.23 \\
1.00                 & 36.77  & 99.19 & 70.42  & 81.00                      & 28.58  & 99.28 & 58.60  & 70.23 \\ \bottomrule
\end{tabular}%
}
\end{table}

\begin{table}[!htbp]
\caption{EMD-NN Aggregating Human and AI (\%) -- Bold numbers represent human-AI team performance at the optimal threshold}
\label{tab:table-c-emd-nn}
\setlength\tabcolsep{2pt}
\resizebox{\textwidth}{!}{%
\begin{tabular}{@{}rrrrrrrrr@{}}
\toprule
\multicolumn{1}{c}{} & \multicolumn{4}{c}{ImageNet}                         & \multicolumn{4}{c}{CUB}        \\ \midrule
\multicolumn{1}{c}{$T$} &
  \multicolumn{1}{c}{\begin{tabular}[c]{@{}c@{}}$\%$ of \\ images\\ handled \\ by AI\end{tabular}} &
  \multicolumn{1}{c}{\begin{tabular}[c]{@{}c@{}}AI-alone\\ accuracy\\ \small{(confidence >= $T$)}\end{tabular}} &
  \multicolumn{1}{c}{\begin{tabular}[c]{@{}c@{}}human\\ accuracy\\ \small{(confidence < $T$)}\end{tabular}} &
  \multicolumn{1}{c|}{\begin{tabular}[c]{@{}c@{}}Aggregated\\ human-AI\\ accuracy\end{tabular}} &
  \multicolumn{1}{c}{\begin{tabular}[c]{@{}c@{}}$\%$ of \\ images\\ handled \\ by AI\end{tabular}} &
  \multicolumn{1}{c}{\begin{tabular}[c]{@{}c@{}}AI-alone\\ accuracy\\ \small{(confidence >= $T$)}\end{tabular}} &
  \multicolumn{1}{c}{\begin{tabular}[c]{@{}c@{}}human\\ accuracy\\ \small{(confidence < $T$)}\end{tabular}} &
  \multicolumn{1}{c}{\begin{tabular}[c]{@{}c@{}}Aggregated\\ human-AI\\ accuracy\end{tabular}} \\ \midrule
0.00                 & 100.00 & 82.39 & n/a    & \multicolumn{1}{r|}{n/a}   & 100.00 & 84.98 & n/a   & NaN   \\
0.05                 & 100.00 & 82.39 & n/a    & \multicolumn{1}{r|}{n/a}   & 100.00 & 84.98 & NaN   & NaN   \\
0.10                 & 99.99  & 82.40 & 100.00 & \multicolumn{1}{r|}{82.40} & 100.00 & 84.98 & NaN   & NaN   \\
0.15                 & 98.19  & 83.63 & 96.10  & \multicolumn{1}{r|}{83.86} & 99.81  & 85.15 & 60.00 & 85.10 \\
0.20                 & 98.19  & 83.63 & 96.10  & \multicolumn{1}{r|}{83.86} & 99.81  & 85.15 & 60.00 & 85.10 \\
0.25                 & 96.36  & 84.72 & 95.24  & \multicolumn{1}{r|}{85.10} & 99.50  & 85.34 & 68.75 & 85.26 \\
0.30                 & 91.86  & 87.12 & 88.36  & \multicolumn{1}{r|}{87.22} & 96.50  & 87.03 & 55.70 & 85.94 \\
0.35                 & 89.14  & 88.37 & 83.11  & \multicolumn{1}{r|}{87.80} & 93.68  & 88.39 & 49.21 & 85.92 \\
0.40                 & 89.14  & 88.37 & 83.11  & \multicolumn{1}{r|}{\cellcolor[HTML]{\thresholdcolor}\textbf{87.80}} & 93.68  & 88.39 & 49.21 & 85.92 \\
0.45                 & 86.25  & 89.59 & 81.49  & \multicolumn{1}{r|}{88.47} & 89.40  & 90.19 & 47.67 & \cellcolor[HTML]{\thresholdcolor}\textbf{85.69} \\
0.50                 & 83.19  & 90.80 & 77.45  & \multicolumn{1}{r|}{88.56} & 83.98  & 92.40 & 48.92 & 85.43 \\
0.55                 & 79.56  & 92.13 & 73.59  & \multicolumn{1}{r|}{88.34} & 78.29  & 94.47 & 48.74 & 84.54 \\
0.60                 & 72.11  & 94.53 & 70.41  & \multicolumn{1}{r|}{87.80} & 68.47  & 96.45 & 49.35 & 81.60 \\
0.65                 & 72.11  & 94.53 & 70.41  & \multicolumn{1}{r|}{87.80} & 68.47  & 96.45 & 49.35 & 81.60 \\
0.70                 & 65.02  & 96.16 & 68.68  & \multicolumn{1}{r|}{86.55} & 57.96  & 97.89 & 50.90 & 78.13 \\
0.75                 & 65.02  & 96.16 & 68.68  & \multicolumn{1}{r|}{86.55} & 57.96  & 97.89 & 50.90 & 78.13 \\
0.80                 & 61.30  & 96.94 & 68.46  & \multicolumn{1}{r|}{85.92} & 51.90  & 98.17 & 52.96 & 76.42 \\
0.85                 & 52.32  & 98.18 & 69.70  & \multicolumn{1}{r|}{84.60} & 39.68  & 99.30 & 55.29 & 72.76 \\
0.90                 & 52.32  & 98.18 & 69.70  & \multicolumn{1}{r|}{84.60} & 39.68  & 99.30 & 55.29 & 72.76 \\
0.95                 & 35.24  & 99.18 & 73.15  & \multicolumn{1}{r|}{82.33} & 19.57  & 99.38 & 60.80 & 68.35 \\
1.00                 & 35.24  & 99.18 & 73.15  & 82.33                      & 19.57  & 99.38 & 60.80 & 68.35 \\ \bottomrule
\end{tabular}%
}
\end{table}

\begin{table}[!htbp]
\caption{EMD-Corr Aggregating Human and AI (\%) -- Bold numbers represent human-AI team performance at the optimal threshold}
\label{tab:table-c-emd-corr}
\setlength\tabcolsep{2pt}
\resizebox{\textwidth}{!}{%
\begin{tabular}{@{}rrrrrrrrr@{}}
\toprule
\multicolumn{1}{c}{} & \multicolumn{4}{c}{ImageNet}                         & \multicolumn{4}{c}{CUB}         \\ \midrule
\multicolumn{1}{c}{$T$} &
  \multicolumn{1}{c}{\begin{tabular}[c]{@{}c@{}}$\%$ of \\ images\\ handled \\ by AI\end{tabular}} &
  \multicolumn{1}{c}{\begin{tabular}[c]{@{}c@{}}AI-alone\\ accuracy\\ \small{(confidence >= $T$)}\end{tabular}} &
  \multicolumn{1}{c}{\begin{tabular}[c]{@{}c@{}}human\\ accuracy\\ \small{(confidence < $T$)}\end{tabular}} &
  \multicolumn{1}{c|}{\begin{tabular}[c]{@{}c@{}}Aggregated\\ human-AI\\ accuracy\end{tabular}} &
  \multicolumn{1}{c}{\begin{tabular}[c]{@{}c@{}}$\%$ of \\ images\\ handled \\ by AI\end{tabular}} &
  \multicolumn{1}{c}{\begin{tabular}[c]{@{}c@{}}AI-alone\\ accuracy\\ \small{(confidence >= $T$)}\end{tabular}} &
  \multicolumn{1}{c}{\begin{tabular}[c]{@{}c@{}}human\\ accuracy\\ \small{(confidence < $T$)}\end{tabular}} &
  \multicolumn{1}{c}{\begin{tabular}[c]{@{}c@{}}Aggregated\\ human-AI\\ accuracy\end{tabular}} \\ \midrule
0.00                 & 100.00 & 82.39 & n/a    & \multicolumn{1}{r|}{n/a}   & 100.00 & 84.98 & n/a    & n/a   \\
0.05                 & 100.00 & 82.39 & n/a    & \multicolumn{1}{r|}{n/a}   & 100.00 & 84.98 & n/a    & n/a   \\
0.10                 & 99.99  & 82.40 & 100.00 & \multicolumn{1}{r|}{82.40} & 100.00 & 84.98 & n/a    & n/a   \\
0.15                 & 98.19  & 83.63 & 95.29  & \multicolumn{1}{r|}{83.84} & 99.81  & 85.15 & 100.00 & 85.17 \\
0.20                 & 98.19  & 83.63 & 95.29  & \multicolumn{1}{r|}{83.84} & 99.81  & 85.15 & 100.00 & 85.17 \\
0.25                 & 96.36  & 84.72 & 95.57  & \multicolumn{1}{r|}{85.11} & 99.50  & 85.34 & 86.67  & 85.35 \\
0.30                 & 91.86  & 87.12 & 89.27  & \multicolumn{1}{r|}{87.29} & 96.50  & 87.03 & 69.70  & 86.43 \\
0.35                 & 89.14  & 88.37 & 85.19  & \multicolumn{1}{r|}{88.02} & 93.68  & 88.39 & 60.36  & 86.62 \\
0.40                 & 89.14  & 88.37 & 85.19  & \multicolumn{1}{r|}{\cellcolor[HTML]{\thresholdcolor}\textbf{88.02}} & 93.68  & 88.39 & 60.36  & 86.62 \\
0.45                 & 86.25  & 89.59 & 82.59  & \multicolumn{1}{r|}{88.62} & 89.40  & 90.19 & 58.70  & \cellcolor[HTML]{\thresholdcolor}\textbf{86.86} \\
0.50                 & 83.19  & 90.80 & 79.17  & \multicolumn{1}{r|}{88.85} & 83.98  & 92.40 & 57.24  & 86.77 \\
0.55                 & 79.56  & 92.13 & 74.67  & \multicolumn{1}{r|}{88.56} & 78.29  & 94.47 & 57.26  & 86.39 \\
0.60                 & 72.11  & 94.53 & 72.40  & \multicolumn{1}{r|}{88.36} & 68.47  & 96.45 & 58.34  & 84.43 \\
0.65                 & 72.11  & 94.53 & 72.40  & \multicolumn{1}{r|}{88.36} & 68.47  & 96.45 & 58.34  & 84.43 \\
0.70                 & 65.02  & 96.16 & 70.44  & \multicolumn{1}{r|}{87.16} & 57.96  & 97.89 & 58.20  & 81.20 \\
0.75                 & 65.02  & 96.16 & 70.44  & \multicolumn{1}{r|}{87.16} & 57.96  & 97.89 & 58.20  & 81.20 \\
0.80                 & 61.30  & 96.94 & 70.71  & \multicolumn{1}{r|}{86.79} & 51.90  & 98.17 & 59.04  & 79.35 \\
0.85                 & 52.32  & 98.18 & 71.74  & \multicolumn{1}{r|}{85.57} & 39.68  & 99.30 & 60.70  & 76.01 \\
0.90                 & 52.32  & 98.18 & 71.74  & \multicolumn{1}{r|}{85.57} & 39.68  & 99.30 & 60.70  & 76.01 \\
0.95                 & 35.24  & 99.18 & 74.63  & \multicolumn{1}{r|}{83.28} & 19.57  & 99.38 & 64.53  & 71.35 \\
1.00                 & 35.24  & 99.18 & 74.63  & 83.28                      & 19.57  & 99.38 & 64.53  & 71.35 \\ \bottomrule
\end{tabular}%
}
\end{table}

\begin{table}[!htbp]
\caption{CHM-NN Aggregating Human and AI (\%)}
\label{tab:table-c-chm-nn}
\setlength\tabcolsep{2pt}
\resizebox{\textwidth}{!}{%
\begin{tabular}{@{}rrrrrrrrr@{}}
\toprule
\multicolumn{1}{c}{} & \multicolumn{4}{c}{ImageNet}                        & \multicolumn{4}{c}{CUB}        \\ \midrule
\multicolumn{1}{c}{$T$} &
  \multicolumn{1}{c}{\begin{tabular}[c]{@{}c@{}}$\%$ of \\ images\\ handled \\ by AI\end{tabular}} &
  \multicolumn{1}{c}{\begin{tabular}[c]{@{}c@{}}AI-alone\\ accuracy\\ \small{(confidence >= $T$)}\end{tabular}} &
  \multicolumn{1}{c}{\begin{tabular}[c]{@{}c@{}}human\\ accuracy\\ \small{(confidence < $T$)}\end{tabular}} &
  \multicolumn{1}{c|}{\begin{tabular}[c]{@{}c@{}}Aggregated\\ human-AI\\ accuracy\end{tabular}} &
  \multicolumn{1}{c}{\begin{tabular}[c]{@{}c@{}}$\%$ of \\ images\\ handled \\ by AI\end{tabular}} &
  \multicolumn{1}{c}{\begin{tabular}[c]{@{}c@{}}AI-alone\\ accuracy\\ \small{(confidence >= $T$)}\end{tabular}} &
  \multicolumn{1}{c}{\begin{tabular}[c]{@{}c@{}}human\\ accuracy\\ \small{(confidence < $T$)}\end{tabular}} &
  \multicolumn{1}{c}{\begin{tabular}[c]{@{}c@{}}Aggregated\\ human-AI\\ accuracy\end{tabular}} \\ \midrule
0.00                 & 100.00 & 82.05 & n/a   & \multicolumn{1}{r|}{n/a}   & 100.00 & 83.28 & n/a   & n/a   \\
0.05                 & 100.00 & 82.05 & n/a   & \multicolumn{1}{r|}{n/a}   & 100.00 & 83.28 & n/a   & n/a   \\
0.10                 & 99.99  & 82.06 & n/a   & \multicolumn{1}{r|}{n/a}   & 100.00 & 83.28 & n/a   & n/a   \\
0.15                 & 98.53  & 83.03 & 94.74 & \multicolumn{1}{r|}{83.20} & 99.86  & 83.36 & 80.00 & 83.35 \\
0.20                 & 98.53  & 83.03 & 94.74 & \multicolumn{1}{r|}{83.20} & 99.86  & 83.36 & 80.00 & 83.35 \\
0.25                 & 96.95  & 83.96 & 88.97 & \multicolumn{1}{r|}{84.11} & 99.52  & 83.56 & 61.54 & 83.45 \\
0.30                 & 92.86  & 86.17 & 81.37 & \multicolumn{1}{r|}{85.82} & 95.79  & 85.64 & 55.14 & 84.36 \\
0.35                 & 90.35  & 87.40 & 80.58 & \multicolumn{1}{r|}{86.75} & 92.18  & 87.25 & 54.92 & 84.72 \\
0.40                 & 90.35  & 87.40 & 80.58 & \multicolumn{1}{r|}{86.75} & 92.18  & 87.25 & 54.92 & 84.72 \\
0.45                 & 87.60  & 88.65 & 79.22 & \multicolumn{1}{r|}{87.48} & 87.04  & 89.63 & 53.19 & \cellcolor[HTML]{\thresholdcolor}\textbf{84.91} \\
0.50                 & 84.55  & 89.90 & 77.88 & \multicolumn{1}{r|}{\cellcolor[HTML]{\thresholdcolor}\textbf{88.05}} & 81.12  & 91.81 & 51.94 & 84.28 \\
0.55                 & 80.85  & 91.27 & 73.72 & \multicolumn{1}{r|}{87.91} & 74.28  & 94.17 & 50.63 & 82.97 \\
0.60                 & 73.48  & 93.72 & 69.48 & \multicolumn{1}{r|}{87.29} & 60.87  & 97.22 & 52.05 & 79.55 \\
0.65                 & 73.48  & 93.72 & 69.48 & \multicolumn{1}{r|}{87.29} & 60.87  & 97.22 & 52.05 & 79.55 \\
0.70                 & 66.57  & 95.65 & 68.99 & \multicolumn{1}{r|}{86.74} & 48.43  & 98.43 & 54.79 & 75.92 \\
0.75                 & 66.57  & 95.65 & 68.99 & \multicolumn{1}{r|}{86.74} & 48.43  & 98.43 & 54.79 & 75.92 \\
0.80                 & 63.00  & 96.34 & 69.23 & \multicolumn{1}{r|}{86.31} & 41.37  & 98.71 & 57.00 & 74.25 \\
0.85                 & 54.32  & 97.77 & 68.79 & \multicolumn{1}{r|}{84.53} & 25.51  & 99.26 & 60.18 & 70.14 \\
0.90                 & 54.32  & 97.77 & 68.79 & \multicolumn{1}{r|}{84.53} & 25.51  & 99.26 & 60.18 & 70.14 \\
0.95                 & 37.96  & 98.97 & 71.49 & \multicolumn{1}{r|}{81.92} & 9.48   & 99.64 & 64.32 & 67.66 \\
1.00                 & 37.96  & 98.97 & 71.49 & 81.92                      & 9.48   & 99.64 & 64.32 & 67.66 \\ \bottomrule
\end{tabular}%
}
\end{table}

\begin{table}[!htbp]
\caption{CHM-Corr Aggregating Human and AI (\%) -- Bold numbers represent human-AI team performance at the optimal threshold}
\label{tab:table-c-chm-corr}
\setlength\tabcolsep{2pt}
\resizebox{\textwidth}{!}{%
\begin{tabular}{@{}rrrrrrrrr@{}}
\toprule
\multicolumn{1}{c}{} & \multicolumn{4}{c}{ImageNet}                        & \multicolumn{4}{c}{CUB}         \\ \midrule
\multicolumn{1}{c}{$T$} &
  \multicolumn{1}{c}{\begin{tabular}[c]{@{}c@{}}$\%$ of \\ images\\ handled \\ by AI\end{tabular}} &
  \multicolumn{1}{c}{\begin{tabular}[c]{@{}c@{}}AI-alone\\ accuracy\\ \small{(confidence >= $T$)}\end{tabular}} &
  \multicolumn{1}{c}{\begin{tabular}[c]{@{}c@{}}human\\ accuracy\\ \small{(confidence < $T$)}\end{tabular}} &
  \multicolumn{1}{c|}{\begin{tabular}[c]{@{}c@{}}Aggregated\\ human-AI\\ accuracy\end{tabular}} &
  \multicolumn{1}{c}{\begin{tabular}[c]{@{}c@{}}$\%$ of \\ images\\ handled \\ by AI\end{tabular}} &
  \multicolumn{1}{c}{\begin{tabular}[c]{@{}c@{}}AI-alone\\ accuracy\\ \small{(confidence >= $T$)}\end{tabular}} &
  \multicolumn{1}{c}{\begin{tabular}[c]{@{}c@{}}human\\ accuracy\\ \small{(confidence < $T$)}\end{tabular}} &
  \multicolumn{1}{c}{\begin{tabular}[c]{@{}c@{}}Aggregated\\ human-AI\\ accuracy\end{tabular}} \\ \midrule
0.00                 & 100.00 & 82.05 & n/a   & \multicolumn{1}{r|}{n/a}   & 100.00 & 83.28 & n/a    & n/a   \\
0.05                 & 100.00 & 82.05 & n/a   & \multicolumn{1}{r|}{n/a}   & 100.00 & 83.28 & n/a    & n/a   \\
0.10                 & 99.99  & 82.06 & n/a   & \multicolumn{1}{r|}{n/a}   & 100.00 & 83.28 & n/a    & n/a   \\
0.15                 & 98.53  & 83.03 & 91.89 & \multicolumn{1}{r|}{83.16} & 99.86  & 83.36 & 100.00 & 83.38 \\
0.20                 & 98.53  & 83.03 & 91.89 & \multicolumn{1}{r|}{83.16} & 99.86  & 83.36 & 100.00 & 83.38 \\
0.25                 & 96.95  & 83.96 & 86.99 & \multicolumn{1}{r|}{84.05} & 99.52  & 83.56 & 53.85  & 83.42 \\
0.30                 & 92.86  & 86.17 & 81.90 & \multicolumn{1}{r|}{85.86} & 95.79  & 85.64 & 72.22  & 85.07 \\
0.35                 & 90.35  & 87.40 & 78.35 & \multicolumn{1}{r|}{86.53} & 92.18  & 87.25 & 71.06  & 85.98 \\
0.40                 & 90.35  & 87.40 & 78.35 & \multicolumn{1}{r|}{86.53} & 92.18  & 87.25 & 71.06  & 85.98 \\
0.45                 & 87.60  & 88.65 & 77.39 & \multicolumn{1}{r|}{87.26} & 87.04  & 89.63 & 63.58  & \cellcolor[HTML]{\thresholdcolor}\textbf{86.25} \\
0.50                 & 84.55  & 89.90 & 76.85 & \multicolumn{1}{r|}{\cellcolor[HTML]{\thresholdcolor}\textbf{87.89}} & 81.12  & 91.81 & 62.92  & 86.35 \\
0.55                 & 80.85  & 91.27 & 73.35 & \multicolumn{1}{r|}{87.84} & 74.28  & 94.17 & 61.15  & 85.68 \\
0.60                 & 73.48  & 93.72 & 70.30 & \multicolumn{1}{r|}{87.51} & 60.87  & 97.22 & 60.52  & 82.86 \\
0.65                 & 73.48  & 93.72 & 70.30 & \multicolumn{1}{r|}{87.51} & 60.87  & 97.22 & 60.52  & 82.86 \\
0.70                 & 66.57  & 95.65 & 70.21 & \multicolumn{1}{r|}{87.15} & 48.43  & 98.43 & 62.36  & 79.83 \\
0.75                 & 66.57  & 95.65 & 70.21 & \multicolumn{1}{r|}{87.15} & 48.43  & 98.43 & 62.36  & 79.83 \\
0.80                 & 63.00  & 96.34 & 70.56 & \multicolumn{1}{r|}{86.80} & 41.37  & 98.71 & 63.37  & 77.99 \\
0.85                 & 54.32  & 97.77 & 69.32 & \multicolumn{1}{r|}{84.78} & 25.51  & 99.26 & 65.20  & 73.89 \\
0.90                 & 54.32  & 97.77 & 69.32 & \multicolumn{1}{r|}{84.78} & 25.51  & 99.26 & 65.20  & 73.89 \\
0.95                 & 37.96  & 98.97 & 71.30 & \multicolumn{1}{r|}{81.80} & 9.48   & 99.64 & 68.29  & 71.26 \\
1.00                 & 37.96  & 98.97 & 71.30 & 81.80                      & 9.48   & 99.64 & 68.29  & 71.26 \\ \bottomrule
\end{tabular}%
}
\end{table}

\clearpage
\subsection{Human-AI team is better than AI-only}
\label{supp:section_human_ai_better_ai_alone}

\begin{figure}[!htbp]
    \centering
    \includegraphics[width=\textwidth]{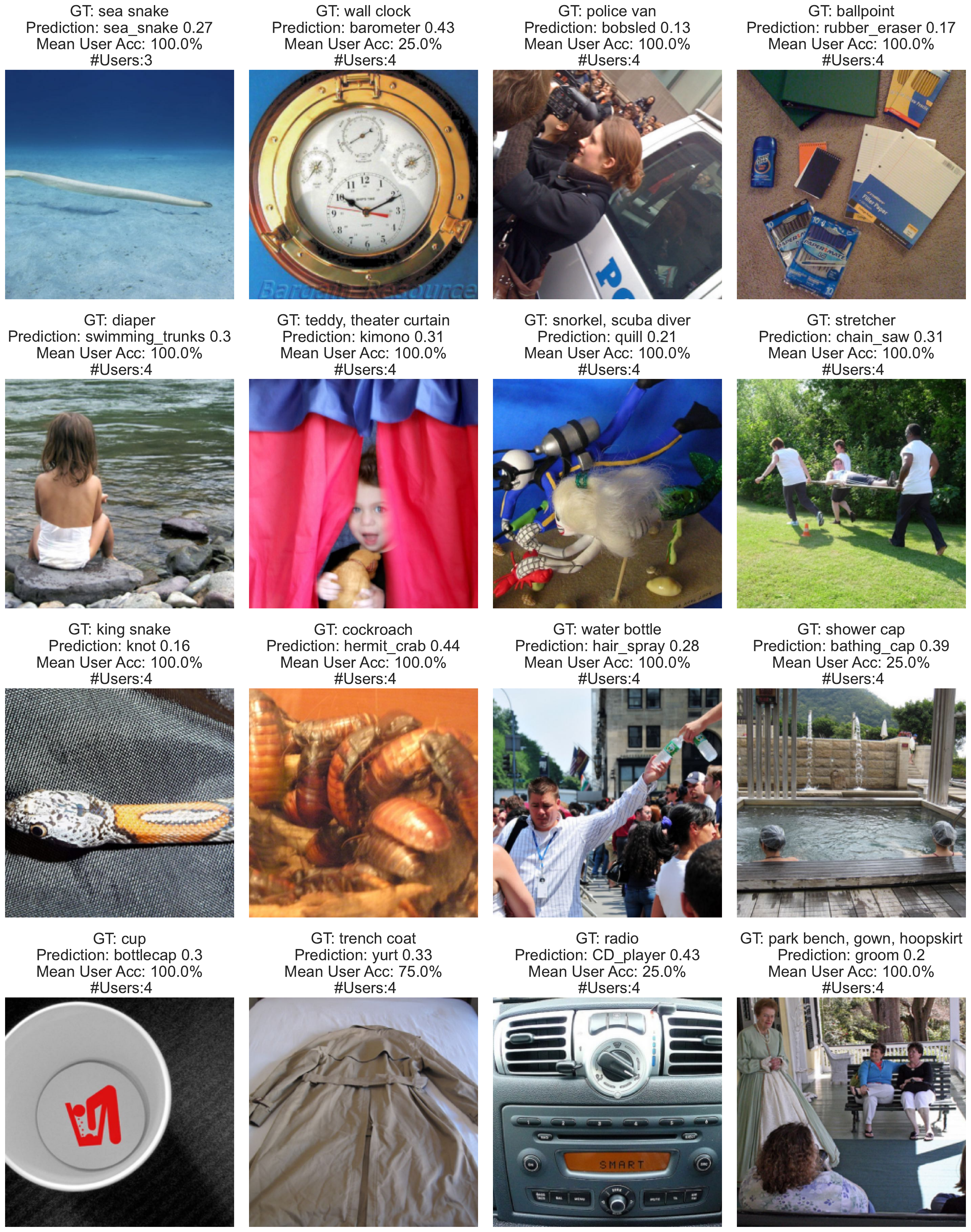}
    \caption{ImageNet samples at $T=0.45$ -- ResNet-50 classifier}
    \label{fig:supp_human_ait_team_imagenet_samples}
\end{figure}

Because there is a subset of images for which AIs are not confident, and have very low accuracy (accuracy = $47/408$ at $T=0.45$) (Table~\ref{fig:supp_human_ait_team_imagenet_samples}).
Therefore, humans helped increase the accuracy by looking at this subset and rejecting AI’s incorrect predictions. These images are easy for humans to reject (Figure \ref{fig:supp_human_ait_team_imagenet_samples}).

\begin{table}[!htbp]
\centering
\caption{Breakdown of the number of trials at different thresholds -- ResNet-50 -- ImageNet}
\label{tab:supp_breakdown_trials_imagenet}
\resizebox{\textwidth}{!}{%
\begin{tabular}{@{}rrrrr@{}}
\toprule
\multicolumn{1}{c}{\textbf{T}} & \multicolumn{1}{c}{\textbf{\begin{tabular}[c]{@{}c@{}}Human\\ Performance\end{tabular}}} & \multicolumn{1}{c}{\textbf{\# Trials}} & \multicolumn{1}{c}{\textbf{\begin{tabular}[c]{@{}c@{}}\# Trials\\ Correct AI Prediction\end{tabular}}} & \multicolumn{1}{c}{\textbf{\begin{tabular}[c]{@{}c@{}}\# Trials\\ Wrong AI Prediction\end{tabular}}} \\ \midrule
0.00 & n/a & n/a & 0 & 1 \\
0.05 & 100.00 & 3 & 0 & 3 \\
0.10 & 100.00 & 16 & 0 & 16 \\
0.15 & 89.09 & 55 & 3 & 52 \\
0.20 & 85.98 & 107 & 8 & 99 \\
0.25 & 89.82 & 167 & 17 & 150 \\
0.30 & 92.41 & 224 & 20 & 204 \\
0.35 & 89.14 & 313 & 22 & 291 \\
0.40 & 86.73 & 392 & 34 & 358 \\
\cellcolor[HTML]{\thresholdcolor}\textbf{0.45} & \cellcolor[HTML]{\thresholdcolor}84.62 & \cellcolor[HTML]{\thresholdcolor}455 & \cellcolor[HTML]{\thresholdcolor}47 & \cellcolor[HTML]{\thresholdcolor}408 \\
0.50 & 83.79 & 543 & 55 & 488 \\
0.55 & 81.52 & 633 & 85 & 548 \\
0.60 & 80.80 & 729 & 124 & 605 \\
0.65 & 80.50 & 800 & 151 & 649 \\
0.70 & 77.83 & 875 & 160 & 715 \\
0.75 & 76.06 & 944 & 188 & 756 \\
0.80 & 76.10 & 996 & 209 & 787 \\
0.85 & 75.65 & 1072 & 258 & 814 \\
0.90 & 75.63 & 1145 & 310 & 835 \\
0.95 & 76.08 & 1292 & 413 & 879 \\
1.00 & 81.52 & 1797 & 891 & 906 \\ \bottomrule
\end{tabular}%
}
\end{table}

\clearpage
\subsection{Human-AI team is better than human-only}
\label{supp:section_human_ai_better_human_alone}

Because humans are not trained explicitly to perform image classification on CUB and ImageNet-ReaL, the mean human-only accuracy is 65.50\% and 81.56\% respectively.
When teaming up with AI, human-AI teams perform slightly better on ImageNet (81.56\% vs. 86.80\%) but substantially better on CUB (65.50\% vs. 87.94\%).
See \cref{table:aialone_and_humanteam}.

\begin{table}[!htbp]
\centering
\caption{Breakdown of the number of trials at different thresholds  -- ResNet-50 -- CUB}
\label{tab:supp_breakdown_trials_cub}
\resizebox{\textwidth}{!}{%
\begin{tabular}{@{}rrrrr@{}}
\toprule
\multicolumn{1}{c}{\textbf{T}} & \multicolumn{1}{c}{\textbf{\begin{tabular}[c]{@{}c@{}}Human\\ Performance\end{tabular}}} & \multicolumn{1}{c}{\textbf{\# Trials}} & \multicolumn{1}{c}{\textbf{\begin{tabular}[c]{@{}c@{}}\# Trials\\ Correct AI Prediction\end{tabular}}} & \multicolumn{1}{c}{\textbf{\begin{tabular}[c]{@{}c@{}}\# Trials\\ Wrong AI Prediction\end{tabular}}} \\ \midrule
0.00 & n/a & n/a & 0 & 1 \\
0.05 & n/a & n/a & 0 & 1 \\
0.10 & n/a & n/a & 0 & 1 \\
0.15 & 100 & 5 & 1 & 4 \\
0.20 & 76.47 & 17 & 1 & 16 \\
0.25 & 79.49 & 39 & 1 & 38 \\
0.30 & 83.93 & 56 & 1 & 55 \\
0.35 & 76.4 & 89 & 1 & 88 \\
0.40 & 72.17 & 115 & 1 & 114 \\
0.45 & 69.36 & 173 & 8 & 165 \\
0.50 & 65.38 & 234 & 19 & 215 \\
\cellcolor[HTML]{\thresholdcolor}\textbf{0.55} & \cellcolor[HTML]{\thresholdcolor}59.27 & \cellcolor[HTML]{\thresholdcolor}329 & \cellcolor[HTML]{\thresholdcolor}31 & \cellcolor[HTML]{\thresholdcolor}298 \\
0.60 & 60.78 & 408 & 46 & 362 \\
0.65 & 57.23 & 484 & 52 & 432 \\
0.70 & 54.56 & 570 & 65 & 505 \\
0.75 & 54.6 & 630 & 84 & 546 \\
0.80 & 52.55 & 685 & 96 & 589 \\
0.85 & 51.72 & 787 & 125 & 662 \\
0.90 & 51.91 & 865 & 151 & 714 \\
0.95 & 54.55 & 1056 & 271 & 785 \\
1.00 & 65.5 & 1800 & 900 & 900 \\ \bottomrule
\end{tabular}%
}
\end{table}

\begin{figure}[!htbp]
    \centering
    \includegraphics[width=\textwidth]{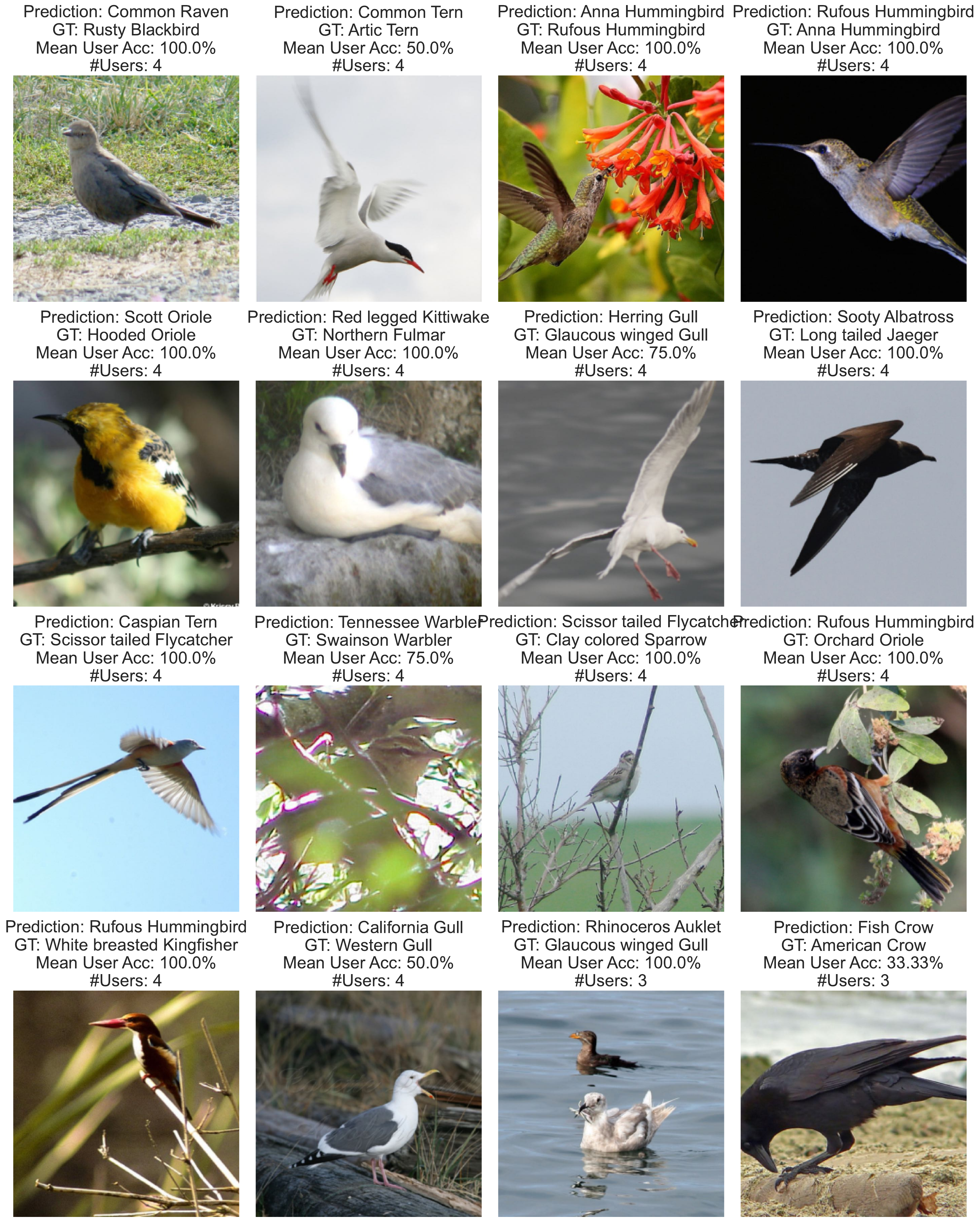}
    \caption{CUB samples at $T=0.55$ -- ResNet-50}
    \label{fig:supp_human_ait_team_cub_samples}
\end{figure}

\clearpage
\section{Analysis for CUB}

\subsection{Correspondences help users to reject wrong AI prediction}
\label{supp:when_xai_helps_group1}


\begin{figure}[!htbp]
        \begin{subfigure}[b]{1\textwidth}
                \includegraphics[width=\linewidth, trim=3cm .0cm .0cm .0cm,clip]{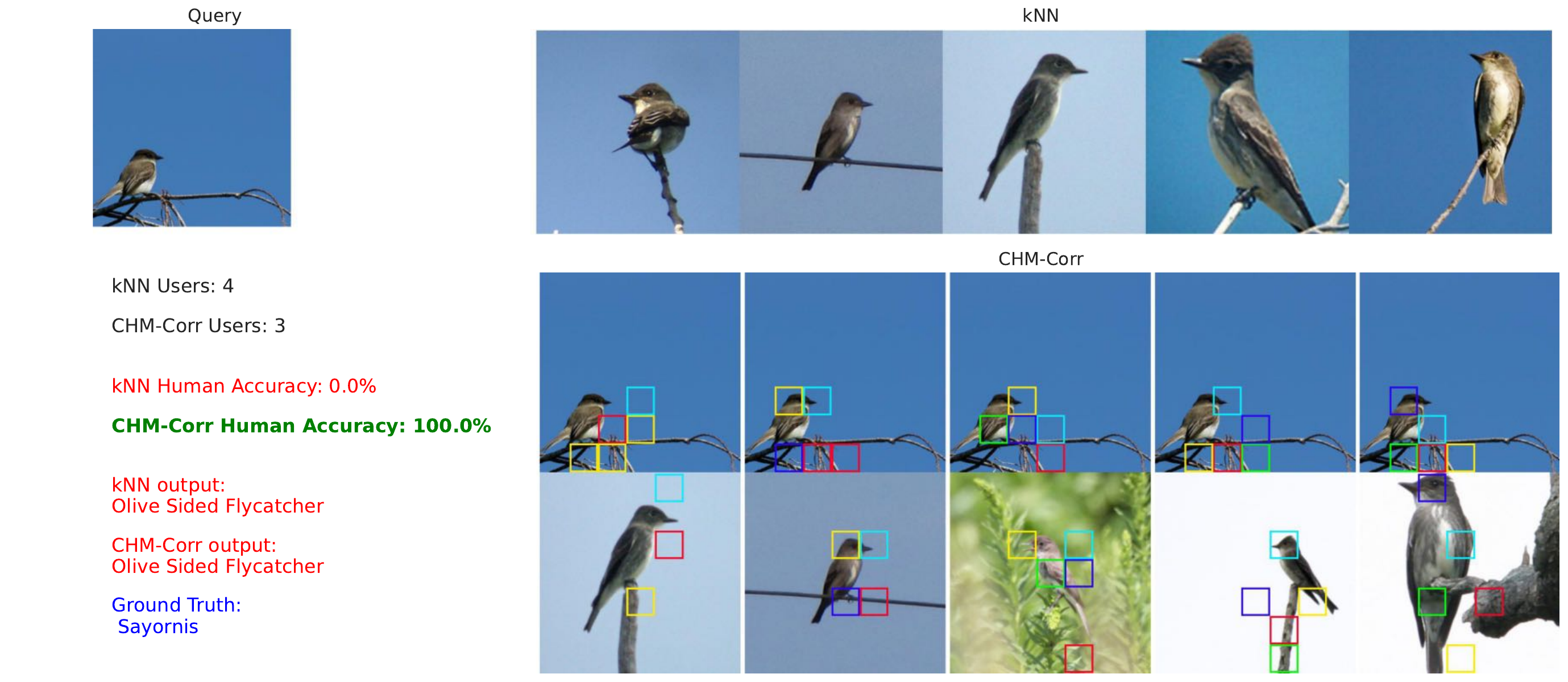}
                \caption{}
                \label{fig:when_xai_helps_group1_knn_chm_a}
        \end{subfigure}\par
        \begin{subfigure}[b]{1\textwidth}
                \includegraphics[width=\linewidth, trim=3cm .0cm .0cm .0cm,clip]{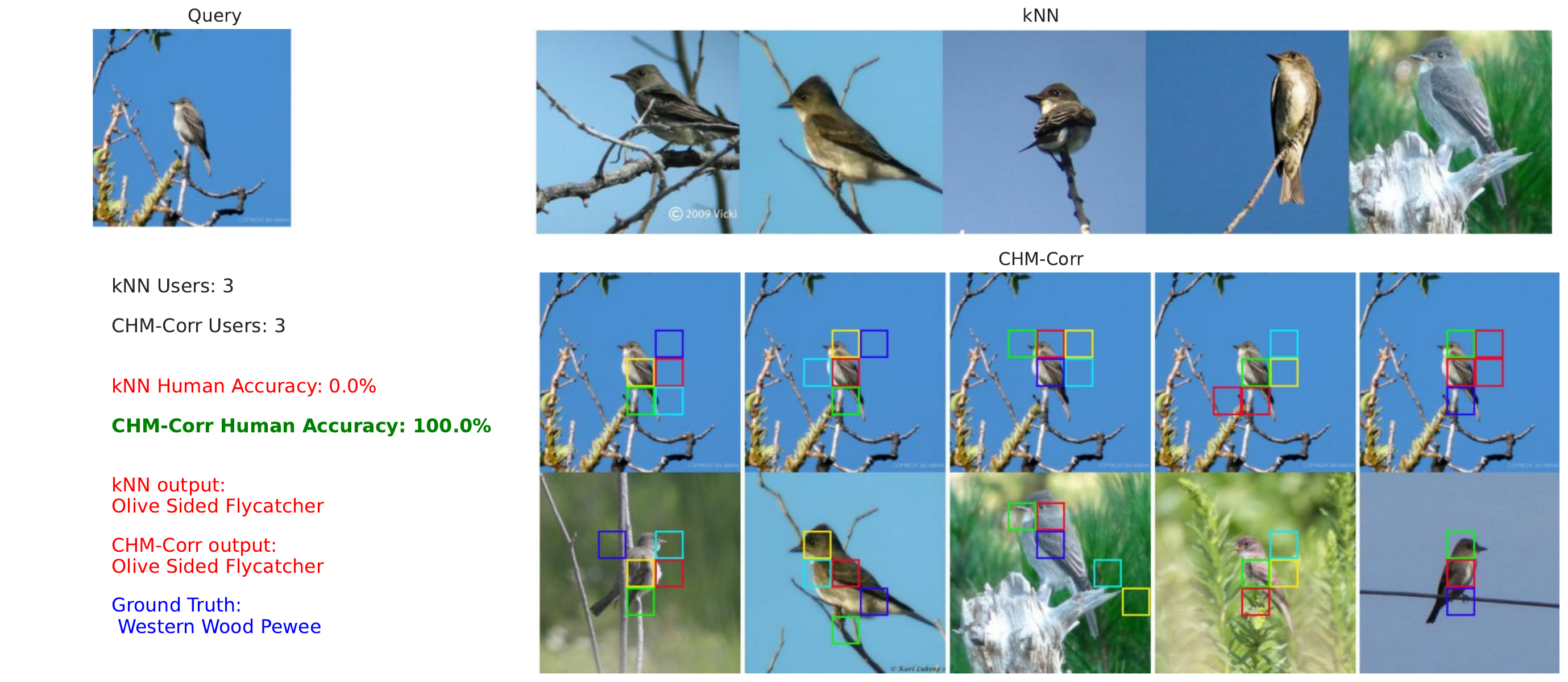}
                \caption{}
                \label{fig:when_xai_helps_group1_knn_chm_b}
        \end{subfigure}\par
        \caption{When correspondences help users to reject wrong AI prediction -- \textbf{(a)} Both kNN and CHM-Corr classifiers misclassified an image of \class{Sayornis} into \class{Olive Sided Flycatcher}. Using kNN explanations, 4/4 of users failed to reject this wrong prediction, while using CHM-Corr explanations, 3/3 of users successfully rejected AI decisions. \textbf{(b)} Both kNN and CHM-Corr classifiers misclassified an image of \class{Western Wood Pewee} into \class{Olive Sided Flycatcher}. Using kNN explanations, 3/3 of users failed to reject this wrong prediction, while using CHM-Corr explanations, 3/3 of users successfully rejected AI decisions.}
        \label{fig:when_xai_helps_group1_knn_chm}
\end{figure}

\begin{figure}[!htbp]
        \begin{subfigure}[b]{1\textwidth}
                \includegraphics[width=\linewidth, trim=3cm .0cm .0cm .0cm,clip]{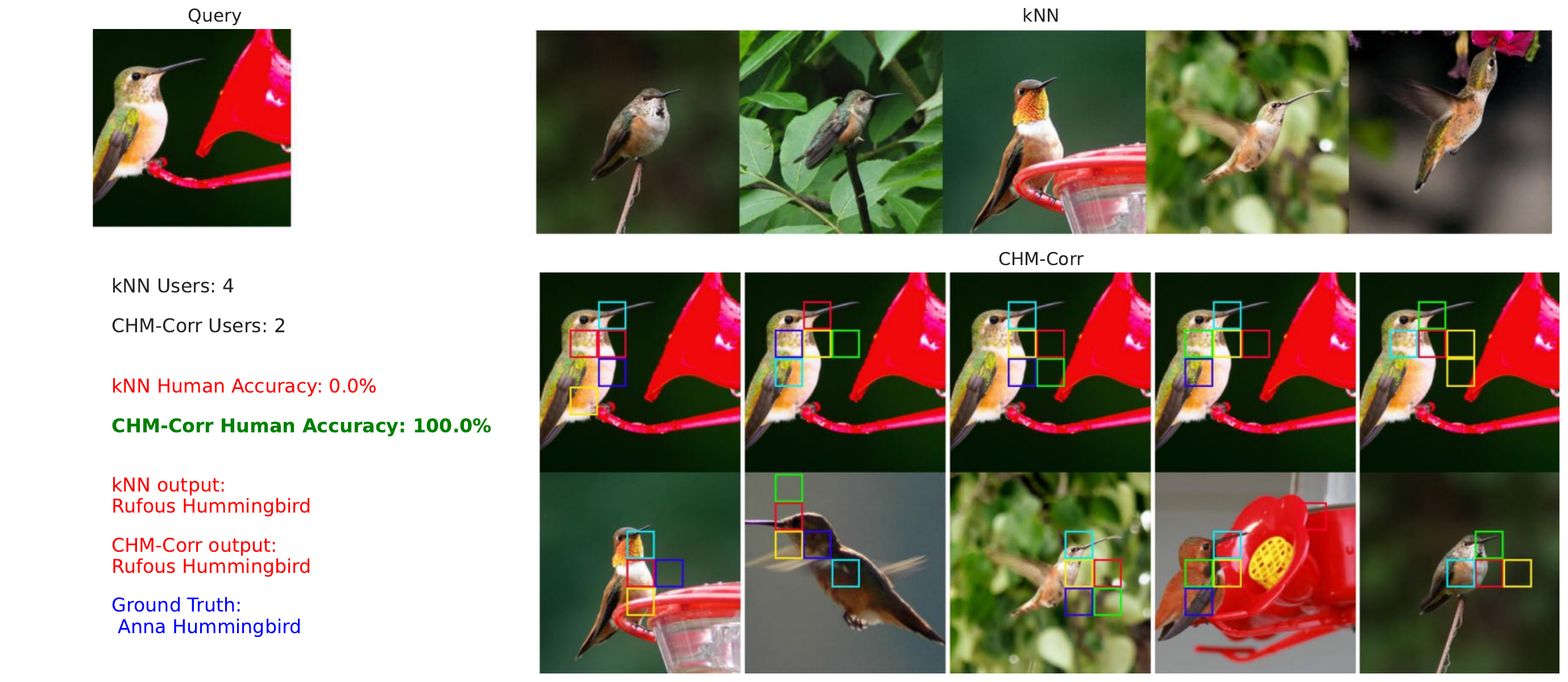}
                \caption{}
                \label{fig:when_xai_helps_group1_knn_chm_2_a}
        \end{subfigure}\par
        \begin{subfigure}[b]{1\textwidth}
                \includegraphics[width=\linewidth, trim=3cm .0cm .0cm .0cm,clip]{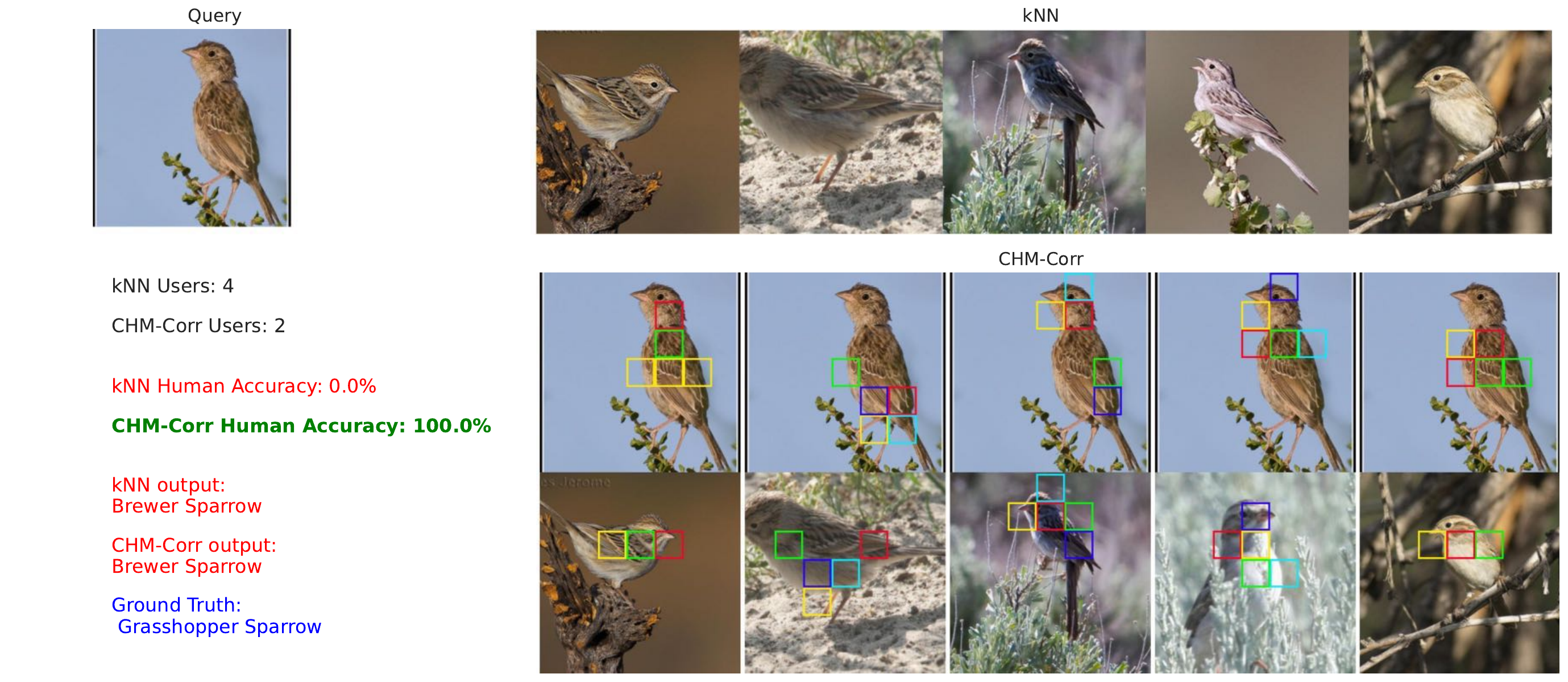}
                \caption{}
                \label{fig:when_xai_helps_group1_knn_chm_2_b}
        \end{subfigure}\par
        \caption{When correspondences help users to reject wrong AI prediction -- \textbf{(a)} Both kNN and CHM-Corr classifiers misclassified an image of \class{Anna Hummingbird} as a \class{Rufous Hummingbird}. Using kNN explanations, 4/4 of users failed to reject this wrong prediction, while using CHM-Corr explanations, 2/2 of users successfully rejected AI decisions. \textbf{(b)} Both kNN and CHM-Corr classifiers misclassified an image of \class{Grasshopper Sparrow} as a \class{Brewer Sparrow}. Using kNN explanations, 4/4 of users failed to reject this wrong prediction, while using CHM-Corr explanations, 2/2 of users successfully rejected AI decisions.}
        \label{fig:when_xai_helps_group1_knn_chm_2}
\end{figure}

\begin{figure}[!htbp]
        \begin{subfigure}[b]{1\textwidth}
                \includegraphics[width=\linewidth, trim=3cm .0cm .0cm .0cm,clip]{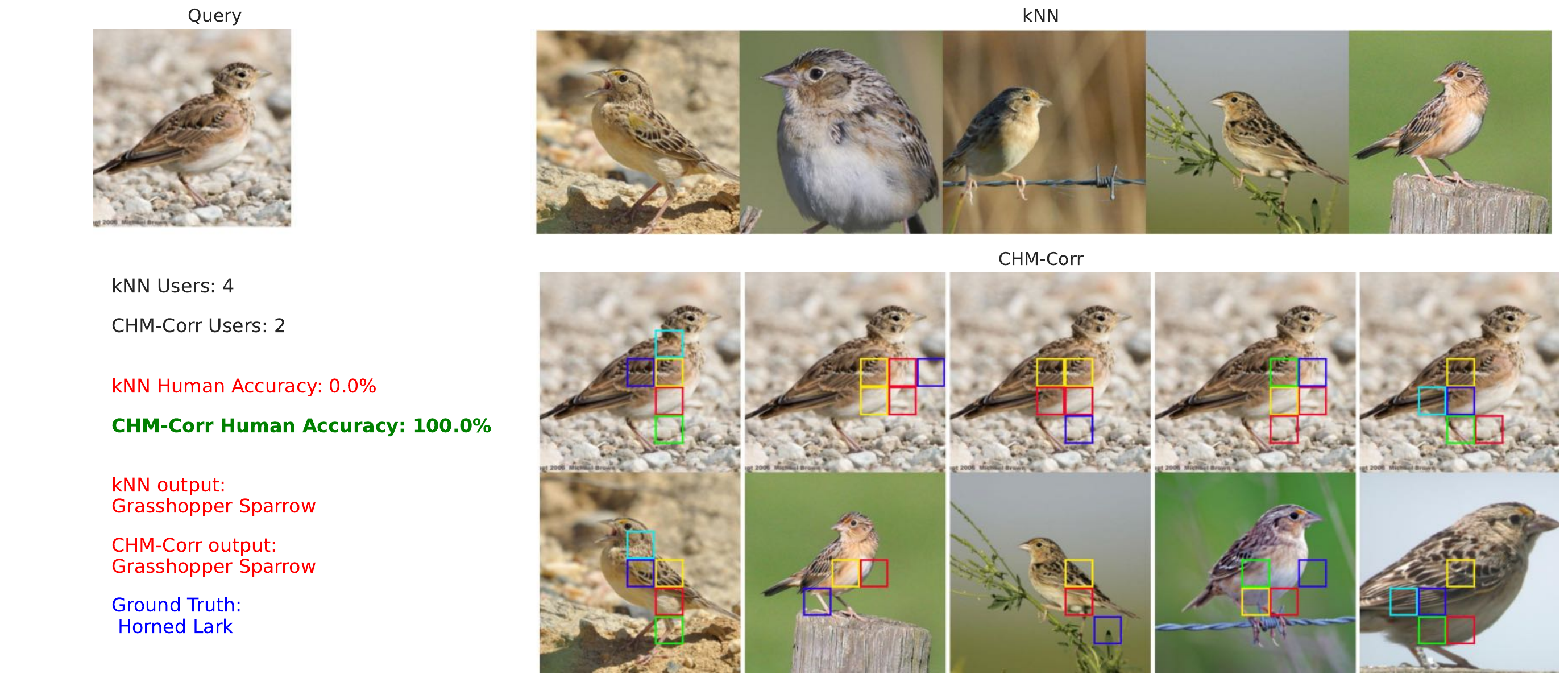}
                \caption{}
                \label{fig:when_xai_helps_group1_knn_chm_3_a}
        \end{subfigure}\par
        \begin{subfigure}[b]{1\textwidth}
                \includegraphics[width=\linewidth, trim=3cm .0cm .0cm .0cm,clip]{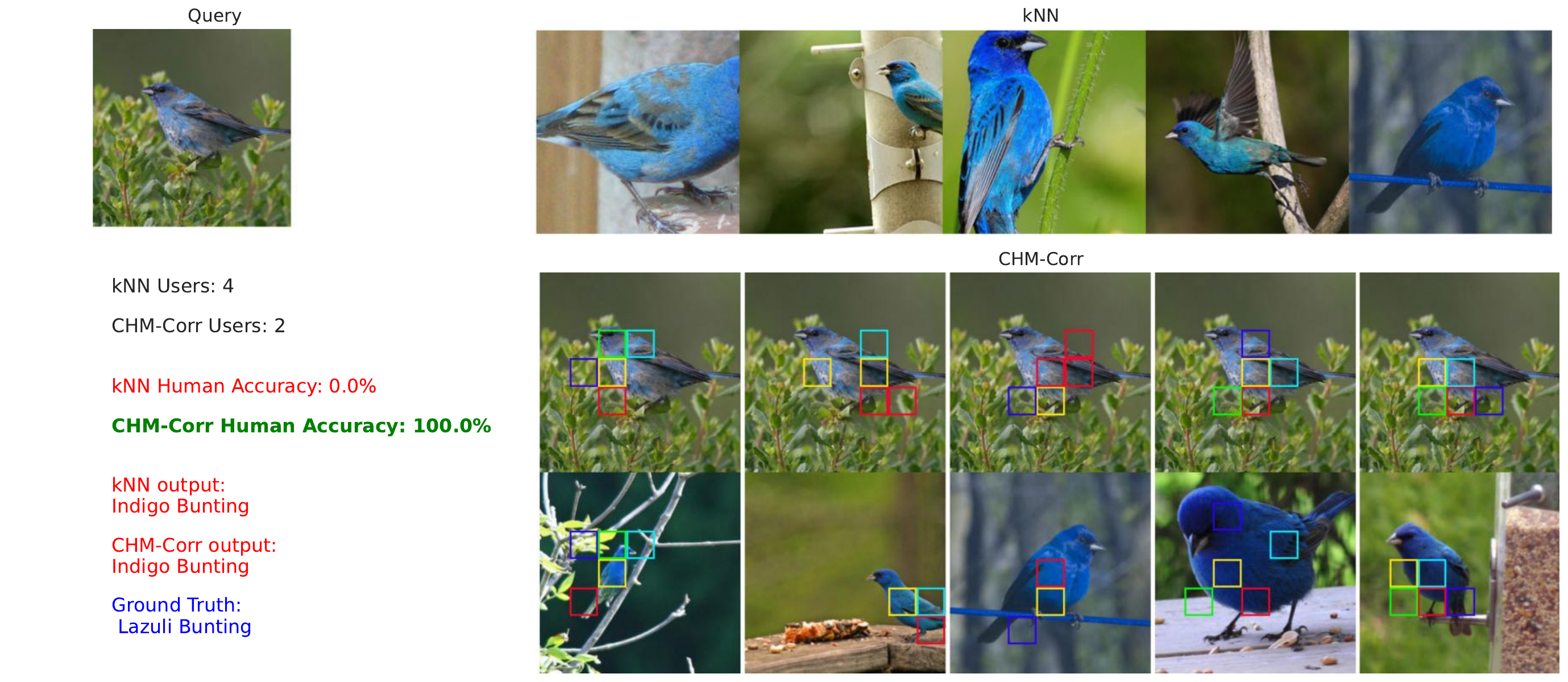}
                \caption{}
                \label{fig:when_xai_helps_group1_knn_chm_3_b}
        \end{subfigure}\par
        \caption{When correspondences help users to reject wrong AI prediction -- \textbf{(a)} Both kNN and CHM-Corr classifiers misclassified an image of \class{Horned Lark} as a \class{Grasshopper Sparrow}. Using kNN explanations, 4/4 of users failed to reject this wrong prediction, while using CHM-Corr explanations, 2/2 of users successfully rejected AI decisions. \textbf{(b)} Both kNN and CHM-Corr classifiers misclassified an image of \class{Lazuli Bunting} as an \class{Indigo Bunting}. Using kNN explanations, 4/4 of users failed to reject this wrong prediction, while using CHM-Corr explanations, 2/2 of users successfully rejected AI decisions.}
        \label{fig:when_xai_helps_group1_knn_chm_3}
\end{figure}

\begin{figure}[!htbp]
        \begin{subfigure}[b]{1\textwidth}
                \includegraphics[width=\linewidth, trim=3cm .0cm .0cm .0cm,clip]{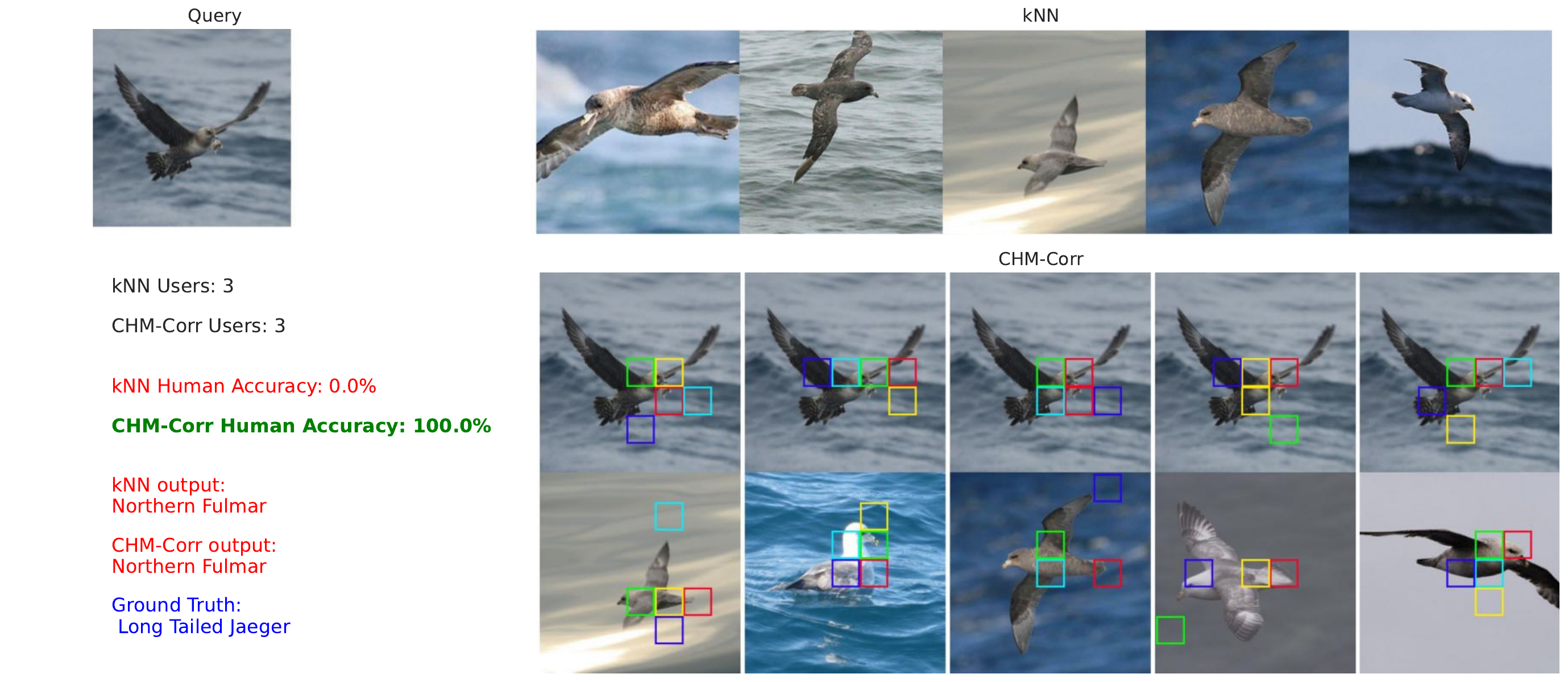}
                \caption{}
                \label{fig:when_xai_helps_group1_knn_chm_4_a}
        \end{subfigure}\par
        \begin{subfigure}[b]{1\textwidth}
                \includegraphics[width=\linewidth, trim=3cm .0cm .0cm .0cm,clip]{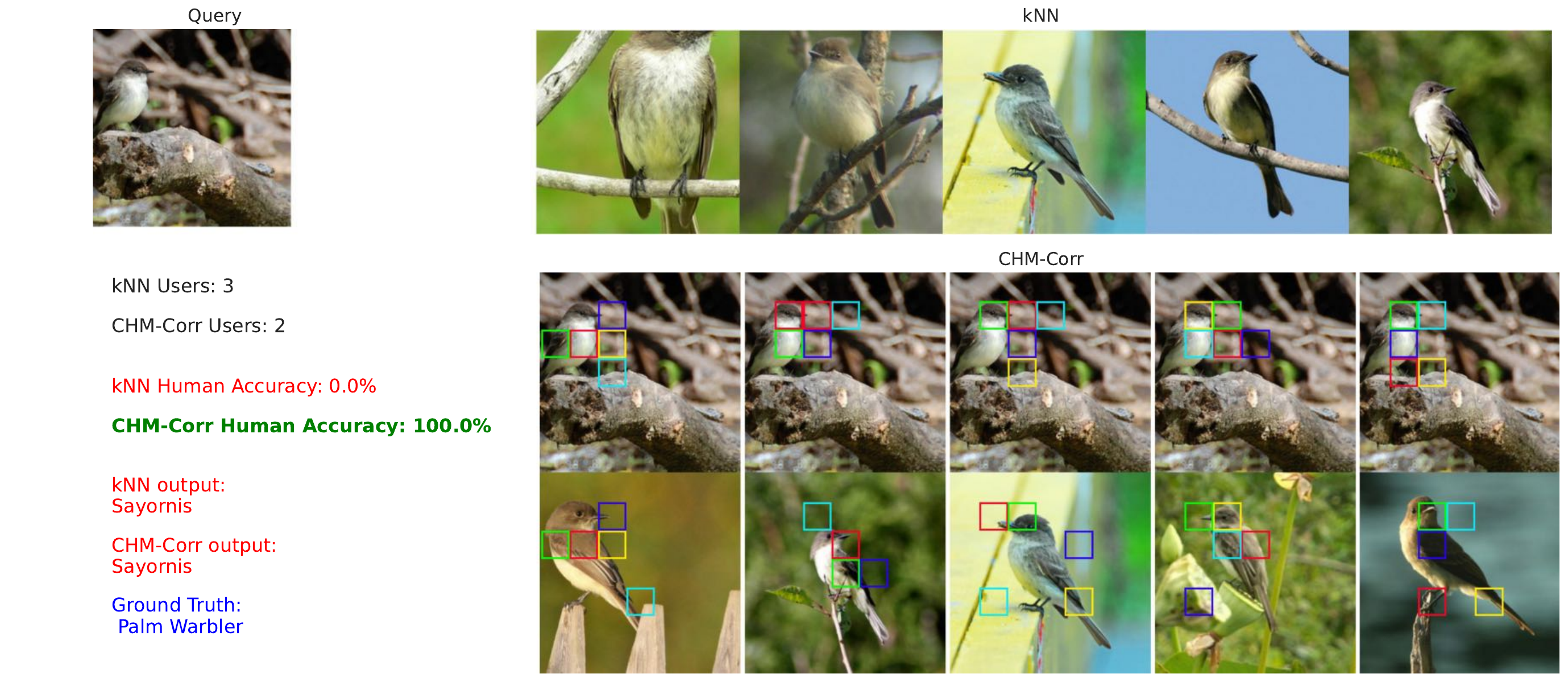}
                \caption{}
                \label{fig:when_xai_helps_group1_knn_chm_4_b}
        \end{subfigure}\par
        \caption{When correspondences help users to reject wrong AI prediction -- \textbf{(a)} Both kNN and CHM-Corr classifiers misclassified an image of \class{Long Tailed Jaege} as a \class{Northern Fulmar}. Using kNN explanations, 3/3 of users failed to reject this wrong prediction, while using CHM-Corr explanations, 3/3 of users successfully rejected AI decisions. \textbf{(b)} Both kNN and CHM-Corr classifiers misclassified an image of \class{Palm Warbler} as a \class{Sayornis}. Using kNN explanations, 3/3 of users failed to reject this wrong prediction, while using CHM-Corr explanations, 2/2 of users successfully rejected AI decisions.}
        \label{fig:when_xai_helps_group1_knn_chm_4}
\end{figure}

\clearpage

\subsection{Diversity of images in kNN and, EMD-Corr, and CHM-Corr explanations}
\label{sec:diversity_test}

We hypothesize that when the AI prediction is wrong, the diversity among the five nearest neighbors of kNN differs from EMD-Corr and CHM-Corr, leading to users rejecting the decision. To this end, we calculated LPIPS and MS-SSIM metrics between all possible pairs of explanations on the relevant queries.

\begin{figure}[!htb]
        \begin{subfigure}[b]{0.5\textwidth}
                \includegraphics[width=\linewidth]{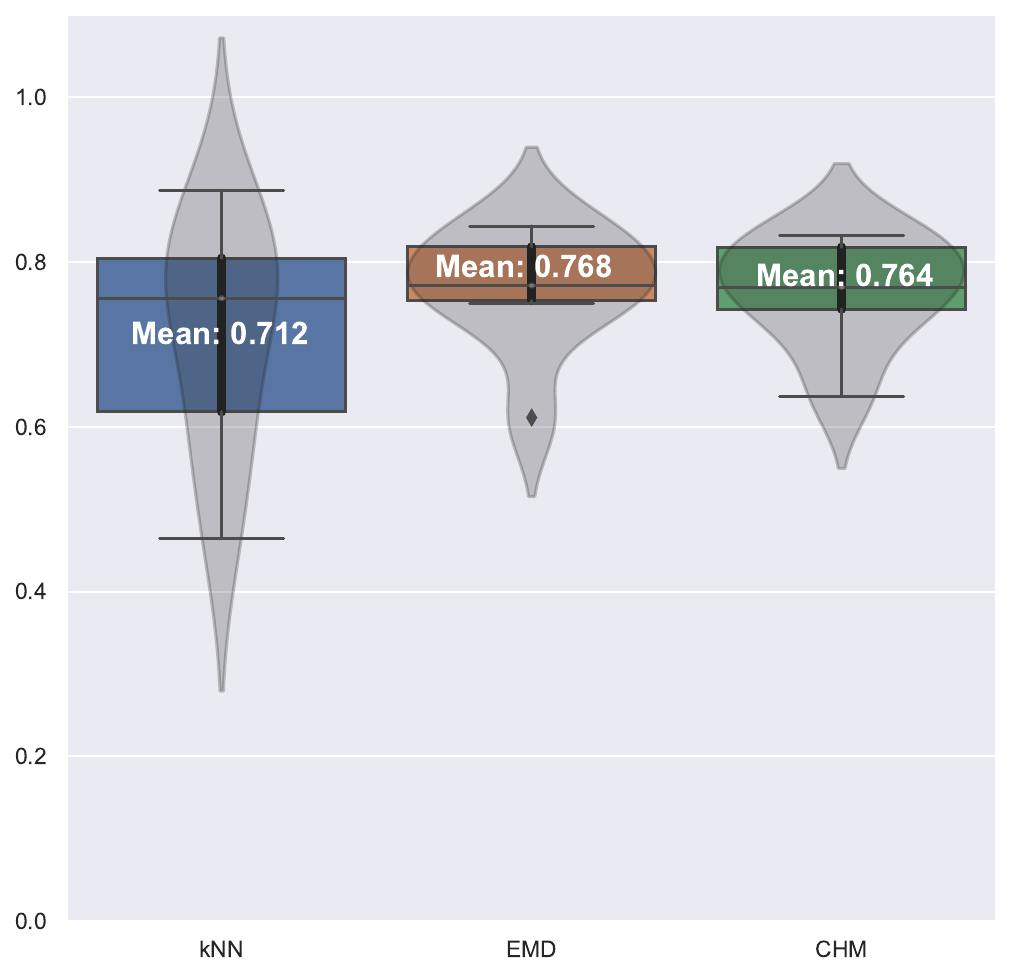}
                \caption{LPIPS score. Higher is more diverse.}
                \label{fig:lpips}
        \end{subfigure}%
        \begin{subfigure}[b]{0.5\textwidth}
                \includegraphics[width=\linewidth]{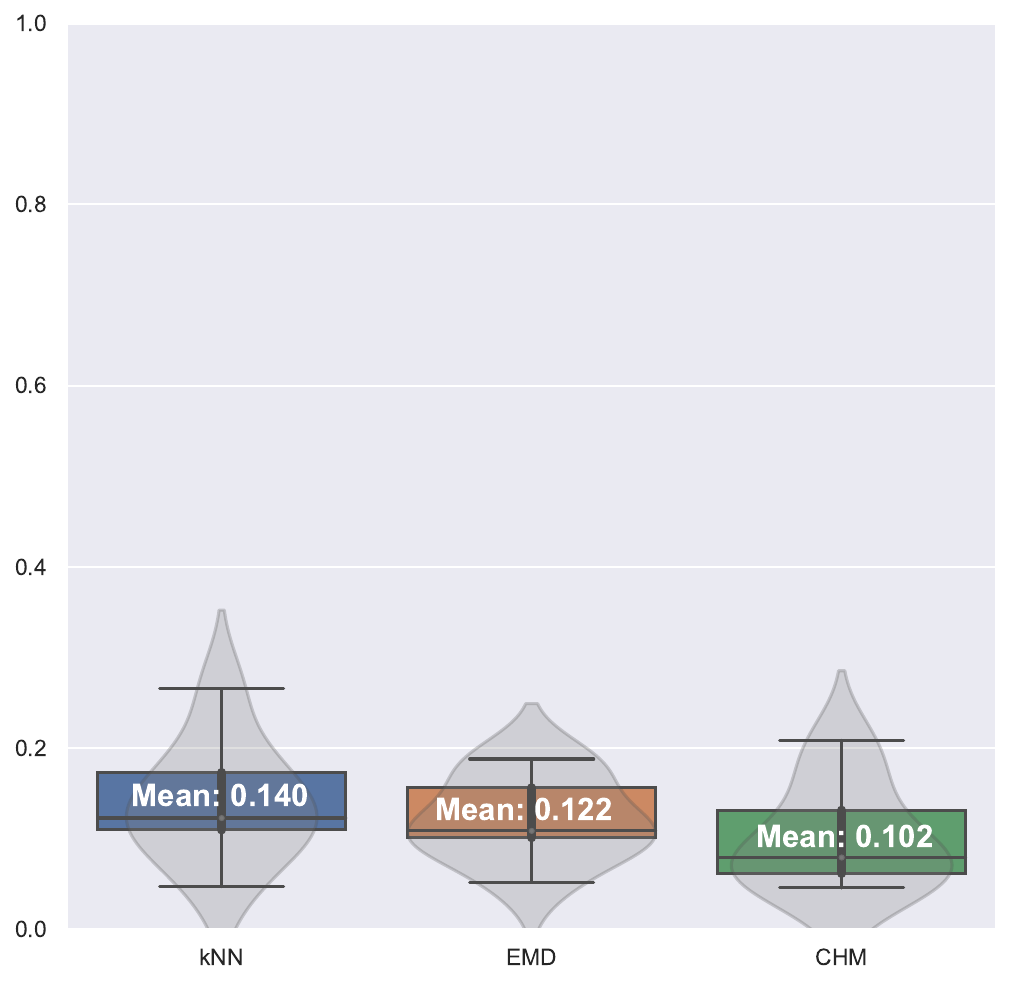}
                \caption{MS-SSIM score. Lower is more diverse.}
                \label{fig:msssim}
        \end{subfigure}%
        \caption{Analysis of the diversity between all 10 possible pairs of five nearest neighbors for queries with the average user's accuracy of 0\% when kNN explanation is provided and the average user's accuracy of 100\% when CHM-Corr explanation is provided (CUB).
        \textbf{The images in kNN explanations are consistently less diverse under both LPIPS (a) and MS-SSIM (b) than those in EMD-Corr and CHM-Corr explanations.}
        That is, this is evidence explaining why kNN users tend to be fooled into accepting kNN wrong decisions the most.
        }
        \label{fig:lpips_and_mmssim}
\end{figure}

\clearpage
\subsection{When the user rejects the correct AI prediction}

This section provides a brief qualitative explanation for the cases where users incorrectly rejected a correct AI prediction.

\begin{figure*}[!hbt]

  \centering
    \begin{subfigure}[b]{\textwidth}
        \centering
        \includegraphics[width=1.0\textwidth, trim=2cm .0cm .0cm .0cm,clip]{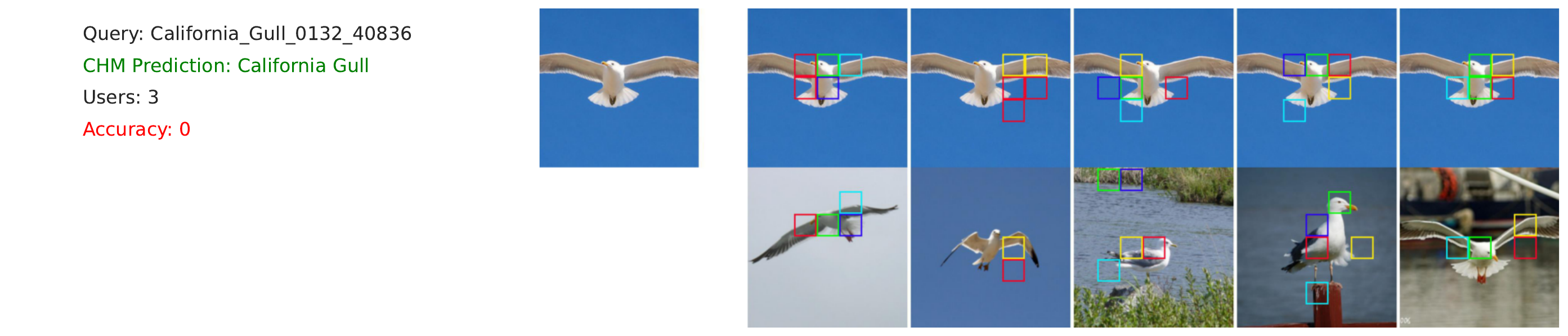}
        \caption{The CHM-Corr classifier missed tips at wings, and the legs’ black tips were occluded.}
        \label{fig:group2_a}
    \end{subfigure}
    
        \centering
    \begin{subfigure}[b]{\textwidth}
        \centering
        \includegraphics[width=1.0\textwidth, trim=2cm .0cm .0cm .0cm,clip]{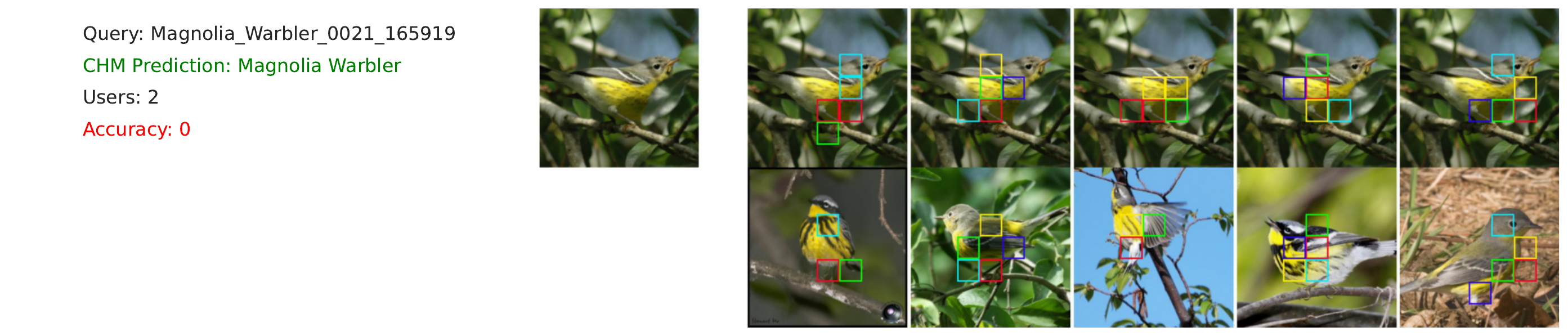}
        \caption{The CHM-Corr classifier missed the stripes at the belly.}
        \label{fig:group2_b}
    \end{subfigure}

    \centering
    \begin{subfigure}[b]{\textwidth}
        \centering
        \includegraphics[width=1.0\textwidth, trim=2cm .0cm .0cm .0cm,clip]{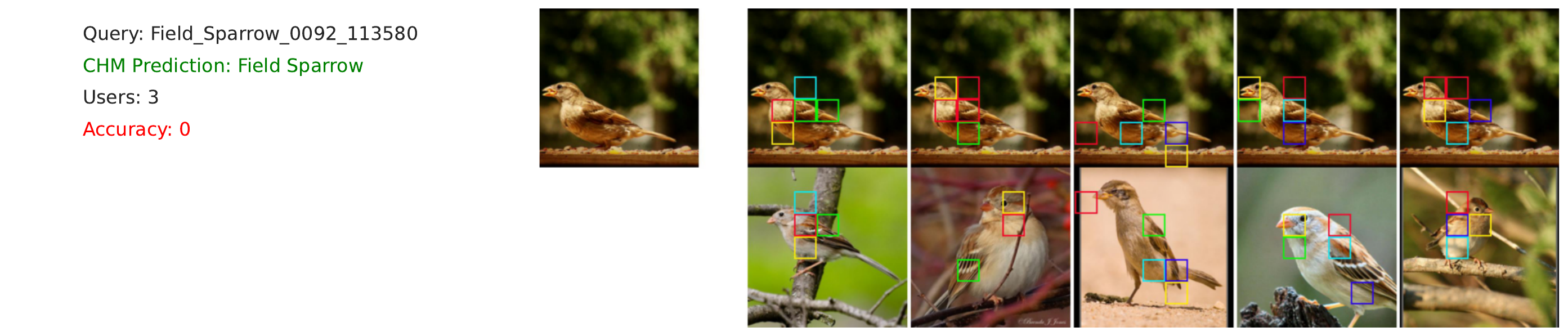}
        \caption{Low-quality query -- No distinctive features can be recognized from the input image.}
        \label{fig:group2_c}
    \end{subfigure}
    
    \centering
    \begin{subfigure}[b]{\textwidth}
        \centering
        \includegraphics[width=1.0\textwidth, trim=2cm .0cm .0cm .0cm,clip]{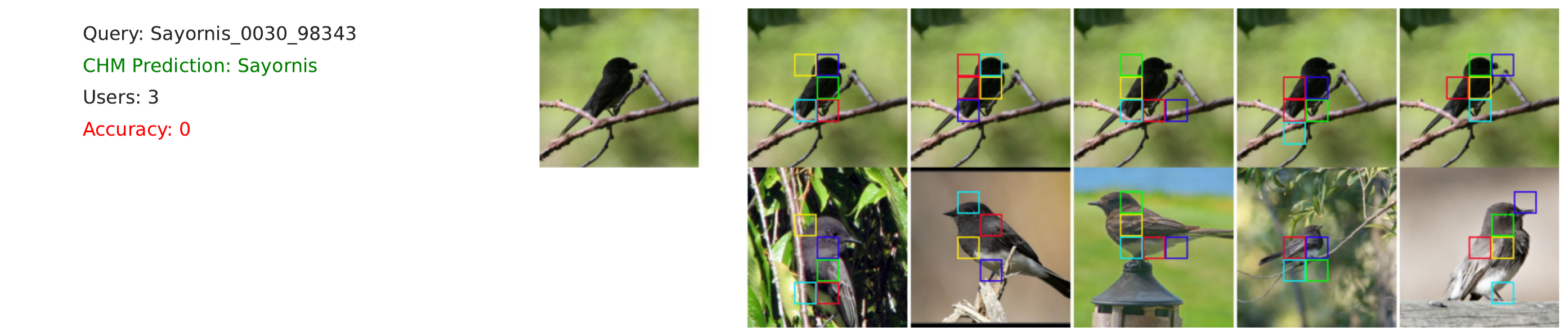}
        \caption{Low-quality query -- No distinctive features can be recognized from the input image.}
        \label{fig:group2_d}
    \end{subfigure}

    \centering
    \begin{subfigure}[b]{\textwidth}
        \centering
        \includegraphics[width=1.0\textwidth, trim=2cm .0cm .0cm .0cm,clip]{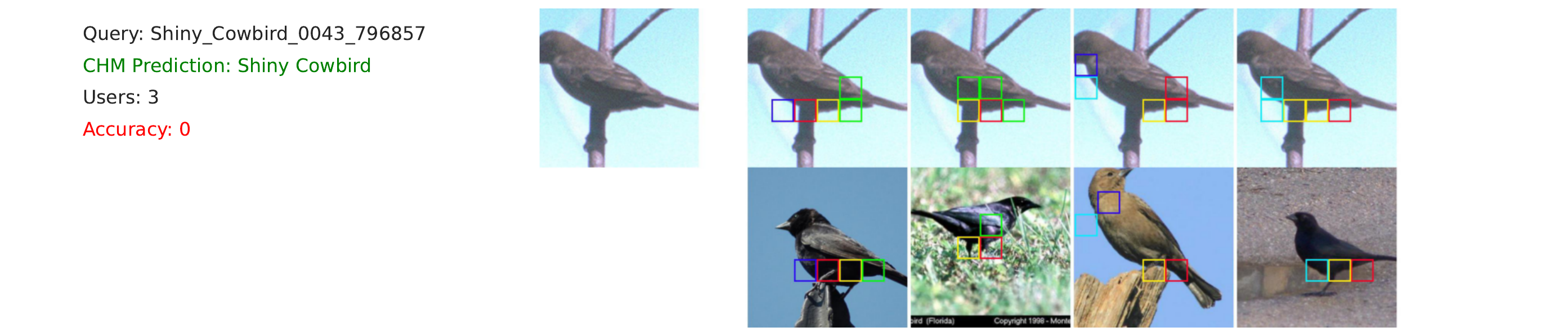}
        \caption{Low-quality query -- No distinctive features can be recognized from the input image.}
        \label{fig:group2_e}
    \end{subfigure}
    
    \caption{Analysis of queries that user's rejected correct CHM-Corr prediction.}
    \label{fig:group2}
\end{figure*}

\clearpage

\clearpage
\section{Comparing explanation methods}
\label{appendix:compare_methods}

This section compares explanations provided by kNN, EMD-Corr, and CHM-Corr for various sets of queries.

\subsection{ImageNet samples}

\begin{figure*}[!hbt]

  \centering
    \begin{subfigure}[b]{\textwidth}
        \centering
        \includegraphics[width=1.0\textwidth]{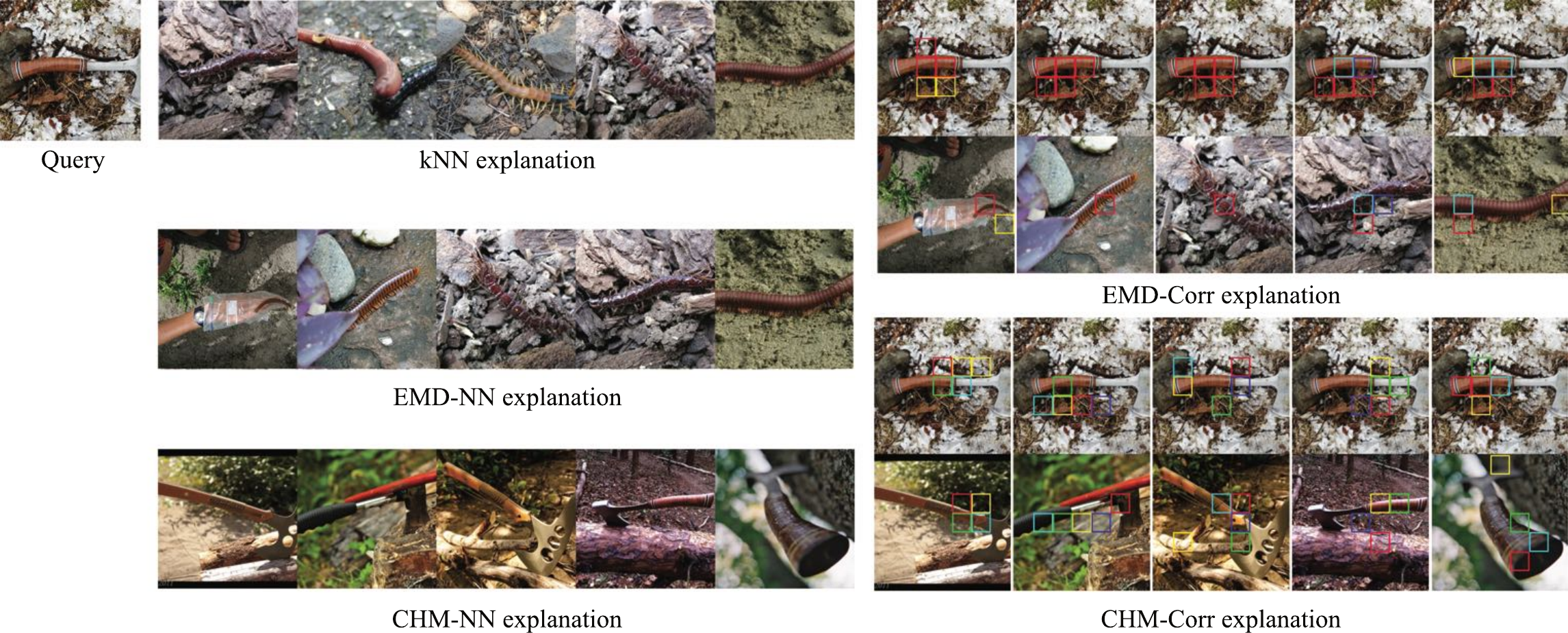}
    \end{subfigure}
    \caption{The kNN and EMD-Corr \textcolor{red}{misclassify} an image of \class{hatchet} as a \class{centipede}. The CHM-Corr \textcolor{ForestGreen}{correctly} classifies this image.}
    \label{fig:placeholder_worm}
\end{figure*}

\begin{figure*}[!hbt]

  \centering
    \begin{subfigure}[b]{\textwidth}
        \centering
        \includegraphics[width=1.0\textwidth]{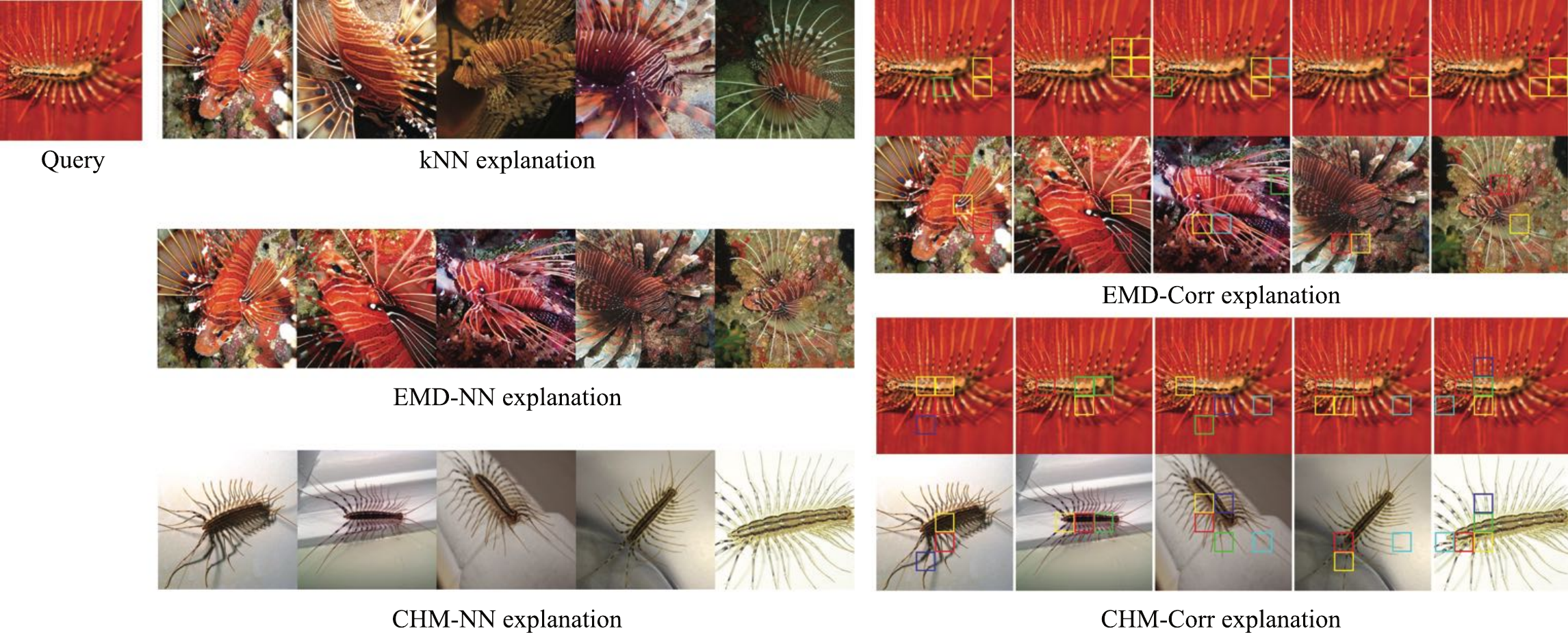}
    \end{subfigure}
    \caption{The kNN and EMD-Corr \textcolor{red}{misclassify} an image of \class{centipede} as a \class{lionfish} due to the dominant red color in the background. The CHM-Corr \textcolor{ForestGreen}{correctly} classifies this image.}
    \label{fig:placeholder_lionfish}
\end{figure*}

\begin{figure*}[!hbt]

  \centering
    \begin{subfigure}[b]{\textwidth}
        \centering
        \includegraphics[width=1.0\textwidth]{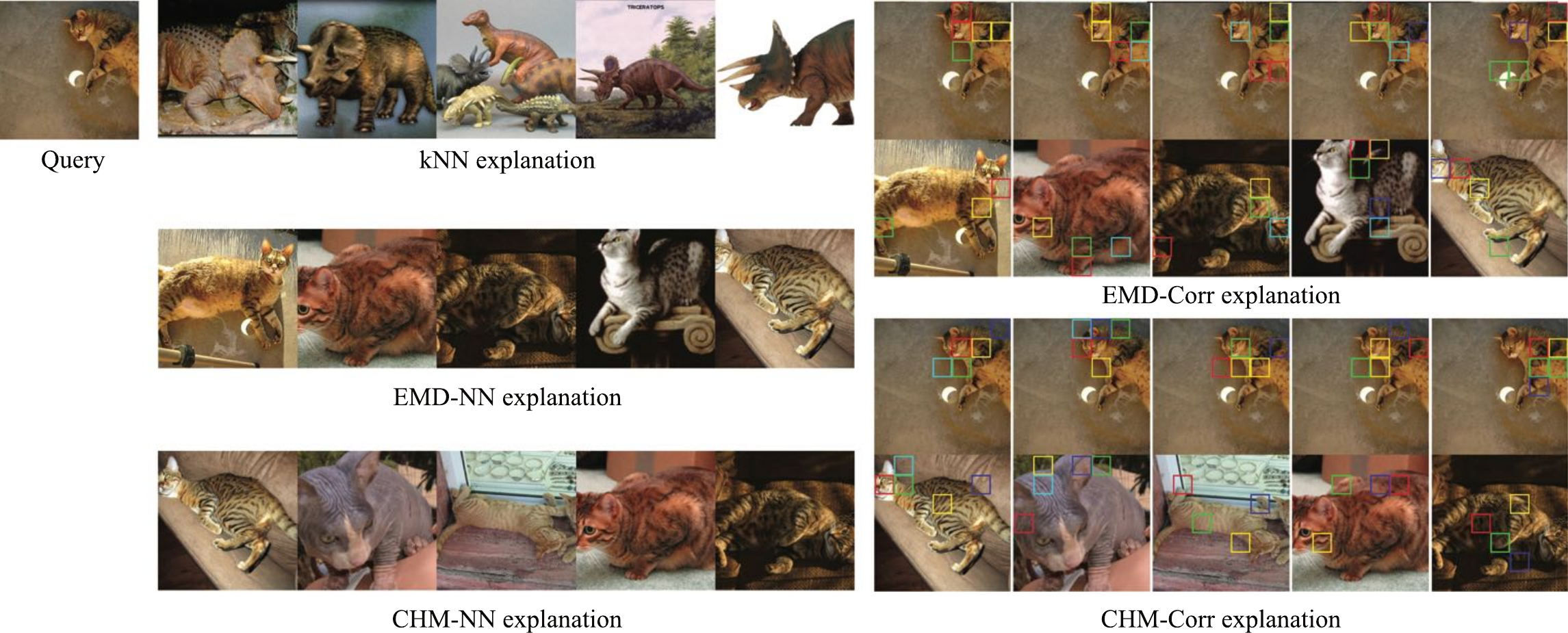}
    \end{subfigure}
    \caption{The kNN \textcolor{red}{misclassifies} an ImageNet image of \class{tiger cat} into \class{triceratops}. The EMD-Corr and CHM-Corr are both correctly classifying this image.}
    \label{fig:placeholder_triceratops}
\end{figure*}

\begin{figure*}[!hbt]

  \centering
    \begin{subfigure}[b]{\textwidth}
        \centering
        \includegraphics[width=1.0\textwidth]{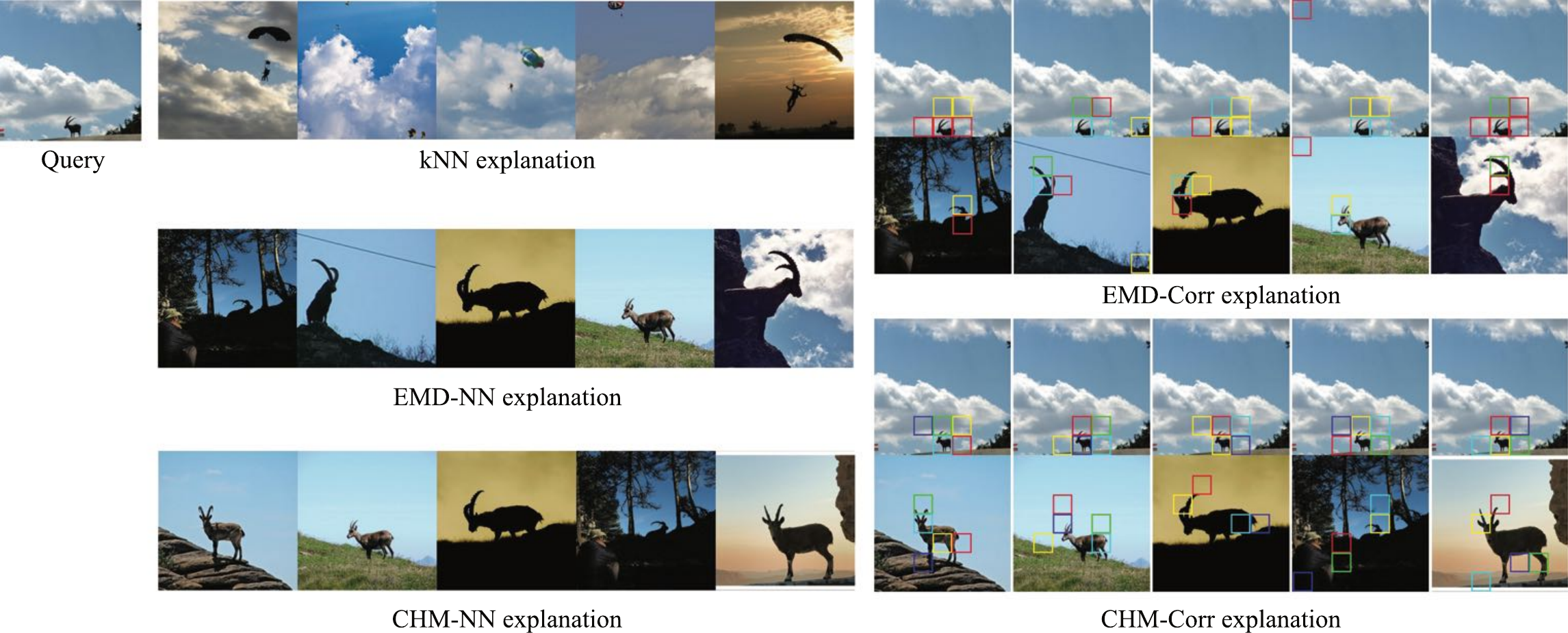}
    \end{subfigure}
    \caption{The kNN \textcolor{red}{misclassifies} an image of \class{ibex} as a \class{parachute} due to the dominant background. The EMD-Corr and CHM-Corr are both \textcolor{ForestGreen}{correctly} classifying this image.}
    \label{fig:placeholder_ibex}
\end{figure*}

\clearpage
\subsection{Adversarial samples}
\label{supp:adversarials}

\begin{figure*}[!hbt]

  \centering
    \begin{subfigure}[b]{\textwidth}
        \centering
        \includegraphics[width=1.0\textwidth]{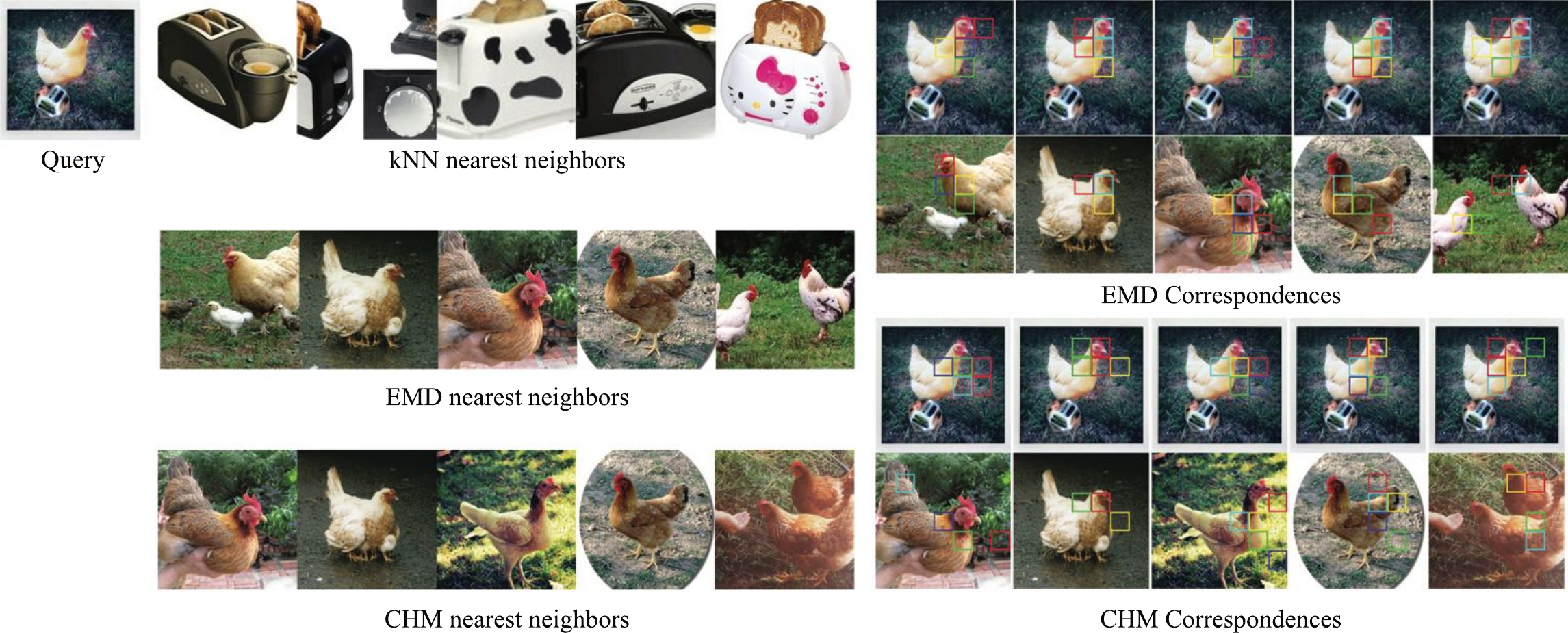}
    \end{subfigure}
    \caption{The kNN \textcolor{red}{misclassifies} an image of \class{hen} as a \class{toaster} due to an adversarial patch. The EMD-Corr and CHM-Corr are both \textcolor{ForestGreen}{correctly} classifying this image.}
    \label{fig:placeholder_hen}
\end{figure*}

\begin{figure*}[!hbt]

  \centering
    \begin{subfigure}[b]{\textwidth}
        \centering
        \includegraphics[width=1.0\textwidth]{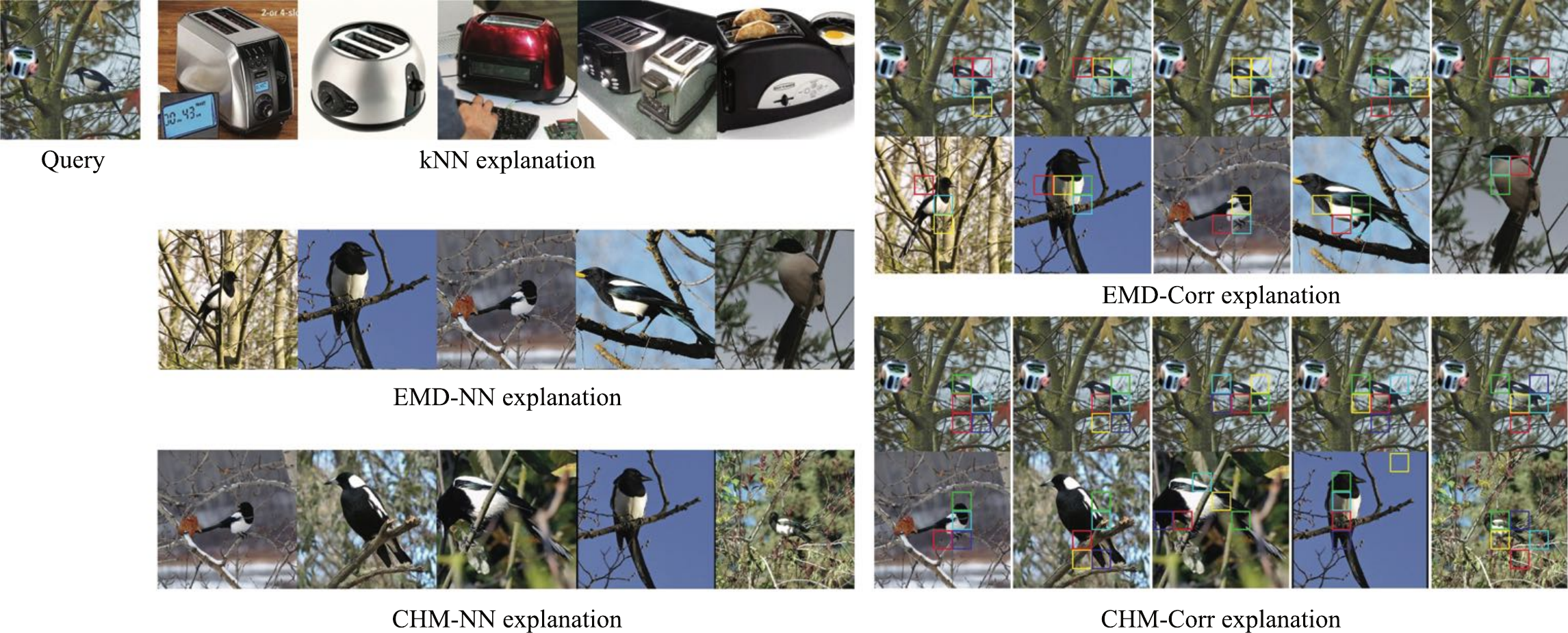}
    \end{subfigure}
    \caption{The kNN \textcolor{red}{misclassifies} an image of \class{magpie} as a \class{toaster} due to adversarial patch. The EMD-Corr and CHM-Corr are both \textcolor{ForestGreen}{correctly} classifying this image.}
    \label{fig:placeholder_magpie}
\end{figure*}

\clearpage
\section{Controlling keypoints in CHM-Corr+ for the CUB dataset}
\label{supp:comapre_method4_methdo4_plus}

Here, we compare CHM-Corr and CHM-Corr+ classifiers to understand the low performance of CHM-Corr+.
Using a set of \textbf{five} keypoints may not help CHM-Corr+ focus on the right patches. Sometimes, the five provided keypoints are not among the discriminative features to correctly classify a bird.



\begin{figure*}[!hbt]

  \centering
    \begin{subfigure}[b]{\textwidth}
        \centering
        \includegraphics[width=1.0\textwidth]{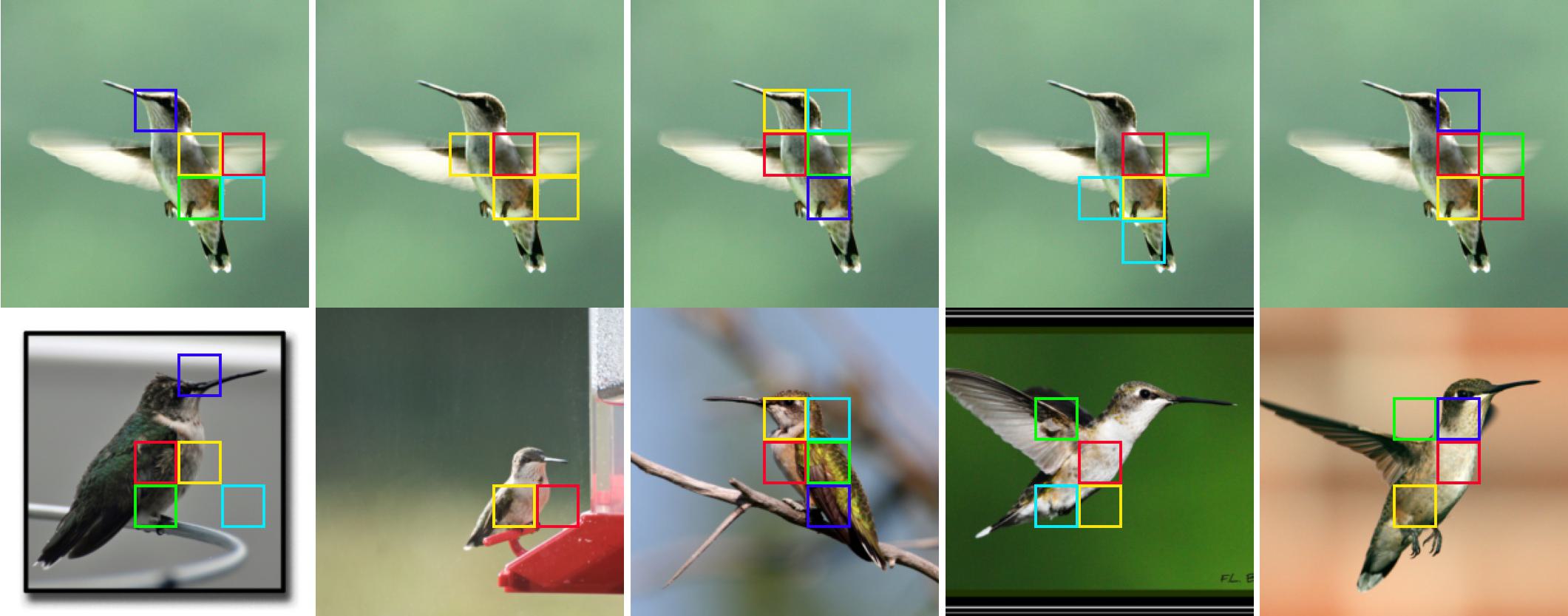}
        \caption{The explanation of a \textcolor{OliveGreen}{correct} classification by CHM-Corr.}
        \label{fig:method4_method4plus1_a}
    \end{subfigure}
    
    \centering
    \begin{subfigure}[b]{\textwidth}
        \centering
        \includegraphics[width=1.0\textwidth]{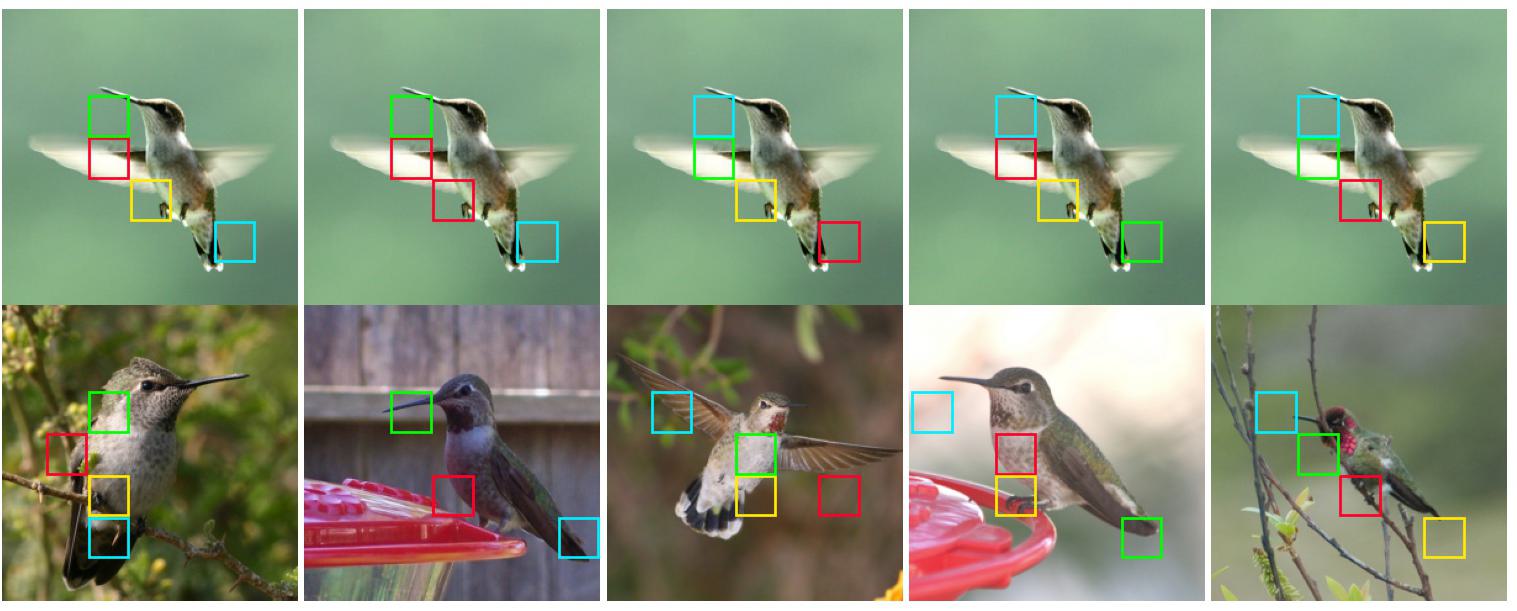}
        \caption{The explanation of a \textcolor{red}{misclassification} by CHM-Corr+}
        \label{fig:method4_method4plus1_b}
    \end{subfigure}
    
    \caption{A \class{Ruby-throated Hummingbird} \textcolor{red}{misclassified} into \class{Anna Hummingbird} by CHM-Corr+. 
    An example of low-quality keypoints leading to selecting and comparing mostly background (uninformative) patches.}
    \label{fig:method4_method4plus1}
\end{figure*}


\begin{figure*}[!hbt]

  \centering
    \begin{subfigure}[b]{\textwidth}
        \centering
        \includegraphics[width=1.0\textwidth]{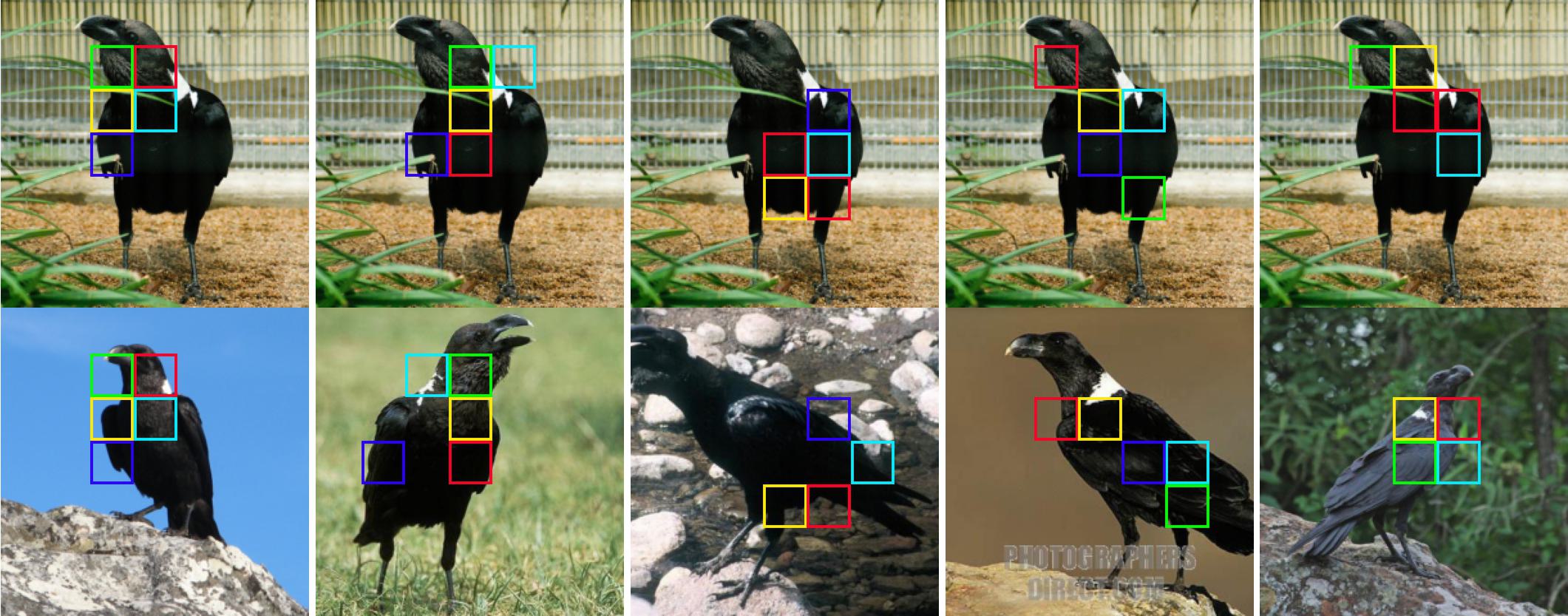}
        \caption{The explanation of a \textcolor{OliveGreen}{correct} classification by CHM-Corr.}
    \end{subfigure}
    
    \centering
    \begin{subfigure}[b]{\textwidth}
        \centering
        \includegraphics[width=1.0\textwidth]{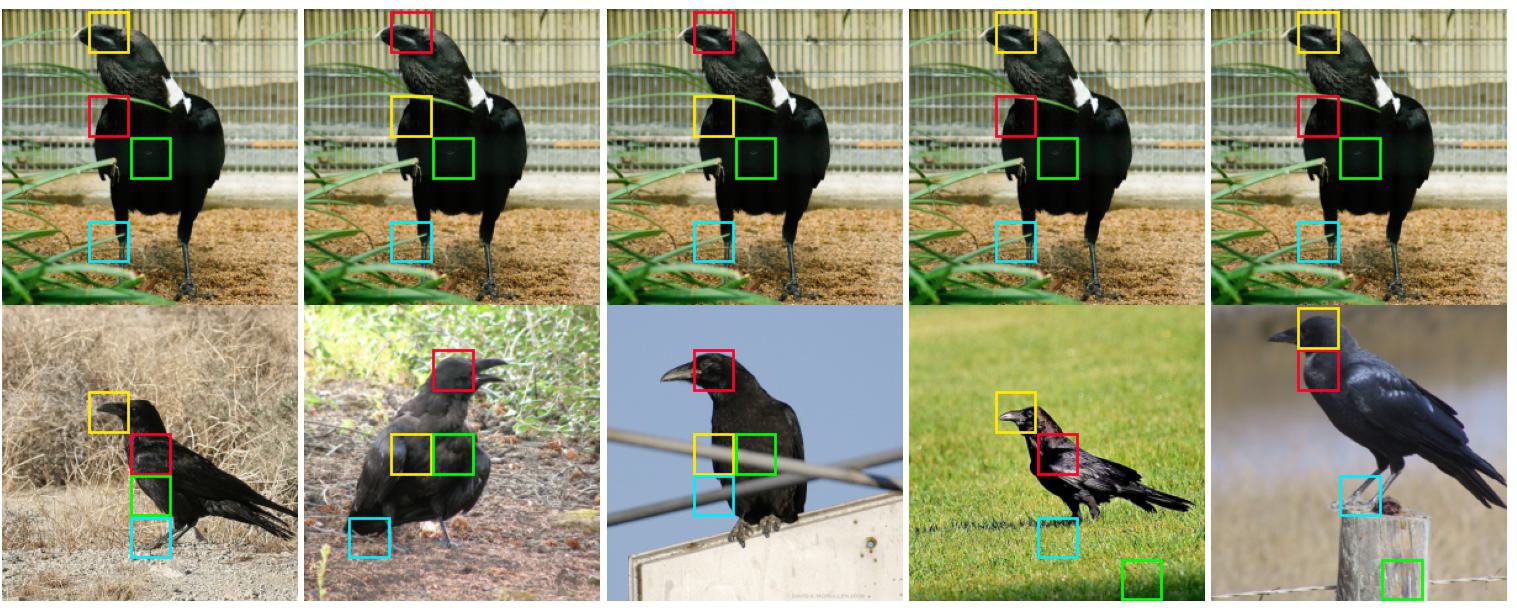}
        \caption{The explanation of a \textcolor{red}{misclassification} by CHM-Corr+.}
    \end{subfigure}
    
    \caption{A \class{White necked Raven} \textcolor{red}{misclassified} as a \class{Common Raven} by CHM-Corr+ -- The distinctive part of the bird is `the white feathers on the neck', which is missed in the keypoints selection step in the CHM-Corr+. The CHM-Corr classifier correctly classifies this image.}
    \label{fig:method4_method4plus2}
\end{figure*}

\begin{figure*}[!hbt]

  \centering
    \begin{subfigure}[b]{\textwidth}
        \centering
        \includegraphics[width=1.0\textwidth]{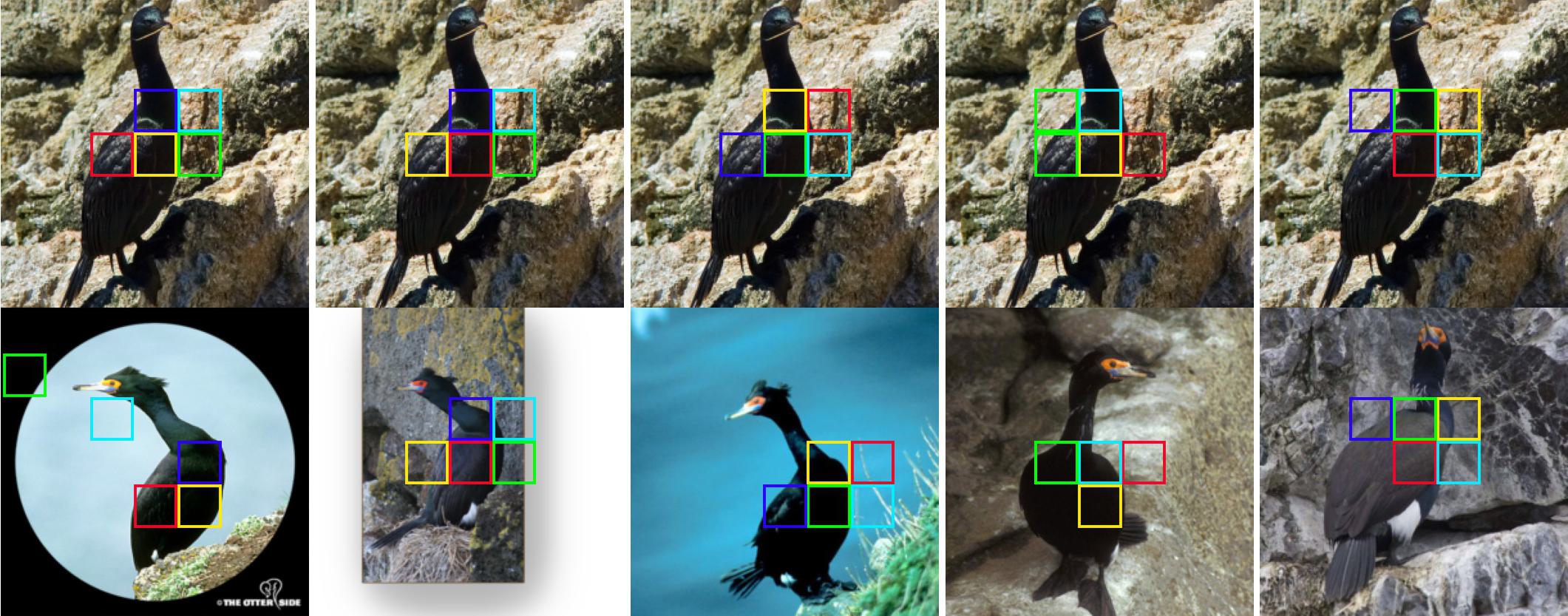}
        \caption{The explanation of a \textcolor{red}{misclassification} by CHM-Corr}
        \label{fig:method4_method4plus3_a}
    \end{subfigure}
    
    \centering
    \begin{subfigure}[b]{\textwidth}
        \centering
        \includegraphics[width=1.0\textwidth]{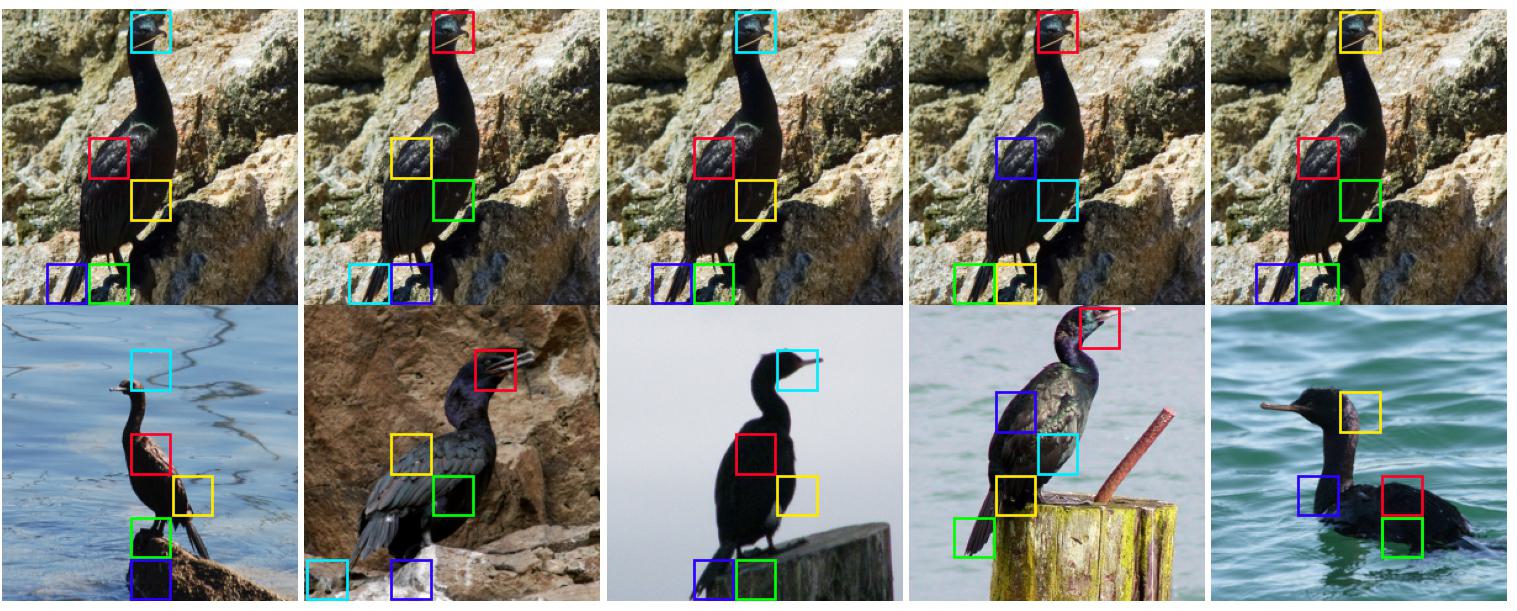}
        \caption{The explanation of a \textcolor{OliveGreen}{correct} prediction by CHM-Corr+.}
        \label{fig:method4_method4plus3_b}
    \end{subfigure}
    
    \caption{A \class{Pelagic Cormorant} \textcolor{red}{misclassified} as a \class{Red Faced Cormorant} by CHM-Corr. The face of the bird was not among the top-5 correspondences picked by CHM-Corr, which led to misclassification. The CHM-Corr+ classifier correctly classifies this image.}
    \label{fig:method4_method4plus3}
\end{figure*}

\clearpage

\section{Samples for ImageNet-Sketch dataset}
\label{supp:imagenet_sketch}

\begin{figure*}[!hbt]

  \centering
    \begin{subfigure}[b]{\textwidth}
        \centering
        \includegraphics[width=0.2\textwidth]{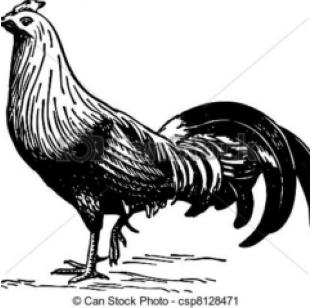}
        \caption{Query -- \class{Cock}}
    \end{subfigure}
    
    \centering
    \begin{subfigure}[b]{\textwidth}
        \centering
        \includegraphics[width=1.0\textwidth]{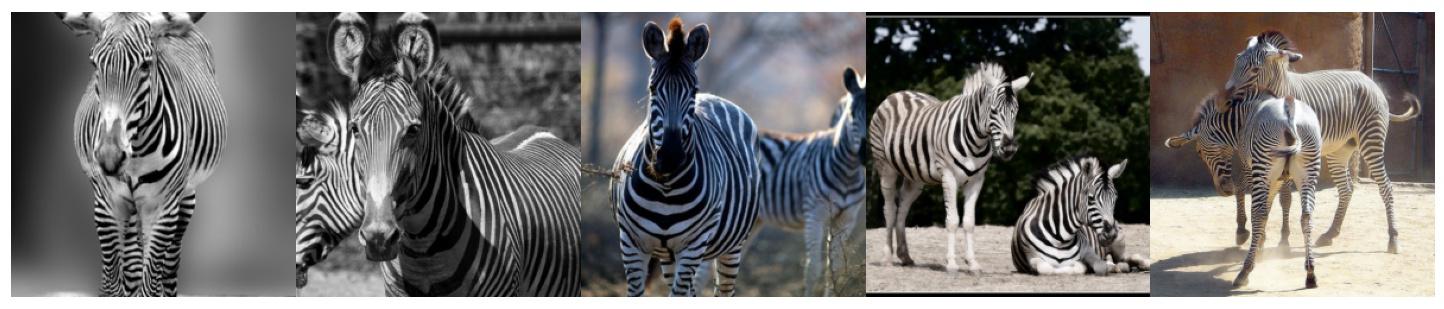}
        \caption{Nearest neighbors using kNN}
    \end{subfigure}
    
    \hfill
    \begin{subfigure}[b]{\textwidth}
        \centering
        \includegraphics[width=1.0\textwidth]{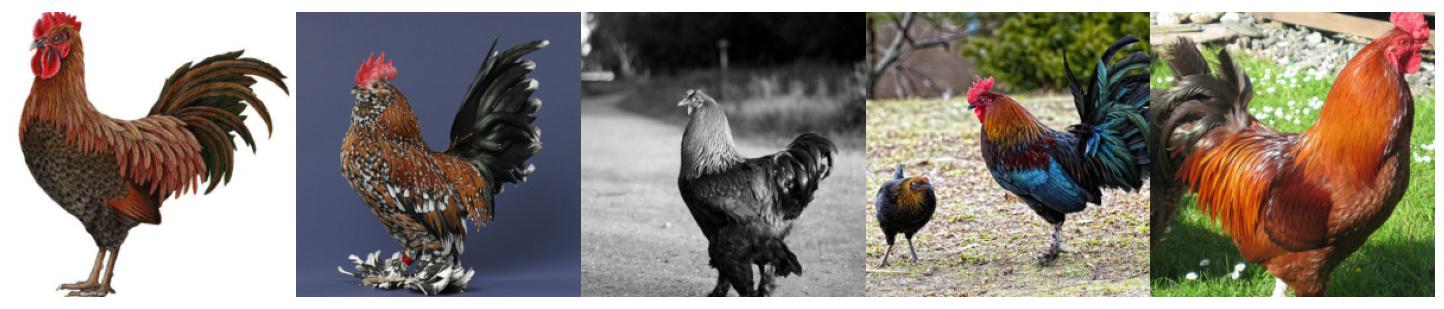}
        \caption{Nearest neighbors after re-ranking using CHM-Corr}
    \end{subfigure}

    \hfill
    \begin{subfigure}[b]{\textwidth}
        \centering
        \includegraphics[width=1.0\textwidth]{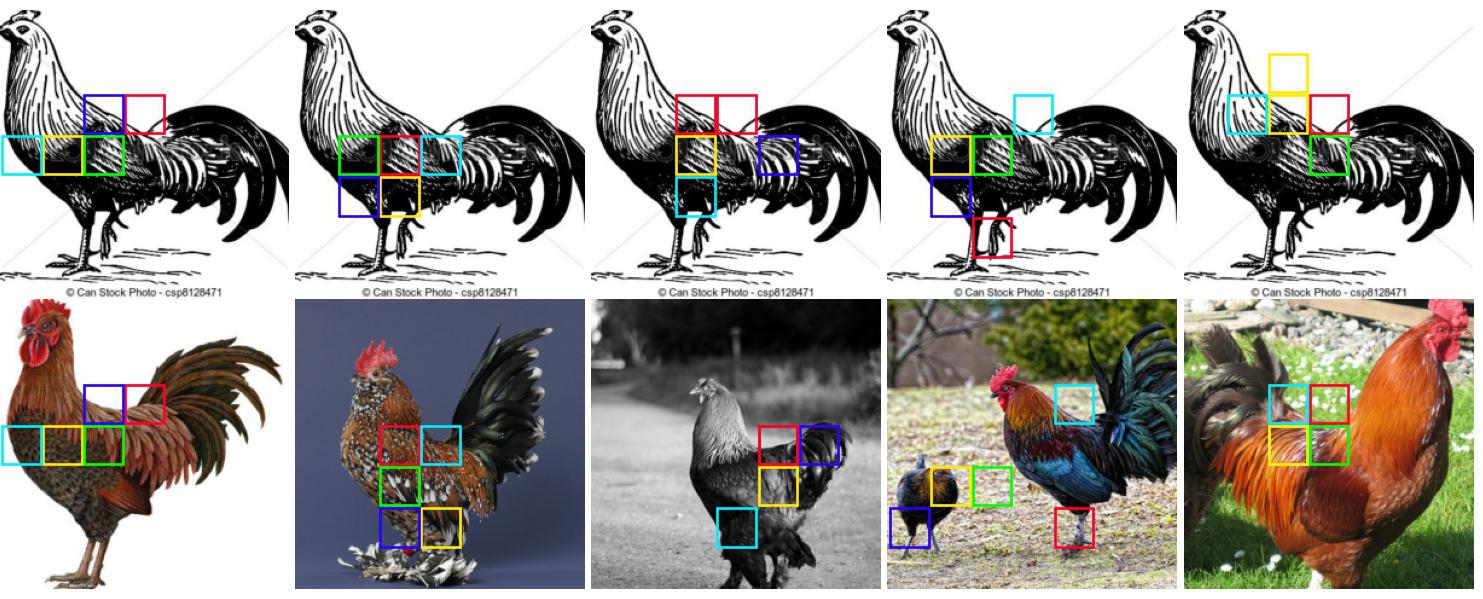}
        \caption{CHM-Corr explanation}
    \end{subfigure}

    \caption{A \textcolor{red}{misclassification} by the kNN classifier. 
    The black-and-white stripe patterns in \class{cock} confuse the kNN classifier, while the CHM-Corr classifier correctly labels the query.}
    \label{fig:imagenet_sketch_sample_1}
\end{figure*}

\begin{figure*}[!hbt]

  \centering
    \begin{subfigure}[b]{\textwidth}
        \centering
        \includegraphics[width=0.2\textwidth]{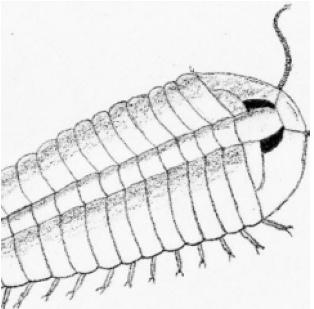}
        \caption{Query -- \class{Trilobite}}
    \end{subfigure}
    
    \centering
    \begin{subfigure}[b]{\textwidth}
        \centering
        \includegraphics[width=1.0\textwidth]{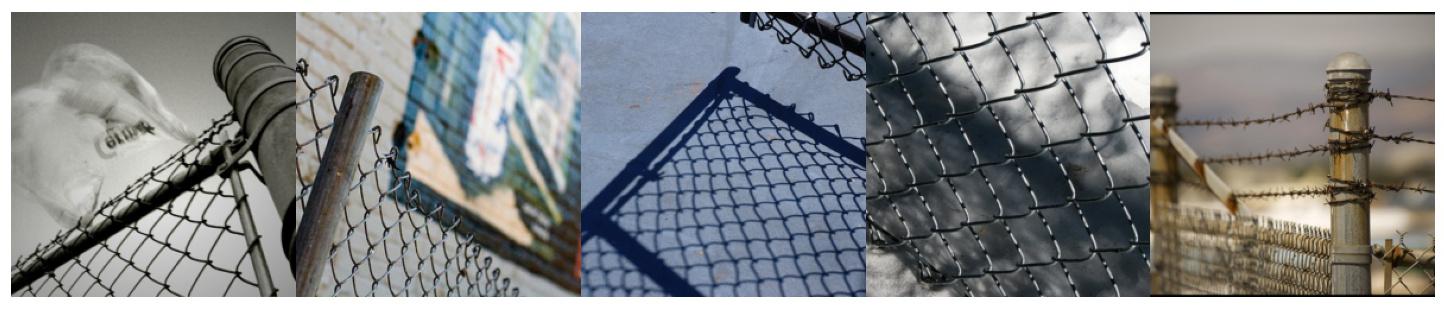}
        \caption{Nearest neighbors using kNN}
    \end{subfigure}
    
    \hfill
    \begin{subfigure}[b]{\textwidth}
        \centering
        \includegraphics[width=1.0\textwidth]{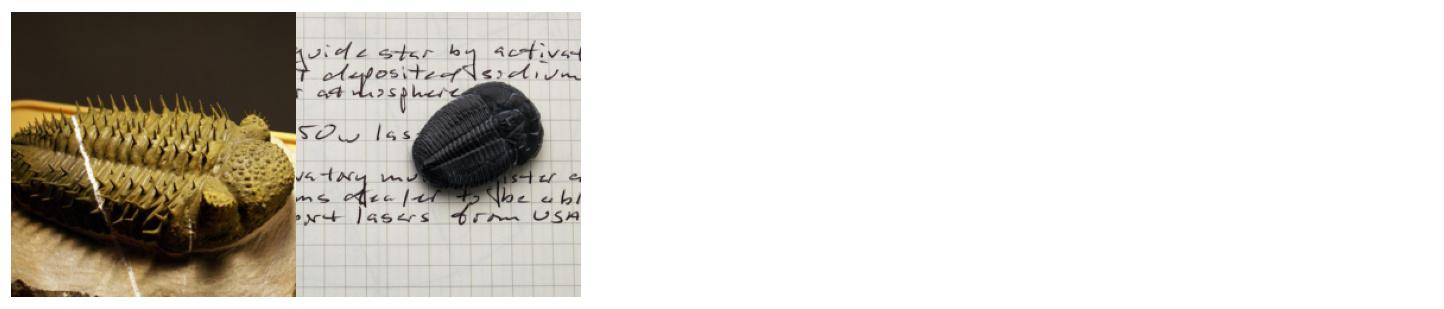}
        \caption{Nearest neighbors after re-ranking using CHM-Corr}
    \end{subfigure}

    \hfill
    \begin{subfigure}[b]{\textwidth}
        \centering
        \includegraphics[width=1.0\textwidth]{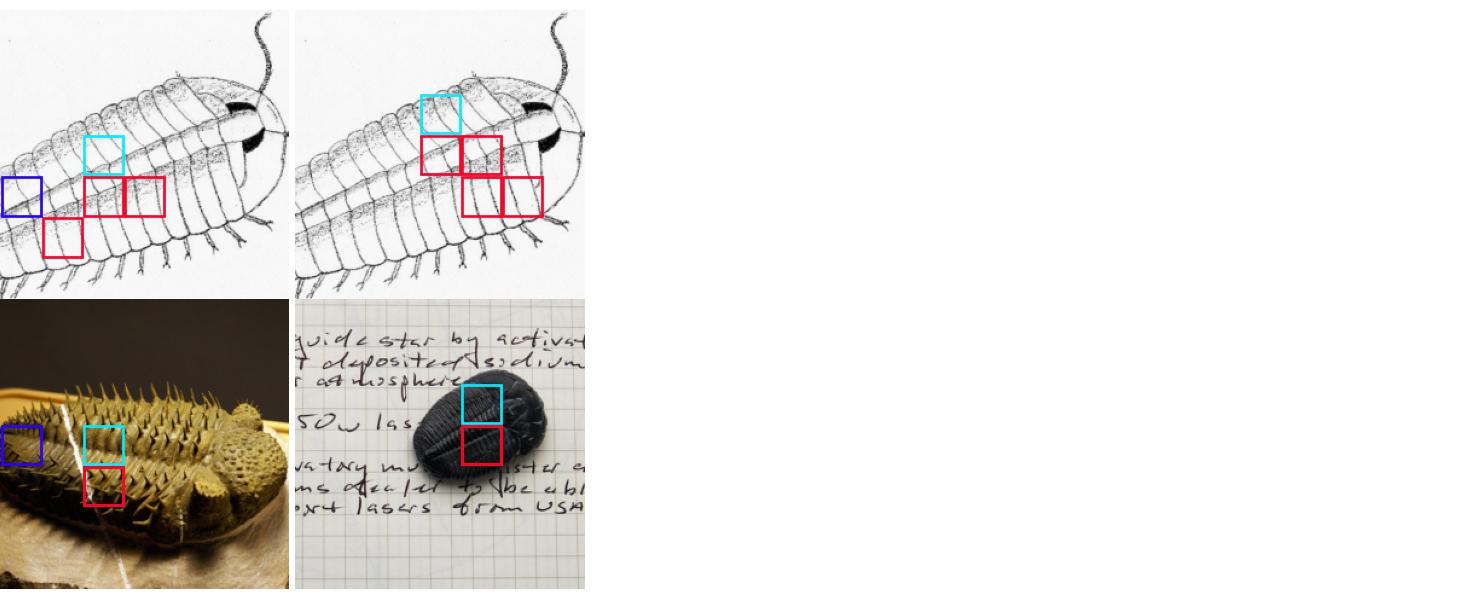}
        \caption{CHM-Corr explanation}
    \end{subfigure}

    \caption{A \textcolor{red}{misclassification} by the kNN classifier. 
    An image of  \class{trilobite} misclassified as a \class{chainlink fence} by the kNN classifier, while the CHM-Corr classifier correctly classifies the query. 
    The confidence score of CHM-Corr is only 2/20, \ie, 10\%.
    That is, only two \class{trilobite} images are among the top 20 candidates.
    }
    \label{fig:imagenet_sketch_sample_2}
\end{figure*}

\clearpage
\section{Removing duplicated images from the ImageNet validation set}
\label{supp:duplicates}

\begin{figure*}[!hbt]

  \centering
    \begin{subfigure}[b]{\textwidth}
        \centering
        \includegraphics[width=1.0\textwidth]{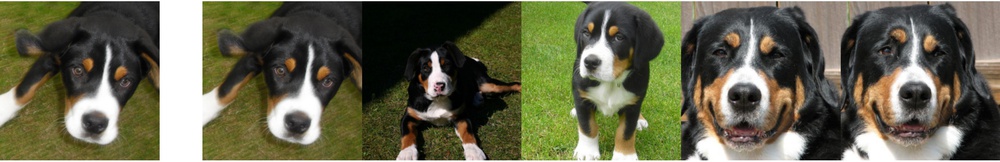}
        \caption{Query: \texttt{ILSVRC2012\_val\_00009877.JPEG}}
    \end{subfigure}
    
    \centering
    \begin{subfigure}[b]{\textwidth}
        \centering
        \includegraphics[width=1.0\textwidth]{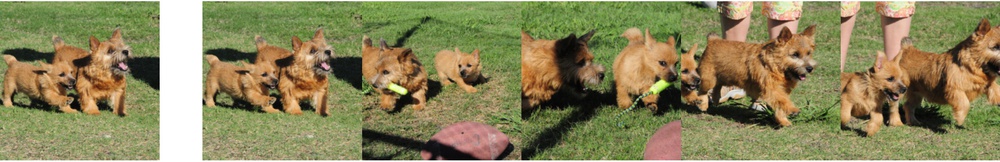}
        \caption{Query: \texttt{ILSVRC2012\_val\_00017380.JPEG}}
    \end{subfigure}
    
    \hfill
    \begin{subfigure}[b]{\textwidth}
        \centering
        \includegraphics[width=1.0\textwidth]{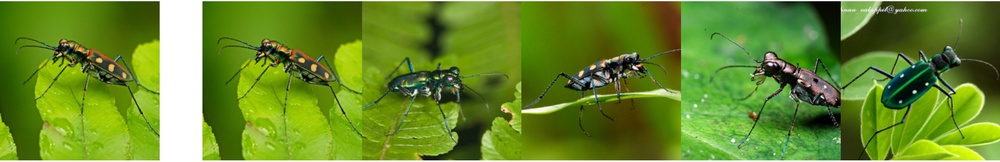}
        \caption{Query:\texttt{ILSVRC2012\_val\_00020013.JPEG}}
    \end{subfigure}

    \hfill
    \begin{subfigure}[b]{\textwidth}
        \centering
        \includegraphics[width=1.0\textwidth]{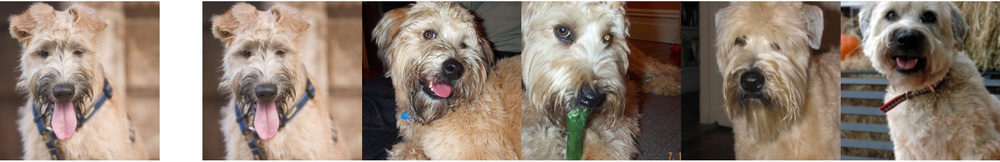}
        \caption{Query: \texttt{ILSVRC2012\_val\_00024875.JPEG}}
    \end{subfigure}


    \hfill
    \begin{subfigure}[b]{\textwidth}
        \centering
        \includegraphics[width=1.0\textwidth]{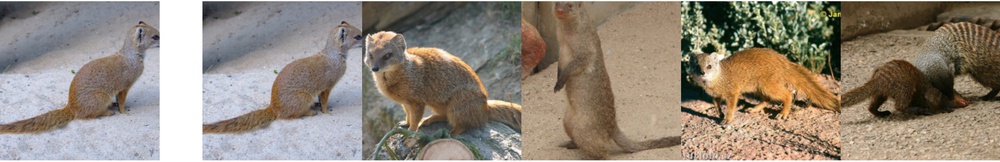}
        \caption{Query: \texttt{ILSVRC2012\_val\_00046136.JPEG}}
    \end{subfigure}

    \hfill
    \begin{subfigure}[b]{\textwidth}
        \centering
        \includegraphics[width=1.0\textwidth]{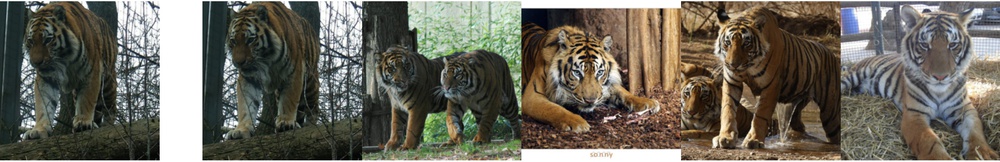}
        \caption{Query: \texttt{ILSVRC2012\_val\_00014815.JPEG}}
    \end{subfigure}
    
    \caption{In each panel, the leftmost image is in the validation set, and the top-5 nearest images on the right are from the training set.
    We find images that exist both in the training set and validation set and remove them from our validation set (in order to not unfairly favor kNN in the study).
    }
    \label{fig:duplicates_1}
\end{figure*}

Some of the images from the ImageNet validation set are also present in the training set. For the human study, we excluded such images from our study. Figure \ref{fig:duplicates_1}, shows some of these samples along with their five nearest neighbor images from the training set.

\section{Human-AI teams outperform both AIs alone and humans alone}
\label{sec:human_ai_team}

In \cref{sec:result_finding4}, we find that user classification accuracy can be improved when humans are provided with AI predictions and explanations.
Here, we leverage the same data collected from the previous ImageNet-ReaL and CUB human studies (\cref{sec:result_finding3,sec:result_finding4}) to estimate the accuracy of a human-AI team that allows both humans and AIs to make final decisions (\cref{fig:interaction-model}; Model 2).

That is, AIs make binary Yes/No predictions on all the $X$\% of query images that they assign a high confidence score $\geq T$ where $T \in [0,1]$ (\cref{fig:interaction-model}b).
Given these images and AI predictions, we compute an accuracy score, $\accai$.
For the set of remaining images (\ie whose AI confidence is $< T$), we take their user predictions and also compute an accuracy score $\acchuman$.
We define the human-AI team accuracy $\accteam$ as: 

\vspace*{-0.3cm}
\[
\accteam = X\% \times \accai + (100 - X)\% \times \acchuman
\]

As the interaction model 2 is more practical and scalable, it is interesting to test how the $\accteam$ compares with the accuracy when users or AIs alone classify all images (\ie when $X = 0$ or $100$).

\paragraph{Experiment}
For each classifier (ResNet-50, kNN, EMD-Corr, and CHM-Corr), we use a 2K-image held-out subset of the ImageNet-ReaL validation\footnote{Because the training set is used by non-parametric classifiers during testing, for ImageNet, we tune $T$ using 2K-image validation images with ImageNet-ReaL labels.
We use 1K test images for CUB.} set to find an optimal threshold $T$ that maximizes the classifier's binary classification accuracy.
Then, we use the remaining $\sim$42K ImageNet-ReaL validation images for testing.
For CUB, we tune $T$ using 1K test images and test on the remaining $\sim$4.7K test-set images.
For both ImageNet and CUB, we did not use the training sets to tune $T$ because the top-1 neighbors retrieved by kNN would be identical to the query all the time, biasing the AIs as well as humans when they perform classification.

After obtaining an AI accuracy score for each value of $T \in \{ 0.05, 0.10, 0.15, ..., 0.95 \}$, we find the best $\accteam$ (at an optimal $T^*$) and repeat the same process to find the best AI-only accuracy.
More details of the sweeps are in \cref{supp:complementary_perf}.

\subsec{Results}
First, across all four classifiers and two datasets, {AI-only accuracy is consistently higher than human-only accuracy} (\cref{table:aialone_and_humanteam} vs. \cref{table:humanaloneperformance}).
That is, letting users make all the AI-assisted decisions is both more labor-intensive and less accurate compared to allowing AIs to classify all the data themselves.
This result is consistent with the prior studies that find AIs to outperform humans \cite{ghaeini2018interpreting,zhang2020effect,ribeiro2016should,feng2019can} (see \cite{bansal2021does} for a summary).

Second, interestingly, human-AI teams consistently outperform the AIs alone (\cref{table:aialone_and_humanteam}) and humans alone (\cref{table:humanaloneperformance}).
That is, lay-users may be considered ``expert'' on ImageNet's everyday objects and therefore, when teaming up with humans to form human-AI teams, the accuracy substantially increases on average by \increasenoparent{2.11} (\cref{table:aialone_and_humanteam}).
On CUB, which is more challenging to lay-users, this benefit of teaming up with users is negligible (\cref{table:aialone_and_humanteam}; \increasenoparent{0.02})

Third, among four classifiers, ResNet-50 yields the highest human-AI team accuracy on both ImageNet-ReaL and CUB (\cref{table:aialone_and_humanteam}).
However, the variance in team accuracy across the classifiers is small. 
Our results interestingly imply that while there is significant evidence that Corr explanations are useful to the AI-assisted decision-making of humans in the interaction model 1 (\cref{table:humanaloneperformance}; CUB), such benefits of XAI models average out in the interaction model 2.




\end{document}